%% file: main.tex
\documentclass{article}

\input{preamble}

\usepackage[preprint]{icml2026}

\usepackage{amsmath}
\usepackage{amssymb}
\usepackage{mathtools}
\usepackage{amsthm}

\theoremstyle{plain}
\newtheorem{theorem}{Theorem}[section]

\theoremstyle{definition}
\newtheorem{definition}[theorem]{Definition}
\theoremstyle{remark}

\input{macros}
\icmltitlerunning{\method: Hyperproperty-Guided Multi-Agent Reinforcement Learning under 
  Partial Observation}

\begin{document}

\twocolumn[
  \icmltitle{\method: Hyperproperty-Guided Multi-Agent Reinforcement Learning  \\ under 
  Partial Observation}



  \icmlsetsymbol{equal}{*}

  \begin{icmlauthorlist}
    \icmlauthor{Arshia Rafieioskouei}{x}
    \icmlauthor{Tzu-Han Hsu}{x}
    \icmlauthor{Matthew Lucas}{x}
    \icmlauthor{Borzoo Bonakdarpour}{x}
  \end{icmlauthorlist}

  \icmlaffiliation{x}{Department of Computer Science and Engineering, Michigan State University, MI, USA}

  \icmlcorrespondingauthor{Arshia Rafieioskouei}{rafieios@msu.edu}
  \icmlcorrespondingauthor{Borzoo Bonakdarpour}{borzoo@msu.edu}

  \icmlkeywords{Reinforcement Learning, Multi-agent Reinforcement Learning, Specification-Based Reinforcement Learning, Formal Methods, Hyperproperty, HyperLTL, StarCraft Benchmark (SMAC)}

  \vskip 0.3in
]



\printAffiliationsAndNotice{}  

\begin{abstract}
	
Formal specification is a powerful tool to guide the learning process and provides significant advantages over reward shaping: 
(1) mathematical rigor; (2) expressiveness to specify objectives and constraints, and 
(3) the ability to define tactics to achieve objectives.
However, these benefits remain largely unexplored in the context of Multi-Agent Reinforcement Learning (MARL).
This paper introduces \method, a novel framework for MARL under {\em partial observability}, where learning is guided by the expressive power of the so-called {\em hyperproperties} and, in particular, the temporal logic \HyperLTL.
We integrate Centralized Training for Decentralized Execution (CTDE) techniques with \method to synthesize \emph{decentralized} policies, and our evaluation on SMAC, MessySMAC, and WildFire benchmark demonstrates clear advantages over baselines.
\renewcommand{\thefootnote}{}
\footnote{\href{https://arshiarafiei.github.io/HyPOLE/}{
     \raisebox{-0.15em}{\includegraphics[height=1.0em]{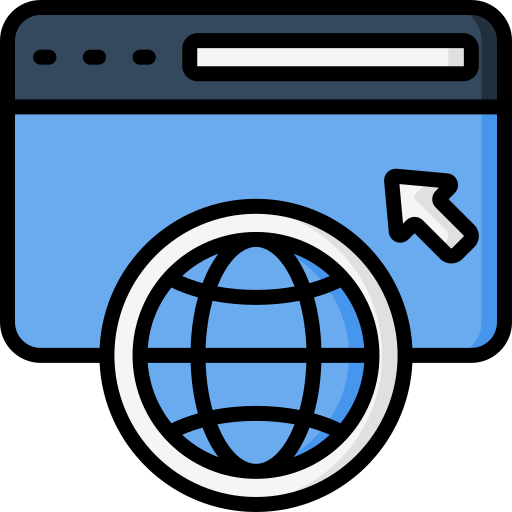}} Paper Webpage
}}
\end{abstract}
\input{intro}

\input{related}

\input{prelim}

\input{problem}
\input{algorithm}

\input{experiment}

\input{conc}

\newpage
\newpage
\section*{Acknowledgment} 
This work is partially sponsored by the United States NSF Award SaTC 2245114.
\section*{Impact Statement} 
This work takes a step toward bringing expressive specifications, in the form of hyperproperties, into practical MARL for partially observable and uncontrolled environments.
By allowing tactics to be stated as precise mathematical formulas, \method can support the design of interpretable and verifiable behaviors in autonomous domains such as drone swarms.
At the same time, the use of expressive specifications introduces potential risks.
Malicious or adversarial specifications could be used to induce harmful behaviors, and poorly designed specifications may lead to unintended consequences.
These risks highlight the importance of careful specification design, validation, and human oversight when deploying specification-guided MARL systems.

\bibliographystyle{icml2026}
\bibliography{bibliography,references}

\newpage
\appendix
\onecolumn

\input{theory-appendix.tex}

\input{exp-appendix}

\end{document}

%% file: preamble.tex
\usepackage[dvipsnames]{xcolor}
\usepackage[T1]{fontenc}
\usepackage[utf8]{inputenc} 
\usepackage[T1]{fontenc}    
\usepackage{hyperref}       
\usepackage{url}            
\usepackage{booktabs}       
\usepackage{amsfonts}       
\usepackage{nicefrac}       
\usepackage{microtype}      
\usepackage{mathtools}
\usepackage{bm}
\usepackage{amssymb}
\usepackage{tikz}
\usepackage{dsfont}
\usepackage{pgfplots}
\usepgfplotslibrary{fillbetween}
\usetikzlibrary{patterns}
\usetikzlibrary{arrows.meta, positioning}
\usepackage{graphicx} 
\usepackage{hyperref} 

\usepackage{xspace}
\usepackage{microtype}

\usepackage{makecell}
\usepackage{orcidlink}

\usepackage{graphicx}
\usepackage{amsmath, amssymb}

\usepackage{lua-visual-debug}
\usepackage{wrapfig}
\usepackage{mdframed}
\usepackage{hyperref}
\hypersetup{%
	colorlinks=true,           
	allcolors=brown!70!black,   
	pdfstartview=Fit,          
	breaklinks=true,
	pdfauthor={},
	pdftitle={{}}
}

\usepackage[capitalize,noabbrev]{cleveref}
\crefname{figure}{Fig.}{Figs.}
\Crefname{figure}{Fig.}{Figs.}

\crefname{table}{Table}{Tabs.}
\Crefname{table}{Table}{Tabs.}

\crefname{section}{Sec.}{Secs.}
\Crefname{section}{Sec.}{Secs.}

\crefname{equation}{Eq.}{Eqs.}
\Crefname{equation}{Eq.}{Eqs.}
\crefname{appendix}{App.}{Apps.}
\Crefname{appendix}{App.}{Apps.}

\usepackage{stmaryrd}
\usepackage{amsfonts}
\usepackage{marvosym}
\usepackage{multirow}

\usepackage{bussproofs}
\usepackage{booktabs}
\usepackage{colortbl}

\usepackage{listings}

\usepackage{subcaption}
\usepackage[textsize=tiny]{todonotes}
\usepackage{graphicx}

\usepackage{stmaryrd}

\usepackage{ltl}

\definecolor{gray}{rgb}{0.5,0.5,0.5}
\definecolor{darkgreen}{rgb}{0,0.4,0}
\definecolor{darkblue}{rgb}{0.2, 0.2, 0.6}
\definecolor{niceblue}{rgb}{0.16, 0.32, 0.75}
\definecolor{niceblack}{rgb}{0.0, 0.18, 0.39}
\definecolor{prettyred}{rgb}{1.0, 0.13, 0.32}
\definecolor{prettypink}{rgb}{0.98, 0.38, 0.5}
\definecolor{prettyblue}{rgb}{0.16, 0.32, 0.75}
\definecolor{prettygreen}{rgb}{0.0, 0.5, 0.0}
\definecolor{prettyorange}{rgb}{0.99, 0.53, 0.05}
\definecolor{prettypurple}{rgb}{0.6, 0.4, 0.8}
\definecolor{prettyyellow}{rgb}{0.99, 0.93, 0.0}
\definecolor{airforceblue}{rgb}{0.36, 0.54, 0.66}
\definecolor{mygreen}{HTML}{0BB357}

\usepackage[many]{tcolorbox}    	
\usepackage{multicol}               

\colorlet{main}{black}

\colorlet{sub}{main!30}

\tcbset{
	sharp corners,
	colback = white,
	before skip = 0.2cm,    
	after skip = 0.5cm      
}                           

\newtcolorbox{boxA}{
	fontupper = \bf,
	boxrule = 1.5pt,
	colframe = black 
}

\newtcolorbox{boxB}{
	fontupper = \bf\color{main}, 
	boxrule = 1.5pt,
	colframe = main,
	rounded corners,
	arc = 5pt   
}

\newtcolorbox{boxC}{
	colback = sub, 
	boxrule = 0pt  
}

\newtcolorbox{boxD}{
	colback = sub, 
	colframe = main, 
	boxrule = 0pt, 
	toprule = 3pt, 
	bottomrule = 3pt 
}

\newtcolorbox{boxE}{
	enhanced, 
	boxrule = 0pt, 
	borderline = {0.75pt}{0pt}{main}, 
	borderline = {0.75pt}{2pt}{sub} 
}

\newtcolorbox{boxF}{
	colback = sub,
	enhanced,
	boxrule = 1.5pt, 
	colframe = white, 
	borderline = {1.5pt}{0pt}{main, dashed} 
}

\newtcolorbox{boxG}{
	enhanced,
	boxrule = 0pt,
	colback = sub,
	borderline west = {1pt}{0pt}{main}, 
	borderline west = {0.75pt}{2pt}{main}, 
	borderline east = {1pt}{0pt}{main}, 
	borderline east = {0.75pt}{2pt}{main}
}

\newtcolorbox{boxH}{
	colback = sub, 
	colframe = main, 
	boxrule = 0pt, 
	leftrule = 6pt 
}

\newtcolorbox{boxI}{
	colback = sub, 
	colframe = main, 
	boxrule = 0pt, 
	toprule = 6pt 
}

\newtcolorbox{boxJ}{
	sharpish corners, 
	colback = sub, 
	colframe = main, 
	boxrule = 0pt, 
	toprule = 4.5pt, 
	enhanced,
	fuzzy shadow = {0pt}{-2pt}{-0.5pt}{0.5pt}{black!35} 
}

\newtcolorbox{boxK}{
	sharpish corners, 
	colframe =prettyblue!50,
	boxrule = 0pt,
	toprule = 4.5pt, 
	enhanced,
	fuzzy shadow = {0pt}{-2pt}{-0.5pt}{0.5pt}{prettyblue!50} 
}

\newtcolorbox{boxL}{
	fontupper = \color{main},
	rounded corners,
	arc = 6pt,
	colback = sub, 
	colframe = main!50, 
	boxrule = 0pt, 
	bottomrule = 4.5pt 
}

\newtcolorbox{boxM}{
	fontupper = \color{white},
	rounded corners,
	arc = 6pt,
    colback=blue!5,
	colback = main!80, 
	colframe = main, 
	boxrule = 0pt, 
	bottomrule = 4.5pt,
	enhanced,
	fuzzy shadow = {0pt}{-3pt}{-0.5pt}{0.5pt}{black!35}
}

\newtcolorbox{problemstatement}[1][]{
    enhanced,
    title={Problem Statement},
    colback=blue!3,
    colframe=blue!50!black!80,
    coltitle=white,
    colbacktitle=blue!50!black!80,
    fonttitle=\bfseries,
    boxrule=1pt,
    arc = 8mm,
    left=2pt,
    right=2pt,
    top=5pt,
    bottom=2pt
}

\newtcolorbox{theoremboxstyle}{
  enhanced,
  colback=green!8,
  colframe=green!60!black,
  boxrule=1pt,
  arc=8mm,
  left=2pt, right=2pt,
  top=5pt, bottom=2pt
}

\newtcolorbox{bluebox}[1][]{
  enhanced,
  colframe=blue!70!black,  
  boxrule=1pt,
  colback=blue!2,
  arc=4mm,                 
  left=1mm, right=1mm, top=0mm, bottom=0.5mm,
  boxsep=0mm,
  #1
}
\newtcolorbox{bluebox2}[1][]{
  enhanced,
  colframe=blue!70!black,  
  boxrule=1pt,
  arc=4mm,                 
  left=2mm, right=2mm, top=2mm, bottom=2mm,
  boxsep=2mm,
  #1
}

\crefname{theorem}{Theorem}{Theorems}
\Crefname{theorem}{Theorem}{Theorems}

\newenvironment{theoremstatement}[2][]{%
  \refstepcounter{theorem}%
  \begin{tcolorbox}[
    enhanced,
    breakable,
    title={Theorem~\thetheorem%
      \if\relax\detokenize{#2}\relax\else\ (\textit{#2})\fi},
    colback=red!4,
    colframe=red!50!black!80,
    coltitle=white,
    colbacktitle=red!60!black!80,
    fonttitle=\bfseries,
    boxrule=1pt,
    arc=8mm,
    left=2pt,
    right=2pt,
    top=5pt,
    bottom=2pt,
    #1
  ]%
}{%
  \end{tcolorbox}%
}

\definecolor{problemblue}{HTML}{C6D8FF}

\newtcolorbox{problemstatementone}[1][]{
    enhanced,
    blanker,
    left=2pt,
    right=2pt,
    top=5pt,
    bottom=2pt,
    #1
}

%% file: macros.tex
{}

{}

{}

\definecolor{c1}{HTML}{602273}
\definecolor{c2}{HTML}{F39F5A}
\definecolor{c3}{HTML}{AE445A}
\definecolor{c4}{HTML}{6e4909}

\colorlet{statementColor}{c3}

\usepackage{scalefnt}

\colorlet{command-color}{black!70}
\definecolor{dkcyan}{rgb}{0.1, 0.3, 0.3}
\definecolor{dkgreen}{rgb}{0,0.3,0}
\definecolor{dkblue}{rgb}{0,0.3,1.0}
\colorlet{comment-color}{black!50}

\lstdefinelanguage{code-lang}{
	keywords={def, repeat, return, if, then, else,or},
	keywordstyle=[1]\color{command-color},
	morekeywords=[2]{faultLocalization,instrument, symbolicExecution,SyGuS},
	keywordstyle=[2]\color{dkgreen},
	morekeywords=[3]{iterativeRepair},
	keywordstyle=[3]\color{dkcyan},
	comment=[l][\color{comment-color}]{//},
	literate=%
	{=}{{{\color{command-color}=}}}1
	{|}{{{\color{command-color}|}}}1
	{:}{{{\color{command-color}:}}}1
	{:=}{{{\color{command-color}:=}}}1
	{@}{ }1
}

\lstdefinestyle{code-style}{
	escapeinside={(*}{*)},
	basicstyle=\ttfamily\fontsize{9}{11}\selectfont,
	columns=fullflexible,
	commentstyle=\sffamily\color{black!50!white},
	framexleftmargin=1em,
	framexrightmargin=1ex,
	keepspaces=true,
	mathescape,
	numbers=left,
	numberblanklines=false,
	numbersep=0.5em,
	numberstyle=\relscale{0.65}\color{gray}\ttfamily,
	showstringspaces=true,
	stepnumber=1,
	xleftmargin=1.2em,
}

\lstdefinelanguage{example-lang}{
	keywords={while,do,if, then,else, observe,or},
	keywordstyle=[1]{\color{c3}},
	morekeywords=[2]{string,int,bool},
	keywordstyle=[2]\color{black!60},
	morekeywords=[3]{phase,title,session,decision,print,password,attack,request,username,date,userPassword,info,LOG,credentials,reviewA,reviewB, reviewerAid,reviewerBid,order,notification,withdraw,balance,amount,ErrorLog,TransactionLog},
	keywordstyle=[3]\color{c1},
	morekeywords=[4]{366,1,0,true},
	keywordstyle=[4]\color{c4},
	morestring=[b]",
	stringstyle=\color{c4},
	comment=[l][\color{comment-color}]{//},
	literate=%
	{=}{{{\color{c3}=}}}1
	{!=}{{{\color{c3}!=}}}1
	{>}{{{\color{c3}>}}}1
	{<}{{{\color{c3}<}}}1
	{+}{{{\color{c3}+}}}1
	{-}{{{\color{c3}-}}}1
	{*}{{{\color{c3}*}}}1
	{@}{ }1
}

\lstdefinestyle{example-style}{
	escapeinside={(*}{*)},
	basicstyle=\ttfamily\fontsize{8}{10}\selectfont,
	columns=fullflexible,
	commentstyle=\sffamily\color{black!50!white},
	framexleftmargin=0em,
	framexrightmargin=0ex,
	keepspaces=true,
	mathescape,
	numbers=left,
	numberblanklines=false,
	numbersep=1.0em,
	numberstyle=\relscale{0.65}\color{gray}\ttfamily,
	showstringspaces=true,
	stepnumber=1,
	xleftmargin=1.2em
}

\lstdefinestyle{example-style-large}{
	escapeinside={(*}{*)},
	basicstyle=\ttfamily\fontsize{10}{12}\selectfont,
	columns=fullflexible,
	commentstyle=\sffamily\color{black!50!white},
	framexleftmargin=0em,
	framexrightmargin=0ex,
	keepspaces=true,
	mathescape,
	numbers=left,
	numberblanklines=false,
	numbersep=0.5em,
	numberstyle=\relscale{0.65}\color{gray}\ttfamily,
	showstringspaces=true,
	stepnumber=1
}

\lstnewenvironment{code}[1][]
{\small
	\lstset{
		style=code-style, 
		language=code-lang,
		#1
	}
}
{}

\lstnewenvironment{exampleCode}[1][]
{\small
	\lstset{
		style=example-style, 
		language=example-lang,
		#1
	}
}
{}

\newcommand{\AP}{\mathsf{AP}}

\newcommand{\realnum}{\mathbb{R}}

\newcommand{\integers}{\mathbb{Z}}

\newcommand{\finite}{*}

\newcommand{\definedas}{\triangleq}

\newcommand{\pomdp}{\mathcal{M}}

\newcommand{\states}{S}
\newcommand{\state}{\mathit{s}}
\newcommand{\labels}{\mathit{L}}
\newcommand{\labelsof}[1]{\mathit{L}(#1)}

\newcommand{\actions}{A}
\newcommand{\action}{\mathit{a}}

\newcommand{\tran}{P}
\newcommand{\init}{I}
\newcommand{\rew}{R}
\newcommand{\ObsSym}{O}
\newcommand{\ObsFun}{{\mathcal{O}}}
\newcommand{\actobshist}{\mathcal{H}}
\newcommand{\obs}{o}

\newcommand{\edgeof}[1]{\xrightarrow{#1}}
\newcommand{\prob}{\mathbb{P}}

\newcommand{\optimal}{\star}

\newcommand{\maxpolicy}{\pi^{\optimal}}
\newcommand{\policy}{\pi}

\newcommand{\traj}{\zeta}
\newcommand{\trajs}{\mathcal{Z}}

\newcommand{\tracevar}{\tau}

\newcommand{\HyperLTL}{{HyperLTL}\xspace}

\newcommand{\LTL}{{LTL}\xspace}

\newcommand{\trajfromto}[3]{{#1}_{\bracketof{{#2}:{#3}}}}

\newcommand{\predicatefunc}{\mathit{f}}
\newcommand{\predicatefuncof}[1]{\predicatefunc\big({#1}\big)}
\newcommand{\constant}{\mathit{c}}

\newcommand{\tr}{\mathit{T}}
\newcommand{\Tr}{\mathcal{T}}
\newcommand{\trace}{\mathit{t}}
\newcommand{\traceof}[1]{\mathsf{Tr}(#1)}
\newcommand{\traceoff}[1]{\mathsf{Tr}\big(#1\big)}
\newcommand{\tracesof}[1]{\mathsf{Traces}(#1)}
\newcommand{\tracesoff}[1]{\mathsf{Traces}\big(#1\big)}

\newcommand{\OR}{\vee}

\newcommand{\always}{\G}
\newcommand{\until}{\mathbin{\mathcal{U}}}

\newcommand{\tru}{\mathsf{true}}
\newcommand{\fals}{\mathsf{false}}
\newcommand{\G}{\LTLsquare}
\newcommand{\F}{\LTLdiamond}
\newcommand{\X}{\LTLcircle}
\newcommand{\U}{\mathbin{\mathcal{U}}}

\newcommand\modified[1]{#1}

\newcommand{\traceVars}{\mathit{Vars(\varphi)}}
\newcommand{\traceVarsOf}[1]{\mathit{Vars(#1)}}

\newcommand{\var}{p}

\newcommand{\quant}{\mathds{Q}}

\newcommand{\rbvalue}{{\rho}}
\newcommand{\rbvalueof}[1]{\rbvalue\big({#1}\big)}

\newcommand{\maximum}{\textsf{max}}
\newcommand{\minimum}{\textsf{min}}
\newcommand{\maxamong}[1]{\underset{#1}{\textsf{max}}}
\newcommand{\minamong}[1]{\underset{#1}{\textsf{min}}}
\newcommand{\distrubition}{\mathcal{D}}
\newcommand{\distributionof}[1]{\distrubition_{#1}}

\newcommand{\setof}[1]{ \{#1\} }
\newcommand{\skolemized}[1]{\mathbf{Skolem}(#1)}

\newcommand{\skolemfunc}{\mathbf{f}}

\newcommand{\allforall}{\mathds{\quant}}
\newcommand{\allexists}{\mathds{\quant}}
\newcommand{\forallsof}[1]{{\allforall^\forall_{#1}}}
\newcommand{\existsof}[1]{{\allexists^\exists_{#1}}}
\newcommand{\sizeof}[1]{{|#1|}}

\newcommand{\traceassignment}{\Pi}

\newcommand{\zip}{\textsf{zip}}
\newcommand{\zipof}[1]{\textsf{zip}(#1)}
\newcommand{\zipofk}[1]{\textsf{zip}_k(#1)}

\newcommand{\discount}{\gamma}
\newcommand{\initdec}{\mathcal{I}}
\newcommand{\jointhistory}{\mathbf{h}}
\newcommand{\jointqvalue}{\mathbf{Q}}

\newcommand{\leafindex}{k}

\newcommand{\expected}{\mathbb{E}}
\newcommand{\Qvalue}{\mathit{Q}}

\newcommand{\tupleof}[1]{\langle #1 \rangle}

\newcommand{\pos}{\texttt{pos}}
\newcommand{\posenm}[1]{\texttt{pos\_enm\_#1}}
\newcommand{\healthenm}[1]{\texttt{health\_enm\_#1}}
\newcommand{\healthenmprev}[1]{\texttt{health\_enm\_prev\_#1}}
\newcommand{\noenem}{\texttt{\#enm\_dead}}
\newcommand{\noenemprev}{\texttt{\#enm\_dead\_prev}}
\newcommand{\win}{\texttt{Win}}

\newcommand{\locmark}{\it} 
\newcommand{\ff}{\textsf{\small FF}\xspace} 
\newcommand{\med}{\textsf{\small Med}\xspace}

\newcommand{\motifdist}{\textsf{dist}}

\newcommand{\shootrate}{\texttt{s\_range}}
\newcommand{\obsrate}[1]{\texttt{o\_range\_enm\_#1}}

\newcommand{\steptrain}{\beta}

\newcommand*\spvertund[1]{\vrule width0pt height0pt depth#1\relax}

\newcommand{\orderedunion}{\sqcup}

\newcommand{\bracketof}[1]{ [ #1 ] }

\newcommand{\method}{\textsc{HyPOLE}\xspace}

\newcommand{\converge}{\overset{\optimal}{\rightarrow}}

\newcommand{\familyofsamples}{\mathbb{S}}
\newcommand{\imageof}[1]{\mathit{Img}({#1})}
\newcommand{\orderedsets}{\bowtie}

\newcommand{\decpomdp}{G}
\newcommand{\statesdec}{\mathcal{S}}
\newcommand{\actionsdec}{\mathcal{A}}
\newcommand{\trandec}{\mathcal{P}}
\newcommand{\rewdec}{\mathcal{R}}
\newcommand{\obsdec}{\mathfrak{O}}

\newcommand{\statedec}{s}
\newcommand{\actiondec}{\mathbf{a}}

\newcommand{\jointpolicy}{\boldsymbol{\pi}}
\newcommand{\Value}{\mathcal{V}}

\newcommand{\maptostate}{\Phi}

\newcommand{\traceThist}{\mathfrak{H}}
\newcommand{\buffer}{\mathcal{D}}
\newcommand{\seqseq}{\texttt{encoder--decoder}\xspace}
\newcommand{\fseq}{f_{\texttt{enc-dec}}}

\newcommand{\select}{\mathsf{Sel}}
\newcommand{\traceToHist}{\Omega}

\newcommand{\focus}{\varphi_{\text{focus-fire}}}
\newcommand{\medivac}{\varphi_{\text{medivac}}}
\newcommand{\kite}{\varphi_{\text{kite}}}
\newcommand{\corridor}{\varphi_{\text{corridor}}}
\newcommand{\defense}{\varphi_{\text{defense}}}
\newcommand{\bad}{\varphi_{\text{bad}}}
\newcommand{\wild}{\varphi_{\text{wildfire}}}

\newcommand{\orangeline}{(\textcolor{orange}{orange line})\xspace}
\newcommand{\redline}{(\textcolor{red}{red line})\xspace}
\newcommand{\blueline}{(\textcolor{blue}{blue line})\xspace}
\newcommand{\purpleline}{(\textcolor{Purple}{purple line})\xspace}
\newcommand{\brownline}{(\textcolor{Brown}{brown line})\xspace}
\newcommand{\grayline}{(\textcolor{Gray}{gray line})\xspace}
\newcommand{\pinkline}{(\textcolor{CarnationPink}{pink line})\xspace}
\newcommand{\greenline}{(\textcolor{Green}{green line})\xspace}

\newcommand{\modif}[1]{\textcolor{black}{#1}}

%% file: intro.tex
\vspace{-7mm}
\section{Introduction}
\label{sec:intro}

\begin{figure*}
    \centering
    \includegraphics[width=0.95\linewidth]{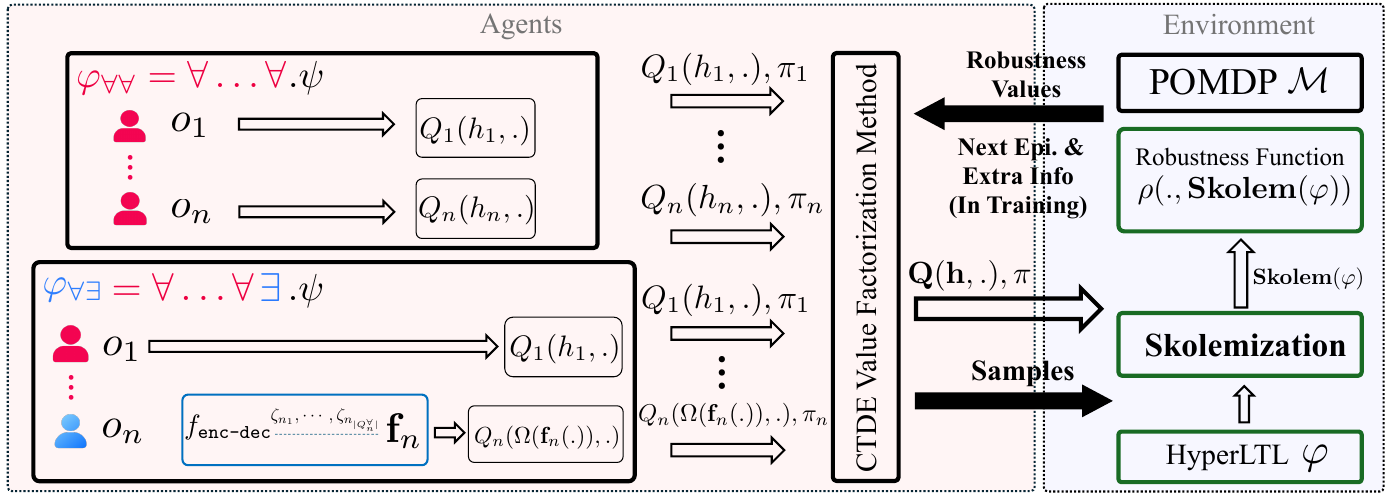}
    \captionof{figure}{Overview of \method.}
    \label{fig:hypole_method}
\end{figure*}

Multi-Agent Reinforcement Learning (MARL) is challenging due to combinatorial scaling, and becomes even more demanding when agents must satisfy multiple interdependent objectives and constraints.
Existing MARL techniques~\cite{pmlr-v80-rashid18a,sunehag2017valuedecompositionnetworkscooperativemultiagent,son2019qtranlearningfactorizetransformation} are strangers to {\em symbolic} techniques that provide a wealth of powerful specification languages that express complex objectives and constraints among 
agents and guide the learning process.
Moreover, shaped reward functions are often insufficient for expressing and guiding learning with respect to {\em relational} objectives among agents and their temporal behaviors~\cite{Spec-RL}.
More importantly, a particular limitation of existing techniques is that shaped rewards for a set of objectives are implicitly {\em universally quantified}; that is, the intended objective must hold for every pair of agents' behaviors.
This inherent universal strictness results in limiting the search space for better policies.

To explain the issue of universality, consider the following formula in predicate logic: $\varphi_1 
= \forall x.\forall y. (x < y)$, where $x$ and $y$ range over integers.
This formula is obviously false, but a weaker variation of this formula $\varphi_2 = \forall x.\exists y. (x < y)$ is valid.
In fact, synthesizing a policy to select $y$ in response to any choice of $x$ is to learn a witnessing {\em Skolem function}~\cite{skolem1920} such as $y = \mathbf{f}(x) = x + 1$.
In the world of MARL, virtually all existing approaches attempt to search for policies for objectives that are of the form $\forall\forall.\psi$, while many multi-agent requirements are of the form $\forall\exists.\psi$, expressing that for every behavior of one agent, there exists a behavior of another agent, such that the objective is satisfied.
For example, \cite{hsu2025hyprl,Beutner_Finkbeiner_2024} formulate multi-agent planning problems, using $\forall\exists.\psi$ specifications.
Moreover, \cite{hsu2025hyprl} extends beyond multi-agent planning by encoding the single-agent planning problem DeepSea Treasure~\cite{vamplew2011empirical} as a multi-agent problem under a $\forall\exists.\psi$ specification.

We propose a specification-guided method that boosts MARL under \emph{partial observability}.
We use the concept of {\em hyperproperties} and, in particular, the temporal logic 
\HyperLTL~\cite{cfkmrs14} as our specification language to express the intended behavior of each agent as well as their {\em relational} (inter-dependent) objectives and constraints. 
Our proposed method \method (\underline{\bf Hy}perproperties for \underline{\bf P}artially \underline{\bf O}bservable \underline{\bf L}earning \underline{\bf E}nvironments), is built on realistic multi-agent settings, assuming that 
agents operate in an environment modeled as a partially observable Markov decision process (POMDP) with the goal of learning a {\em decentralized} set of policies, one per agent (see~\Cref{fig:hypole_method}).

\paragraph{Our Contributions.} (1) We formulate the policy synthesis problem for a multi-agent system as a learning problem, where the goal is to maximize the probability of satisfying a \HyperLTL specification that expresses a set of objectives and constraints under partial observability (\Cref{sec:problem}).
(2) We solve this problem by first transforming symbolic satisfaction of the \HyperLTL specification for a POMDP into an optimization problem by defining continuous robustness functions, then use off-the-shelf Centralized Training for Decentralized Execution (CTDE) algorithms to learn an optimal collection of decentralized policies (\Cref{sec:algorithm}), and 
(3) we evaluate \method on scenarios from (i) the StarCraft~II Multi-Agent Challenge (SMAC)~\cite{samvelyan19smac}, (ii) MessySMAC~\cite{pmlr-v202-phan23a}, and (iii) the WildFire benchmark.
Our results show that \method coupled with CTDE algorithms is more efficient and effective 
than the vanilla versions in handling complex requirements. 
We also show how \HyperLTL assists in (1) faster learning of better policies, and (2) elegantly express various tactics (\Cref{sec:eval}).

%% file: related.tex
\section{Related Work}
\label{sec:related}

We believe that the work closest to \method is \textsc{HypRL}~\cite{hsu2025hyprl}, which studies hyperproperty-guided MARL in a fully observable and centralized setting. 
Both approaches leverage Skolemization, a well-known technique in logic for handling quantifier alternations. 
However, \method introduces a new challenge from the decentralized setting: the Skolem function does not have access to the traces of other agents when reasoning about its own behavior relative to theirs. 
Moreover, because \textsc{HypRL} operates in a centralized setting, it does not scale well with increasing environment complexity or the number of agents. In contrast, \method addresses this scalability challenge by learning decentralized policies in partially observable multi-agent environments. 
Other work on specification-guided RL include~\cite{jothimurugan2021compositional,li2017reinforcement,kuo2020encoding,de2019foundations,hasanbeig2018logically,xu2019transfer,zun2019modular,NEURIPS2024_d4857f72}. 
However, these approaches are either limited to single-agent tasks or fail to scale effectively in MARL settings due to the composition of MDPs.
The extensions to multi-agent systems ~\cite{jothimurugan2022specification,elsayedaly2022logicbasedrewardshapingmultiagent,liu2024guidingmultiagentmultitaskreinforcement,leon2020extendedmarkovgameslearn,hammond2021multiagentreinforcementlearningtemporal} either rely on centralized constructions, employ enumerate-and-verify strategies, or support only co-safety properties.
Several work integrate temporal logic constraints into POMDP planning, including finite-state controllers and barrier-based methods for LTL and DTL specifications~\cite{sharan2014finite,ahmadi2020barrier,wang2021reinforcement,liu2021leveraging}, but typically focus on single-agent or small settings.
Furthermore, Inverse RL~\cite{IRL11} recovers rewards from expert demonstrations, whereas \method uses only a hyperproperty specification and assumes no expert behavior.

%

%% file: prelim.tex
\vspace{-2mm}
\section{Preliminaries}

\newcommand{\explain}[1]{\textcolor{blue}{#1}}

Let $\Delta(X) \triangleq \left\{ p : X \mapsto [0,1] \;\middle|\; \sum_{x \in X} p(x) = 1 \right\}$ 
be
the probability simplex of a set $X$ of random variables.
We abbreviate tuples $\tupleof{x_1, \ldots, x_n}$ as $\tupleof{x_i}_{i \in \setof{1, \ldots, n}}$.

\subsection{Partially Observable Markov decision Processes}
\begin{definition}
A \emph{partially observable Markov decision process (POMDP)} is a tuple $\pomdp = \tupleof{\states,\init, \actions, \tran, \rew, \ObsFun,\ObsSym, \AP, \labels, \discount}$, where 
$\states$ is a finite set of {\em states}, $\init$ is {\em initial} distribution over $\states$, i.e., $\init \in \Delta(\states)$, $\actions$ is a finite {\em action} space, the transition function $\tran(\state' \mid \state, \action)$ specifies the conditional probability of transitioning to $s' \in \states$ after taking action $\action \in \actions$ in $s \in \states$.
By taking $\action$ in $\state$, the agent receives a scalar {\em reward} $\rew(\state, \action)$ and $\discount \in (0,1]$ is discount factor.
An {\em observation} $\obs \in \ObsSym$ is produced by observation probability function
$\ObsFun(\obs \mid s, \action)$ given the current $\state$ and $\action$.
$\AP$ denotes a finite set of {\em atomic propositions}, and $\labels: \states \rightarrow 2^\AP$ is a {\em labeling} function.
\qed
\end{definition}

\begin{figure}[t!]
    \centering
    \includegraphics[width=0.92\linewidth]{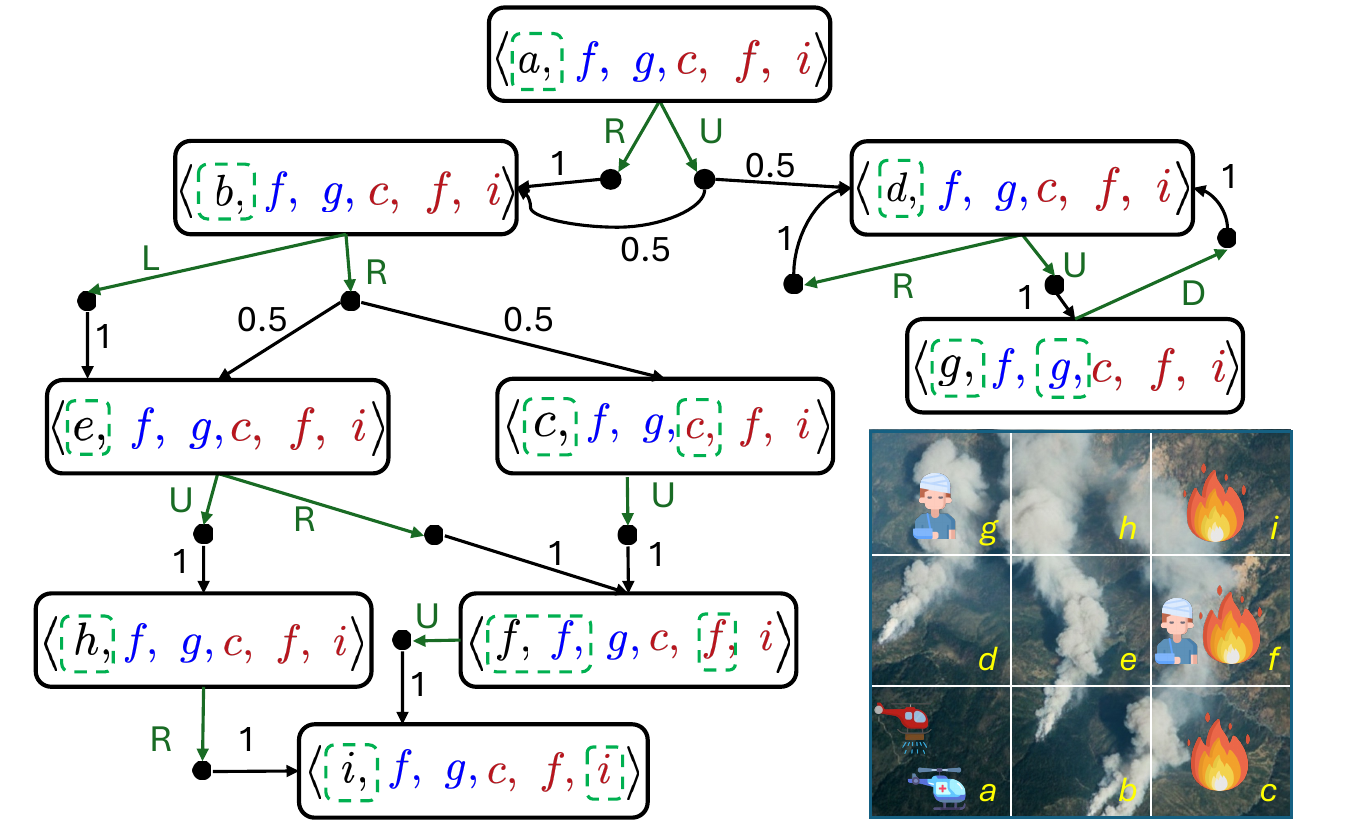}
    \captionof{figure}{WildFire scenario and its POMDP.}
    \label{fig:pomdp}
    \vspace{-4mm}
\end{figure}

\vspace{-2mm}
\paragraph{Example.}
Consider the environment in~\Cref{fig:pomdp}, a WildFire scenario on a $3\times 3$ grid world with cells labeled $\{{\locmark a}, {\locmark b}, \ldots, {\locmark i}\}$.
A firefighter drone ($\ff$ \raisebox{-0.15em}{\includegraphics[height=1.2em]{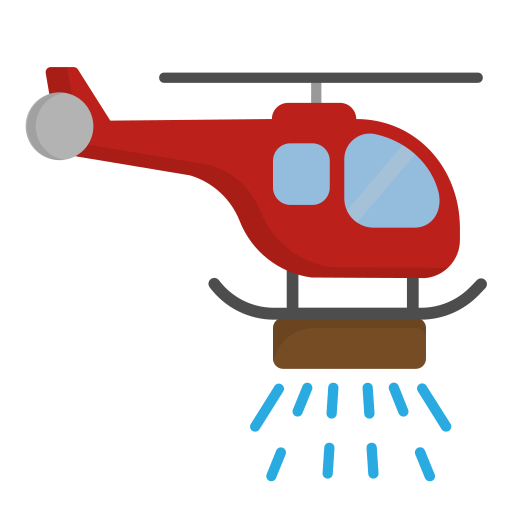}}) is tasked to extinguish three fire zones \raisebox{-0.15em}{\includegraphics[height=1.2em]{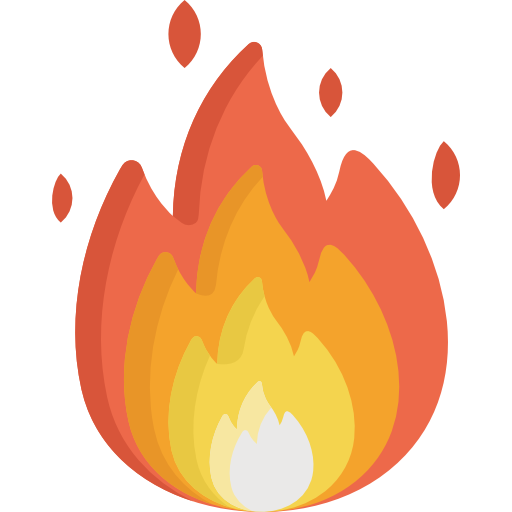}}, and a medical drone ($\med$ \raisebox{-0.15em}{\includegraphics[height=1.2em]{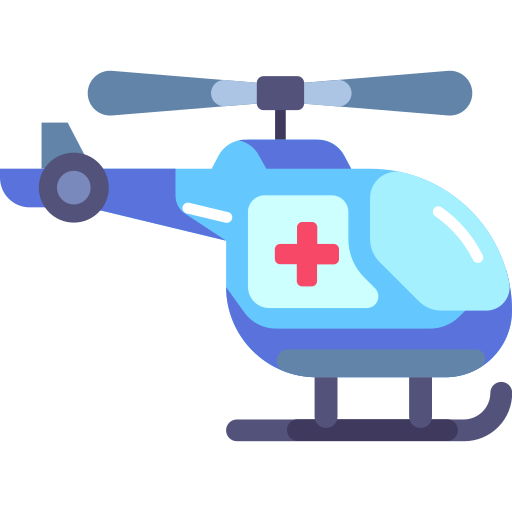}}) whose objective is to rescue two victims \raisebox{-0.15em}{\includegraphics[height=1.2em]{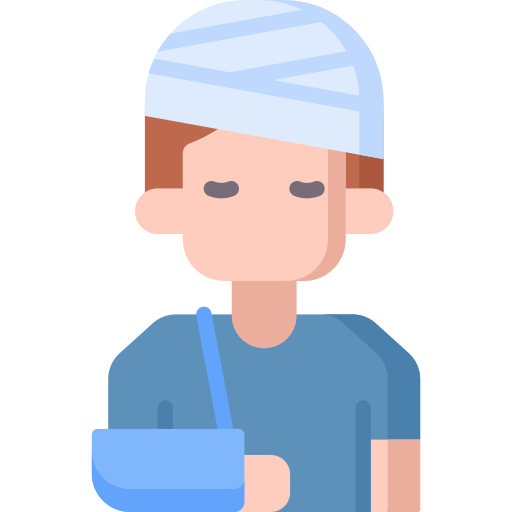}}.
Agents can communicate only within a distance of three.
$\med$ is not allowed to enter fire-zone.
The map is partially observable due to smoke and the agents can detect fire or victims only when they are in the same cell. 
The underlying state consists of the agents' positions and the locations of the two victims and three fire zones. 
\Cref{fig:pomdp} illustrates a POMDP, with state
$\tupleof{\texttt{pos}, \textcolor{blue}{\texttt{V1}, \texttt{V2}}, \textcolor{red}{\texttt{F1}, \texttt{F2}, \texttt{F3}}}$
(agent position, victim positions, and fire positions). 
Observations are partial; e.g., in state $\tupleof{a,\textcolor{blue}{f,g,}\textcolor{red}{c,f,i}}$ the agent's observation is \tikz[baseline=(X.base)] 
\node[
  draw=mygreen,
  dashed,
  inner sep=0.5pt,
  rectangle
] (X) {$\tupleof{a}$};.

A \emph{history} is an observation-action sequence $h \triangleq \obs_0 \edgeof{\action_0} \obs_1 \edgeof{a_1} \cdots $, where for all $i \geq 0$, $\obs_i \in \ObsSym$ and $a_i \in A$.
A {\em sub-history} $h_{\bracketof{\ell:k}}$ is a finite segment $\obs_l \edgeof{\action_l} \cdots \edgeof{a_{k-1}} \obs_k$, for $0 \leq \ell < k < \sizeof{h}$. 
We denote the set of all finite histories by $\actobshist^\finite$.
A \emph{policy} $\pi : \actobshist^\finite \rightarrow \actions$ maps a history to a fixed action.
A {\em path} in $\pomdp$ is a sequence of states $\traj \triangleq \state_0 \edgeof{\action_0} \state_1 \edgeof{\action_1} \cdots$, where $\state_i \in \states$ and $\action_i \in \actions$ for all $i \geq 0$. 
A {\em sub-path} $\traj_{\bracketof{\ell:k}}$ is a finite segment $\state_\ell \xrightarrow{\action_\ell} \cdots \xrightarrow{\action_{k-1}} \state_k$, for $0 \leq \ell < k < \sizeof{\traj}$. 
Given a path $\traj$, the {\em trace} $\trace$ of $\traj$ is a sequence of labels 
$\traceof{\traj} \triangleq \trace(0)\trace(1)\cdots$, such that $\trace(i) = 
\labels(\state_i)$ for all $i \geq 0$. 
We write $\trajs^\finite$ to denote the set of all finite paths, and
$\tracesof{\trajs^\finite}$ to denote the set of all finite traces of $\pomdp$.

\begin{figure*}[t!]
\begin{problemstatement}
Given a POMDP $\pomdp$ 
and a \HyperLTL formula $\varphi =
\quant_1 \tracevar_1. \ldots 
\quant_n \tracevar_n.~ 
\psi
$,
our goal is to identify a tuple of $n$ policies
$ \tupleof{\maxpolicy_1, \ldots, \maxpolicy_n}$, such that:
\vspace*{-4mm}
\begin{align*}
    \tupleof{&\maxpolicy_i}_{i \in 
        \{1, \ldots,n\}}
    \in
    \bigg[
    \underset{
        \tupleof{
            {
                \policy_i
            }
        }
    }
    {{\arg\max}}
    \notag ~~ {\prob} 
    \Big[ 
   \tupleof{ \tracesoff{ \hspace{-3mm}\bigcup_{h \in (\actobshist_i \sim \distributionof{\policy_i})}
\hspace{-3mm} \{ \arg\max_{\traj \in \trajs^*}
\maptostate(h, \traj) \} }}
    \models 
    \varphi
    \Big]
    \bigg]_{i \in \{1, \ldots,n\}}
     \\[-6mm]
\end{align*}
where $\distributionof{\policy_1},\ldots,\distributionof{\policy_n}$ 
are the distributions over a set of histories  
drawn by policies $\policy_1,\ldots, \policy_n$, 
and $\bigcup_{h \in (\actobshist_i \sim \distributionof{\policy_i})} \{ \arg\max_{\traj \in \trajs^*} \maptostate(h, \traj) \}$ is a set of paths $\trajs_{i}$ associated with policy $\policy_i$.
\end{problemstatement}
\vspace{-5mm}
\caption{Formal problem statement of \method.}
\vspace{-5mm}
\label{fig:hypole}
\end{figure*}

\vspace{-2mm}
\subsection{Finite Semantics for HyperLTL}
\label{sec:hyperltl}

\HyperLTL syntax~\cite{cfkmrs14} is as follows:
%
{\small
\begin{align*}
\varphi  ::= \exists \tracevar . \varphi \mid \forall \tracevar. \varphi \mid \psi \quad ~
\psi  ::= \var_\tracevar \mid \lnot \psi \mid \psi \OR \psi \mid \X \psi \mid \psi \until \psi,
\end{align*}}
\hspace*{-1mm}where $\var \in \AP$ is an atomic proposition and $\tracevar$ is a trace variable. 
The second rule yields standard LTL formulas, with Boolean connectives $\lnot$, $\OR$, and temporal operators $\X$ (`next') and $\until$ (`until').
Other Boolean and temporal operators are 
defined as 
syntactic sugar: $\tru \triangleq \var_\tracevar \vee \neg \var_\tracevar$, $\fals \triangleq \neg \tru$, $\psi_1 \rightarrow \psi_2 \triangleq \neg \psi_1 \vee \psi_2$, $\F \psi \triangleq \tru \until \psi$, and $\always \psi \triangleq \neg \F \neg \psi$, where $\F$ and $\always$ denote the `\emph{eventually}' and `\emph{always}'
, respectively.
In $\varphi$, $\exists \tracevar$ means ``along some trace $\tracevar$'', and $\forall \tracevar$ means ``along all traces $\tracevar$''. 
We write $\traceVars$ for the set of trace variables appearing in $\varphi$. 
A formula is \emph{closed} if all $\tracevar \in \traceVars$ are quantified exactly once.
%

Since our setting is episodic and each episode terminates at a finite time, we adopt the finite-trace semantics of \HyperLTL~\cite{bsb17}. 
%
I.e., formulas are interpreted over finite trace assignments, given by a partial mapping 
$\Pi : \traceVars \rightharpoonup (2^\AP)^\finite$ 
that assigns each $\tracevar \in \traceVars$ to a finite trace.
Given a trace assignment $\Pi$, a trace variable $\tracevar$, and a finite trace $\trace \in (2^\AP)^\finite$, we write $\Pi[\tracevar \mapsto \trace]$ for the assignment that agrees with $\Pi$ except that $\tracevar$ is mapped to $\trace$. 
We denote by $\Pi_\emptyset$ the empty trace assignment, and $\trace \in \Pi$ to refer to traces in the image of $\Pi$.

Given a closed \HyperLTL formula $\varphi \definedas \quant_1 \tracevar_1 \ldots \quant_n \tracevar_n.~ \psi$, where each $\quant_i \in \{\forall, \exists\}$, 
and $\psi$ is an inner \LTL formula, an \emph{interpretation} is a tuple $\Tr \triangleq\tupleof{\tr_{\tracevar_i}}_{i\in\setof{1,\ldots,\sizeof{\traceVarsOf{\varphi}}}}$, where $\tr_{\tracevar_i} = \tracesof{\trajs_i}$ is a set of traces that may be assigned to the corresponding trace variable $\tracevar_i$. 
The satisfaction relation $\models$ maps a formula $\varphi$ to a model $(\Tr, \Pi, i)$, where $i \in \integers_{\geq 0}$ indicates the current evaluation position.
Formally: 
\vspace{-2mm}
{\small
\[
\begingroup
\renewcommand{\arraystretch}{1.2}
{
\setlength{\arraycolsep}{3pt}
\begin{array}{@{}ll@{\hspace{.5em}}l@{\hspace{1em}}l@{}}

(\Tr, \Pi,0) & \models \exists \tracevar.\ \modified{\psi} & \text{iff} &
\exists t \in \tr_{\tracevar}.(\Tr,\Pi[\tracevar \rightarrow t],0) \models \psi\\

(\Tr, \Pi, 0) & \models \forall \tracevar.\ \modified{\psi} & \text{iff} &
\forall t \in \tr_{\tracevar}. (\Tr,\Pi[\tracevar\rightarrow t],0) \models \psi\\

(\Tr, \Pi, i) & \models \var_\tracevar & \text{iff} &
\var \in \Pi(\tracevar)(i)\\

(\Tr, \Pi,i) &\models \neg \psi & \text{iff} &
(\Tr, \Pi, i) \not\models \psi\\

(\Tr, \Pi,i) &\models \psi_1 \OR \psi_2 & \text{iff} &
(\Tr, \Pi,i) \models \psi_1 \text{or }(\Tr, \Pi,i) \models \psi_2\\

(\Tr, \Pi,i) &\models \X \psi & \text{iff} &
(\Tr,\Pi,i+1)\models\psi\\
&&& \text{and } \forall \trace \in \Pi.\ |\trace| > i+1\\

(\Tr, \Pi, i) &\models \psi_1 \until \psi_2 & \text{iff} &
\exists j \geq i \text{ with } j < \min_{\trace \in \Pi} |\trace|,\\
&&& \text{s.t. } (\Tr,\Pi, j) \models \psi_2\\
&&& \text{and } \forall k \in [i, j), (\Tr,\Pi, k)\models \psi_1
\end{array}
}
\endgroup
\]}
We say that an interpretation $\Tr$ satisfies a \HyperLTL formula $\varphi$, written as $\Tr \models 
\varphi$, if $(\Tr, \Pi_\emptyset, 0) \models \varphi$.

\vspace{-2mm}
\paragraph{Example.} Consider the following arbitrary \HyperLTL formula for the WildFire scenario:
\vspace{-2mm}
{\small
\[
\varphi_{\text{exp}} \definedas \forall \tracevar_1 \exists \tracevar_2 . ~ 
\textsf{Dist}(\texttt{pos}_{\tracevar_1}, \texttt{pos}_{\tracevar_2})<3 ~~ \U ~~ 
(\texttt{pos}_{\tracevar_1} = i )
\]}
\hspace*{-1mm}Here, $\ff$ corresponds to $\tracevar_1$ and $\med$ corresponds to $\tracevar_2$. The predicate $\textsf{Dist}(\texttt{pos}_{\tracevar_1}, \texttt{pos}_{\tracevar_2}) < 3$ enforces that they remain within distance $3$, (until \ff reaches cell $i$). 
Importantly, $\forall\exists$ captures inter-agent dependency by letting $\med$'s decisions depend on $\ff$'s, whereas most specification-RL work focuses on $\forall\forall$, which cannot express this dependency.

%% file: problem.tex
\vspace{-2mm}
\section{Problem Statement}
\label{sec:problem}

To bridge POMDP histories to \HyperLTL interpretation, we define a function
$
\maptostate : \actobshist^\finite \times \trajs^\finite \rightarrow [0,1]
$,
which maps an action-observation history and a path to a scalar value in $[0,1]$. 
We say that $\maptostate$ is \emph{path-consistent}, meaning that for any history $h$ generated by a latent path $\traj$,
$
\traj \in \arg\max_{\traj' \in \trajs^\finite} \maptostate(h,\traj').$
For a \HyperLTL formula $\varphi \definedas \quant_1 \tracevar_1 \ldots \quant_n \tracevar_n.~ \psi$, where each $\quant_i \in \{\forall, \exists\}$, let $\distributionof{\policy_i}$ denote the distribution over observation–action histories induced by a policy $\policy_i$ and let $\actobshist_i \sim \distributionof{\policy_i}$ denote a {\em set} of histories sampled from this distribution, for $1 \leq i \leq n$ and $\actobshist_i \subseteq \actobshist^*$. 
The policy $\policy_i$ is associated with trace variable $\tracevar_i$, for each $i \in \setof{1,\ldots,|\traceVarsOf{\varphi}|}$, so that each trace variable ranges over the possible behaviors generated by its corresponding policy.
The mapping $\maptostate$ can be constructed by any technique such as belief-states constructions~\cite{poupart2013valuedirectedbeliefstateapproximation,NIPS1999_158fc2dd}.
For each trace variable $\tracevar_i \in \traceVarsOf{\varphi}$, we introduce the set $\trajs_i$ of paths, where each path $\traj$ in $\trajs_i$ has the highest probability for each $h \in \actobshist_i$ drawn from distribution $\distributionof{\pi_i}$ using policy $\pi_i$: 
{\small
\[
\trajs_i
\definedas \hspace{-3mm}
\bigcup_{h \in (\actobshist_i \sim \distributionof{\policy_i})}
\hspace{-3mm} \big\{ \arg\max_{\traj \in \trajs^*}
\maptostate(h, \traj) \big\}
\]}
\hspace*{-1.5mm}Then, the \emph{family} of paths for each set $\trajs_i$ of paths is defined as:
$
\familyofsamples \definedas \tupleof{\trajs_i}_{i\in\setof{1,\ldots,\sizeof{\traceVarsOf{\varphi}}}}. 
$ 
We write, $\Tr = \tracesof{\familyofsamples}$ as the tuple of sets of sampled traces. 
Also, we say a family of paths $\familyofsamples$ (induced by each 
$\policy_\tau$ associated with each $\tracevar \in \traceVars$), satisfies a 
formula $\varphi$ if $\tupleof{\tracesof{\trajs_i}}_{i\in\setof{1,\ldots,\sizeof{\traceVarsOf{\varphi}}}} \models \varphi$.
The formal problem statement is shown in~\cref{fig:hypole}, where `$\star$' denotes optimality (e.g., $\maxpolicy$ denotes an optimal policy). 
That is, the tuple of policies $\tupleof{\maxpolicy_1, \ldots, \maxpolicy_n}$ 
maximizes the probability $\prob$ 
such that the generated tuple of sets of traces from $\pomdp$ satisfies $\varphi$. 

\vspace{-2mm}
\paragraph{Example.} Consider the POMDP in \cref{fig:pomdp} and 
$\varphi_{\text{exp}}$.
Suppose $\ff$ samples histories with policy $\policy_{1}$, $\actobshist_{\tracevar_1}= \{ 
h_{1}^1, h_1^2 \}$:
\vspace{-3mm}
{\small
\begin{align*}
 h_1^1= & \tupleof{a}\overset{\text{R}}{\rightarrow}\tupleof{b} 
\overset{\text{R}}{\rightarrow}
\tupleof{c, \textcolor{red}{c}}\overset{\text{U}}{\rightarrow} 
\tupleof{f, \textcolor{blue}{f} , \textcolor{red}{f}}\overset{\text{U}}{\rightarrow} \tupleof{i, \textcolor{red}{i}} 
\notag \\
    h_1^2= & 
\tupleof{a}\overset{\text{R}}{\rightarrow}
\tupleof{b}\overset{\text{R}}{\rightarrow}
\tupleof{e}\overset{\text{U}}{\rightarrow}
\tupleof{h}\overset{\text{R}}{\rightarrow} \tupleof{i, \textcolor{red}{i}} \\[-5.5mm] 
\end{align*}}
Likewise, agent $\med$, corresponding to $\tracevar_2$, samples histories using policy 
$\policy_{2}$, $\actobshist_{\tracevar_2}= \{ h_{2}^1, h_2^2 \}$:
\vspace{-1mm}
{\small
\begin{align*}
 h_2^1 = & \tupleof{a}\overset{\text{U}}{\rightarrow} 
\tupleof{d}\overset{\text{U}}{\rightarrow}
\tupleof{g, \textcolor{blue}{g}}\overset{\text{D}}{\rightarrow}
\tupleof{d}\overset{\text{R}}{\rightarrow}\tupleof{d} \notag ~~~~~\\ 
    h_2^2 = &
\tupleof{a}\overset{\text{U}}{\rightarrow}
\tupleof{b}\overset{\text{L}}{\rightarrow} \tupleof{e} 
\overset{\text{R}}{\rightarrow} \tupleof{f, \textcolor{blue}{f} , \textcolor{red}{f}}
\overset{\text{U}}{\rightarrow}\tupleof{i, \textcolor{red}{i}} \\[-6mm]
\end{align*}}
%
%
The set of paths associated with $\actobshist_{\tracevar_i}$ is
$\trajs_{\tracevar_i}=\{\arg\max_{\traj\in\trajs^*}\maptostate(h_i^1,\traj),\,\arg\max_{\traj\in\trajs^*}\maptostate(h_i^2,\traj)\}$,
 for $i \in \{1, 2\}$.
We now compute the probability of satisfying $\varphi_{\text{exp}}$ using $\trajs_{\tracevar_1}$ and $\trajs_{\tracevar_2}$ as follows, for $j\in \{1,2\}$:

{\small
\begin{align*}
\tracesof{\tupleof{ \{\arg\max_{\traj\in\trajs^*}\maptostate(h_1^j,\traj)\},  \trajs_{\tracevar_2}}} \models \varphi_{\text{exp}}
\end{align*}}
\hspace*{-1mm}Here, $\arg\max_{\traj\in\trajs^*}\maptostate(h_2^2,\traj)$ serves as a witness for $\tracevar_2$ in both satisfaction relations; hence the satisfaction probability of $\varphi_{\text{exp}}$ under $\actobshist_{\tracevar_1}$ and $\actobshist_{\tracevar_2}$ is $1$. 
However, if we replace the quantifiers in $\varphi_{\text{exp}}$ with $\forall\forall$, the satisfaction 
probability drops to $0.75$ (check~\Cref{appendix:example} for details).

%% file: algorithm.tex
\vspace{-3mm}
\section{Algorithmic Details of \method} 
\label{sec:algorithm}

Our solution to the problem in~\Cref{fig:hypole} has three steps.
First, we apply Skolemization~\cite{skolem1920} to eliminate quantifier alternation (\Cref{sec:skolemization}).
Next, we define robustness functions to quantify and optimize satisfaction of the \HyperLTL formula (\Cref{sec:quantsemantics}).
Finally, we lift the shared-environment POMDP to a Decentralized POMDP and apply CTDE algorithms to learn policies to solve the problem in~\Cref{fig:hypole}.

\begin{figure*}[t!]
\begin{bluebox}
\vspace{-3mm}
{\footnotesize   
\begin{align}
\tupleof{ \maxpolicy_j}_j
\in &
\bigg[
\underset{
\tupleof{
    {
        {\policy_{{j}}}
    }
}
}
{{\arg\max}} ~\prob 
\Big[ 
\tupleof{ 
\imageof{\skolemfunc_i}
}
\orderedsets	
\tupleof{ 
\tracesof{\hspace{-5mm}\bigcup_{h \in (\actobshist_{i_j} \sim \distributionof{\policy_{i_j}})}
\{ \arg\max_{\traj \in \trajs^*} \maptostate(h, \traj) \} }
}
\models
\skolemized{\varphi}
\Big]
\bigg]_{{i \in \existsof{}}, {j \in \forallsof{}}} 
\label{eq:problem_statement_2}
\\[-5pt]
\tupleof{\maxpolicy_\ell}_{\ell}
\in &
\bigg[ 
\underset{
\tupleof{
\policy_{{\ell}}
}
}
{{\arg\max}}~
{\prob}
\Big[ 
\rho \big(
\zip 
\tupleof{
\traceof{(\arg\max_{\traj \in \trajs^*} \maptostate(h_{\ell} \sim \distributionof{\policy_{\ell}} , \traj))_{[0:k_l]}}
}
\big), 
\psi \big)
\converge 
\rbvalue_{\mathit{max}}
\Big]
\bigg]_{\ell \in \setof{1,...n}}
\label{eq:robustzip}
\\[-5pt]
\tupleof{\maxpolicy_i}
\orderedunion
\tupleof{\maxpolicy_j}
\in &
\Bigg[
\underset{
	\tupleof{\policy_i}
	\orderedunion
	\tupleof{\policy_j}
}
{{\arg\max}} 
~{\prob}
\Big[
\rbvalue
\Big(
\zip
\big(
\tupleof{\traceoff{(\arg\max_{\traj \in \trajs^*} \maptostate(h_i \sim \distributionof{\policy_{i}} , 
\traj))_{[0:k_i]}}\big} \;
\orderedunion \, 
\label{eq:opt}
\\[-16pt] 
& ~~~~~~~~~~~~~~~~~~~~~~~~~~~~~~~~~~
\tupleof{\traceoff{(\arg\max_{\traj \in \trajs^*} \maptostate(h_j \sim \distributionof{\policy_{j}} , \traj))_{[0:k_j]}}\big}
\big),
\skolemized{\psi}
\Big)
\converge 
\rbvalue_{\mathit{max}}
\Big]\Bigg]_{{i \in \existsof{}}, {j \in \forallsof{}}} \notag
\end{align}
}
\end{bluebox}
\vspace*{-5mm}
\label{fig:allformulas}
\vspace*{-4mm}
\end{figure*}

\vspace{-3mm}
\subsection{Step 1: HyperLTL Skolemization}
\label{sec:skolemization}
Let
$
\varphi =
\quant_1 \tracevar_1 \ldots \quant_n \tracevar_n.~ 
\psi(\tracevar_1,\ldots,\tracevar_n)
$
be a \HyperLTL formula.
Following~\cite{hsu2025hyprl}, we Skolemize $\varphi$ to eliminate quantifier alternation. 
Let $\existsof{}=\{i \mid \quant_i=\exists\}$ and $\forallsof{}=\{j \mid \quant_j=\forall\}$.
For each $i\in\existsof{}$, let $\forallsof{i}=\{\,j\in\forallsof{}\mid j<i\,\}$ denote the set of indices of all \emph{preceding} universal quantifiers.
For each $i\in\existsof{}$, a Skolem function $\skolemfunc_i:\Tr^{|\forallsof{i}|}\rightarrow\Tr$ is defined, reducing to a constant function when $\forallsof{i}=\emptyset$.
A trace assignment $\traceassignment$ is consistent with $\skolemfunc_i$ if, for all $i\in\existsof{}$, 
$\traceassignment(\tracevar_j)\in\Tr$ for all $j\in\forallsof{i}$ and
$
\traceassignment(\tracevar_i)=
\skolemfunc_i\big(
\traceassignment(\tracevar_{i_1}), \traceassignment(\tracevar_{i_2}),\ldots,
\traceassignment(\tracevar_{i_{|\forallsof{i}|}})
\big)
$
for all $i \in \existsof{}$, where $\forallsof{i}=\{i_1<\cdots<i_{|\forallsof{i}|}\}$.
If $(\Tr,\traceassignment,0)\models\varphi$ holds for all trace assignments consistent with all $\skolemfunc_i$, then each $\skolemfunc_i$ witnesses the satisfaction of $\varphi$~\cite{wz24}.
For the inner \LTL formula $\psi$, we obtain $\skolemized{\psi}$ by substituting each proposition $p_{\tau_i}$ with $p_{\skolemfunc_i}$ for all $p \in \AP$ and $i \in \existsof{}$, thereby instantiating existential traces with their Skolem witnesses.x
The Skolemized formula is:
\vspace{-2mm}
{\small
\begin{align*}
\skolemized{\varphi} = 
\underbrace{\exists \skolemfunc_{i}(\tau_{i_1}, \ldots, \tau_{i_{|\forallsof{i}|}})}_{\text{for each } i \in 
\existsof{}}.
\underbrace{\spvertund{1.5ex}\forall \tracevar_{j}.}_{\text{for each }j \in \forallsof{}}\skolemized{\psi} \\[-7mm]
\end{align*}
}
\hspace*{-1mm}Based on this transformation, we rewrite the problem statement in~\cref{fig:hypole} of~\Cref{sec:problem}.  
The updated statement is shown in~\cref{eq:problem_statement_2}, where $\imageof{\skolemfunc_i}$ denotes the set of traces obtained by applying the Skolem function to the preceding universally quantified traces (as detailed in~\Cref{appnedix:image}).

\vspace{-2mm}
\subsection{Step 2: Learning with Quantitative Semantics}
\label{sec:quantsemantics}

We reformulate the \HyperLTL satisfaction problem as an optimization problem under its quantitative semantics, where the Skolemized formula $\skolemized{\varphi}$ is evaluated on tuples of histories $\tupleof{h_1,\ldots,h_n}$ sampled from the POMDP $\pomdp$.
Let $\mathbb{R}$ be the set of real numbers, $\Psi$ the set of all \LTL formulas, and $\predicatefunc : 2^{\AP} \rightarrow \mathbb{R}$ be a \emph{valuation function} assigning real values to sets of atomic propositions.
For a state $\state \in \states$ of $\pomdp$, quantitative semantics are defined by predicates of the form $\predicatefuncof{\labelsof{\state}} < \constant$, where $\constant$ is a user-specified threshold (see~\Cref{fig:quantsemantics} in Appendix for full semantics).
The robustness function $\rbvalue:\tracesof{\trajs^\finite}\times\Psi \rightarrow \mathbb{R}$ assigns a real-valued score to a finite trace and an \LTL formula, bounded by $\rbvalue_{\mathit{max}}$ and $\rbvalue_{\mathit{min}}$. 
Formally, given an \LTL formula $\psi$ and a POMDP $\pomdp$, we formulate the policy synthesis problem as:
\vspace{-1mm}
{\small
\[
    \maxpolicy \hspace{-0.3mm} \in \hspace{-0.5mm}
\underset{
	\policy
}
{{\arg\max}} \hspace{-1mm}
\underset{h \sim \distributionof{\policy}}
{\prob}
\Big[ \notag 
\rbvalueof{\traceoff{(\arg\max_{\traj \in \trajs^*} \maptostate(h, \traj))_{[0:k]}}, \psi} \converge \rbvalue_{\mathit{max}} 
\Big]
\]}
\hspace*{-1mm}where $\converge$ means convergence. 
I.e., $\maxpolicy$ maximizes the probability that paths induced by histories under $\policy$ satisfy $\psi$.
Next, to compute robustness over multiple (universally quantified) traces, we define a $\zip$ operator
as follows:
\vspace{-1mm}
{\small
\[
\zipof{\tupleof{\trace_\ell}_{\ell\in\setof{1,\ldots,n}}}
\definedas
\tupleof{\trace_\ell(0)}_{\ell\in\setof{1,\ldots,n}} \cdots
\tupleof{\trace_\ell(k)}_{\ell\in\setof{1,\ldots,n}}
\]
}
\hspace*{-1mm}where
$ k \definedas \min_{\ell\in\setof{1,\ldots,n}} m_\ell.
$
E.g., given $\trace_1 = a_1a_2a_3$ and $\trace_2 = b_1b_2b_3$, their zipped trace is
$
\zipof{\tupleof{\trace_1,\trace_2}}
=
\tupleof{a_1,b_1}\,\tupleof{a_2,b_2}\,\tupleof{a_3,b_3}.
$
Thus, the optimization problem of computing a tuple of policies
$\tupleof{\maxpolicy_1, \ldots, \maxpolicy_n}$
that maximizes robustness can be stated as~\cref{eq:robustzip}.

Now, let $\tupleof{\trace_i}_{i\in I}$ and $\tupleof{\trace_j}_{j\in J}$ be two trace tuples, where
$I \cup J = \setof{1,\ldots,n}$ and $I \cap J = \emptyset$.
To preserve index order when combining them, we use $\orderedunion$, which produces a single tuple ordered by path indices.
Based on this construction, satisfaction of the inner \LTL body $\psi$ is defined over tuples of histories.
For each $i\in\existsof{}$ and $j\in\forallsof{}$, a tuple $\tupleof{h_1,\ldots,h_n}$ satisfies $\psi$ if and only if, after combining the paths associated with histories via ordered union and zipping them into a joint trace, the resulting robustness score converges to $\rbvalue_{\mathit{max}}$ (details in~\cref{appendix:quant}).
Accordingly, we formulate the optimization problem in~\eqref{eq:opt}.

Observe~\cref{eq:opt}, the robustness value $\rbvalue$ for an $\exists$-quantified trace instantiated by a Skolem function depends on whether the trace induced by $h_i \sim \distributionof{\policy_i}$ can serve as a valid witness for the preceding $\forall$-quantified traces.
Concretely, the robustness of a Skolem witness $\skolemfunc_i$ with respect to $\psi$ is defined as
$
\rbvalueof{\skolemfunc_i,\psi}
\definedas
\rbvalueof{\skolemfunc_i\!\big(\traceof{\traj_{i_1}},\ldots,\traceof{\traj_{i_{|\forallsof{i}|}}}\big),\psi}.
$
As a result, the optimization of $\tupleof{\maxpolicy_i}$ is inherently coupled with $\tupleof{\maxpolicy_j}$, which is crucial for correctly capturing the semantics of \HyperLTL formulas with quantifier alternation.

\begin{theoremstatement}{}\label{theorem:skolem}
Given a POMDP $\pomdp$ and a \HyperLTL formula $\varphi$, 
if $\maptostate$ is path-consistent, then any tuple of policies
$\tupleof{\maxpolicy_i}_{i\in\existsof{}} \orderedunion \tupleof{\maxpolicy_j}_{j\in\forallsof{}}$
that optimizes the Skolemized formula $\skolemized{\varphi}$
also optimizes the probability of satisfying $\varphi$ in $\pomdp$ as defined in~\Cref{fig:hypole}.
\end{theoremstatement}

\vspace{-3mm}
\subsection{Step 3: MARL for \HyperLTL}
\label{sec:step}
\vspace{-1mm}

Now, we aim to synthesize 
policies for the universally quantified path variables and to learn Skolem 
functions that witness the existential quantifiers in the Skolemized 
\HyperLTL formula.
To this end, we adopt value-based CTDE paradigm, i.e., 
VDN~\cite{sunehag2017valuedecompositionnetworkscooperativemultiagent},
QMIX~\cite{pmlr-v80-rashid18a}, and 
QTRAN~\cite{son2019qtranlearningfactorizetransformation}.
In CTDE, training leverages global information, such as other 
agents’ observations or the global state, while execution is decentralized 
and each agent acts based on its local observation.

\vspace{-1mm}
Our objective is to compute tuples of optimal policies
$\tupleof{\maxpolicy_i}_{i \in \existsof{}}$ and
$\tupleof{\maxpolicy_j}_{j \in \forallsof{}}$
that solve~\eqref{eq:opt}.
For each $j \in \forallsof{}$, we construct an optimal policy
$\maxpolicy_j(h_{j[0:k]})$ that depends solely on $h_j$.
For each $i \in \existsof{}$, the corresponding policy takes a history
induced by the associated Skolem witness as an input. To formalize this, we introduce a
trace-to-history consistency map
$\traceThist : \Tr \to 2^{\actobshist}$, where
$
\traceThist(\trace)
\definedas
\{
h \in \actobshist
\mid
\traceof{\arg\max_{\traj \in \trajs^*} \maptostate(h,\traj)} = \trace
\}.
$
That is, $\traceThist(\trace)$ denotes the set of histories whose most likely
latent path under $\maptostate$ induces trace $\trace$.
We assume that 
$\traceThist(\trace)\neq\emptyset$ for every
$\trace\in\Tr$ under consideration.
Since policies require a single history as input, we fix a function $\select$, 
that maps a nonempty set $X$ to an arbitrary member of $X$, and introduce
$\traceToHist : \Tr \to \actobshist$, and defined as,
$
\traceToHist(\trace) \definedas \select\big(\traceThist(\trace)\big).
$
Finally, for each $i \in \existsof{}$, we construct an optimal policy that takes
a history induced by the corresponding Skolem witness as an input, namely,
$\maxpolicy_i (
\traceToHist(
\skolemfunc_i(
\traceof{\trajfromto{\traj_{i_1}}{0}{k}}, \ldots,
\traceof{\traj_{i_{\sizeof{\forallsof{i}} \bracketof{0:k}}}}
)
)
).$
This construction implies that the decisions of the optimal policies associated with existential quantifiers depend on the optimal policies of the preceding universal quantifiers for capturing the inter-agent dependencies.

\vspace{-3mm}
\paragraph{From POMDP to Dec-POMDP.}
Value-based CTDE algorithms are defined over Decentralized 
POMDPs
(Dec-POMDP)~\cite{oliehoek2016concise}, represented as a tuple 
$\decpomdp = \langle \statesdec,\initdec, \actions, \trandec, \rewdec,   
\ObsFun, \obsdec, n, \AP, \labels, \discount \rangle$.
It is important to note that a POMDP is a special case of a Dec-POMDP~\cite{LMPJPJ20}, and our framework simultaneously samples all the histories associated with each quantifier from the POMDP.
To make the shared POMDP compatible with CTDE algorithms, we formally 
lift the shared POMDP $\pomdp$
for $n$ agents into a Dec-POMDP. 
In this transformation, the state space remains the same, i.e., $\statesdec \definedas 
\states$ and $\initdec\in\Delta(\statesdec)$.
The joint action space is defined as the Cartesian product of the individual 
agent action spaces, i.e., $\actionsdec \definedas \actions^n$ (we assume the same 
action space for all agents), and a joint action is given by \modif{$\actiondec 
\definedas \tupleof{\action_\ell}_{\ell \in \setof{1,\ldots,n}}$}.
The transition function $\trandec(\statedec' \mid \statedec, \actiondec)$ specifies the conditional probability of transitioning to state $\statedec' \in \statesdec$ after taking joint action $\actiondec$ in $\statedec \in \statesdec$.
When taking $\actiondec$ in $\statedec$, the agents receive a scalar reward $\rewdec(\state_\leafindex, \actiondec_\leafindex)$.
The components $\discount$, $\AP$, and $\labels$ are defined as in the POMDP case.
The observation space in $G$ is defined as \modif{$\obsdec \definedas 
\ObsSym^n$}, 
where each agent $\ell$ receives an individual observation from $o_\ell \in \ObsSym$ based on the observation probability function $\ObsFun(\obs_\ell \mid s, \action_\ell)$.
The joint history is denoted by
$\jointhistory \definedas \tupleof{h_\ell}_{\ell\in \{1,\ldots n\}}$,
where $h_\ell$ is the local history of agent $\ell \in \{1,\ldots, n \}$.
Each $h_\ell$ is obtained using $\ObsFun$.

\vspace{-1mm}
\paragraph{Robustness Values as Reward Signals.}
Let the state of a zipped path at position $k$ as 
$s_k = \zipofk{\tupleof{\trace_\ell}_{\ell\in\{ 1,\ldots, n \}}}$.
The immediate reward signal at step $\leafindex$ is based on the scalar robustness value of the zipped trace from 
\cref{eq:opt} defined as:
{\small
\begin{align*}
\rewdec &(\state_\leafindex, \actiondec_\leafindex)
\definedas
\Big[
\rbvalue
\Big(
\zip
	\big(
	\tupleof{
	\traceof{(\arg\max_{\traj \in \trajs^*} \maptostate(h_{i}, \traj))_{[0:\leafindex+1]}}
	}
\;	\orderedunion \notag \\&
	\tupleof{
	\traceof{(\arg\max_{\traj \in \trajs^*} \maptostate(h_{j}, \traj))_{[0:\leafindex+1]}}
	}
	\big), 
	\skolemized{\psi}
\Big)
\Big]_{{i \in \existsof{}},{j \in \forallsof{}}}
\end{align*}}
\hspace*{-1mm}Since paths are defined over state–action sequences, $\actiondec_k$ is included in the zipped paths prefix to evaluate $\rho$.
Moreover, because Dec-POMDP rewards are defined over the underlying state and joint action, we are allowed to assume that $\maptostate$ is \emph{path-consistent} to compute robustness-based reward and feedback from observation histories.
This assumption is used only for reward computation during training, and is not needed to generate the optimal policies.

\vspace{-1mm}
\paragraph{MARL Optimization Problem.}
The formal optimization problem is to find a joint policy
$\jointpolicy(\jointhistory) \definedas \tupleof{\policy_\ell(h_\ell)}_{\ell \in 
\setof{1,\ldots,n}}$ that maximizes the joint value function, defined as, $\Value^{\jointpolicy}(\jointhistory_{[0:k]})
=
\expected
[
\rewdec(\state_\leafindex, \actiondec_\leafindex) + \discount  \Value^{\jointpolicy} (\jointhistory_{[0:k+1]})]$, where $\actiondec_k = \jointpolicy(\jointhistory_{[0:k]})$.

As an alternative, many RL methods work with the joint history--action 
value function (Q-function) 
$\jointqvalue^{\jointpolicy}(\jointhistory,\actiondec)$, which evaluates 
taking joint action $\actiondec$ at joint history $\jointhistory$ and 
continuing according to $\jointpolicy$, 
$\jointqvalue^{\jointpolicy}(\jointhistory_{[0:k]},\actiondec_k)
=
\expected
[
\rewdec(\state_\leafindex, \actiondec_\leafindex)
+
\discount
\expected[
\jointqvalue^{\jointpolicy}(\jointhistory_{[0:k+1]}, \jointpolicy(\jointhistory_{[0:k+1]}))
]
\mid
\actiondec_k
].$

CTDE methods learn per-agent utilities that are combined (e.g., via a mixing network) into a joint Q-function $\jointqvalue$ during centralized training.
At execution time, agents act greedily using only local utilities, enabling decentralized control.
Formally, CTDE aims to approximate
$\jointqvalue^*(\jointhistory,\actiondec)=\max_{\jointpolicy}\jointqvalue^{\jointpolicy}(\jointhistory,\actiondec)$
and induce the joint optimal policy
$\jointpolicy^\optimal(\jointhistory)\in\arg\max_{\actiondec}\jointqvalue^*(\jointhistory,\actiondec)$.

\vspace{-1.0mm}
\paragraph{Constructing Policies Using CTDE.}
From the given joint policy $
\jointpolicy \definedas \tupleof{\policy_\ell}_{\ell \in \{1,\ldots, n\}}
$, we construct two tuples of policies 
to the corresponding indices:
\(
\tupleof{\policy_i}_{i \in \existsof{}}
\)
and
\(
\tupleof{\policy_j}_{j \in \forallsof{}}
\).

In CTDE algorithms, each agent (corresponding to a trace variable in our setting) selects actions during execution based solely on its own local observation history.
Thus, for the policies associated with universally quantified traces, denoted 
by $\tupleof{\policy_j}_{j \in \forallsof{}}$, where each policy takes as input 
its local history $h_{j[0:k]}$, we can directly apply CTDE methods.
In particular, we use robustness values as reward signals to approximate $\jointqvalue$ during centralized training, enabling the iterative construction of the policies $\tupleof{\policy_j}_{j \in \forallsof{}}$.

For policies associated with existentially quantified traces,
denoted by $\tupleof{\policy_i}_{i \in \existsof{}}$,
each agent’s policy requires as input the history induced by the corresponding Skolem witness, which is produced by the Skolem function $\skolemfunc_i$.
To compute $\skolemfunc_i$, access to the paths associated with the preceding universally quantified trace variables,
namely $\traj_{i_1[0:k]}, \ldots, \traj_{i_{|\forallsof{i}|}[0:k]}$,
is required.
Such information is available during the centralized training, but not during decentralized execution; hence, during training, we construct a replay buffer $\buffer$ that stores sequences of the form
$(\obs_i, \state_{i_1}, \ldots, \state_{i_{|\forallsof{i}|}})$
for each $i \in \existsof{}$.

Next, we train
$\fseq : \actobshist \rightarrow \trajs^{|\forallsof{i}|}$ model.
During centralized training, we sample sequences from the replay buffer $\buffer$ and use them to train $\fseq$ to predict the paths
$(\hat{\traj}_{i_1[0:k]}, \ldots, \hat{\traj}_{i_{|\forallsof{i}|}[0:k]})$
given $h_{i[0:k]}$.
The predicted paths serve as inputs to the Skolem function $\skolemfunc_i$, enabling the construction of policies for existentially quantified traces; in particular, for each $i\in\existsof{}$, we have,
$\policy_i(
\traceToHist\ (
\skolemfunc_i(
\traceof{\hat{\traj}_{i_1 {[0:k]}} },
\ldots,
\traceof{\hat{\traj}_{i_{|\forallsof{i}|} {[0:k]}} }
)
)
).$

Up to now, we have described how to learn
$\tupleof{\policy_i}_{i \in \existsof{}}$ and
$\tupleof{\policy_j}_{j \in \forallsof{}}$ using CTDE.
However, it is important that CTDE algorithms are designed to approximate a joint action–value function and to induce an optimal joint policy $\jointpolicy^\optimal$.
This objective does not, in general, imply that the resulting
$\tupleof{\maxpolicy_i}_{i \in \existsof{}}
\orderedunion
\tupleof{\maxpolicy_j}_{j \in \forallsof{}}$
is optimal in all CTDE methods.
 In~\cite{son2019qtranlearningfactorizetransformation}, a formal property called
\emph{Individual--Global--Max} (IGM) is introduced.
IGM property states that maximizing the joint Q-function is equivalent to independently maximizing each agent’s local function.
\begin{definition}
For a joint action--value function $\jointqvalue(\jointhistory,\actiondec)$,
if there exist a tuple of functions $\tupleof{\Qvalue_\ell}_{\ell\in \{1,\ldots,n \}}$ such that:
\vspace{-2mm}
\begin{align*}
\arg\max_{\actiondec} \jointqvalue(\jointhistory,\actiondec)
=
\tupleof{
\arg\max_{\action_\ell} \Qvalue_\ell(h_\ell,\action_\ell)
}_{\ell \in \{1,\ldots, n \}}\\[-3em]
\end{align*}
\end{definition}
where $\tupleof{\Qvalue_\ell}_{\ell\in \{1,\ldots,n \}}$ satisfies \emph{IGM} for $\jointqvalue$ given $\jointhistory$.
Among the CTDE algorithms considered, QTRAN satisfies IGM more generally, while QMIX satisfies it via monotonicity.

\begin{theoremstatement}{}\label{theorem:Qlearning}
Let $\pomdp$ be a POMDP and $\varphi$ a \HyperLTL formula.
If a CTDE algorithm learns an optimal joint action--value function $\jointqvalue^\optimal$ satisfying the IGM property,
then 
$\tupleof{\maxpolicy_i}_{i \in \existsof{}}$ and
$\tupleof{\maxpolicy_j}_{j \in \forallsof{}}$
induced by the joint policy $\jointpolicy^\optimal$ will optimize the satisfaction probability of $\skolemized{\varphi}$.
\end{theoremstatement}
\vspace{-3mm}
\Cref{theorem:Qlearning} establishes the premise of \cref{theorem:skolem}, which in turn solves the original problem \Cref{fig:hypole} (proofs in \cref{sec:proof}).

%% file: experiment.tex
\vspace{-3mm}
\section{Experiments and Results}
\label{sec:eval}

\begin{figure}[t]
    \centering
    \includegraphics[width=0.99\linewidth]{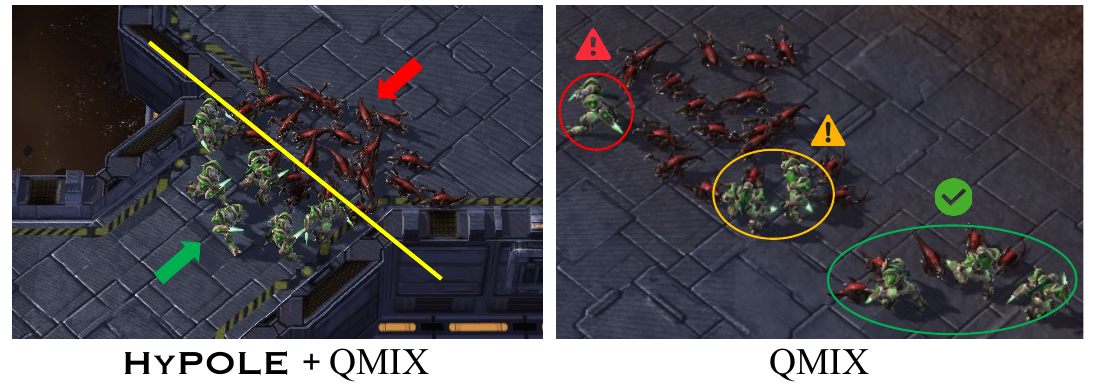}
    \captionof{figure}{\method+QMIX vs. QMIX on SMAC corridor map.}
    \label{fig:corridor_tac}
    \vspace{-5mm}
\end{figure}

\begin{figure*}[t]
    \centering

\vspace{-1mm}
    
    \begin{subfigure}[t]{\textwidth}
        \centering
        \includegraphics[width=0.99\linewidth]{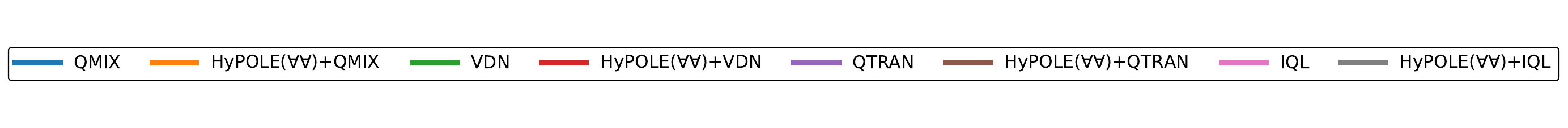}
    \end{subfigure}

\vspace{-5mm}
    \begin{subfigure}[t]{0.3\textwidth}
        \centering
        \includegraphics[width=\linewidth]{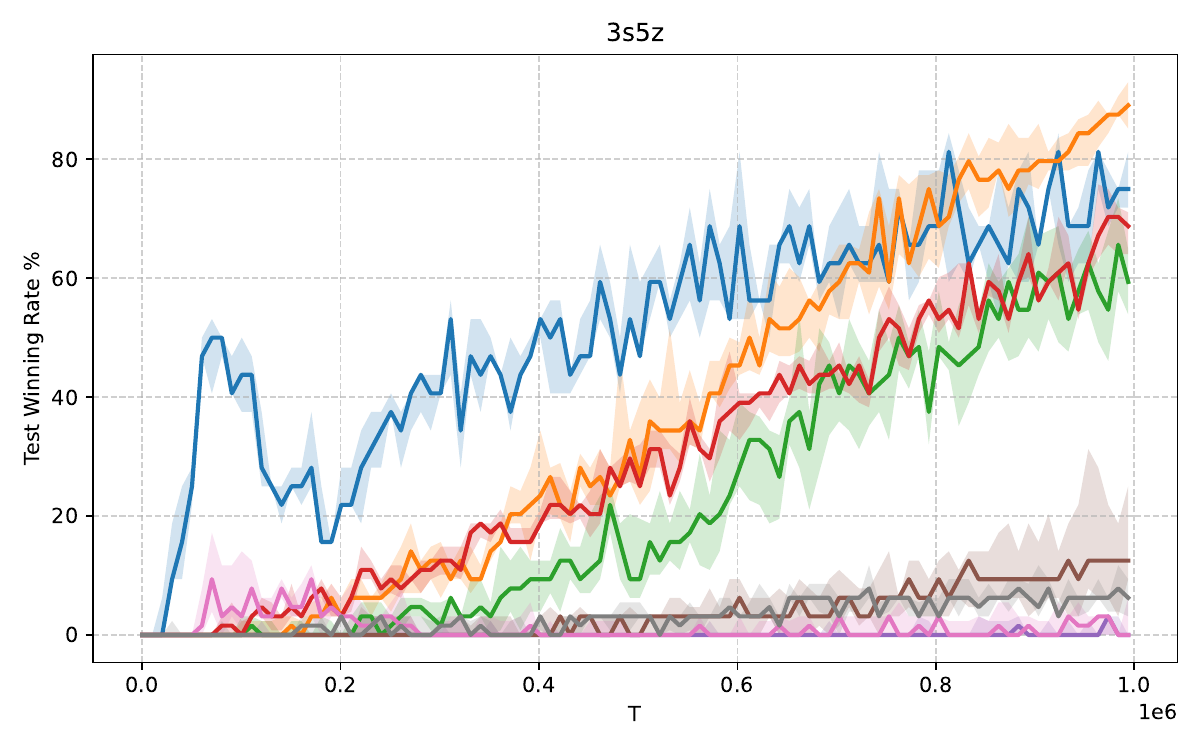}
        \vspace*{-8mm}
        \caption{3s5z}
        \label{fig:3s5z}
    \end{subfigure}\hfill
    \begin{subfigure}[t]{0.3\textwidth}
        \centering
        \includegraphics[width=\linewidth]{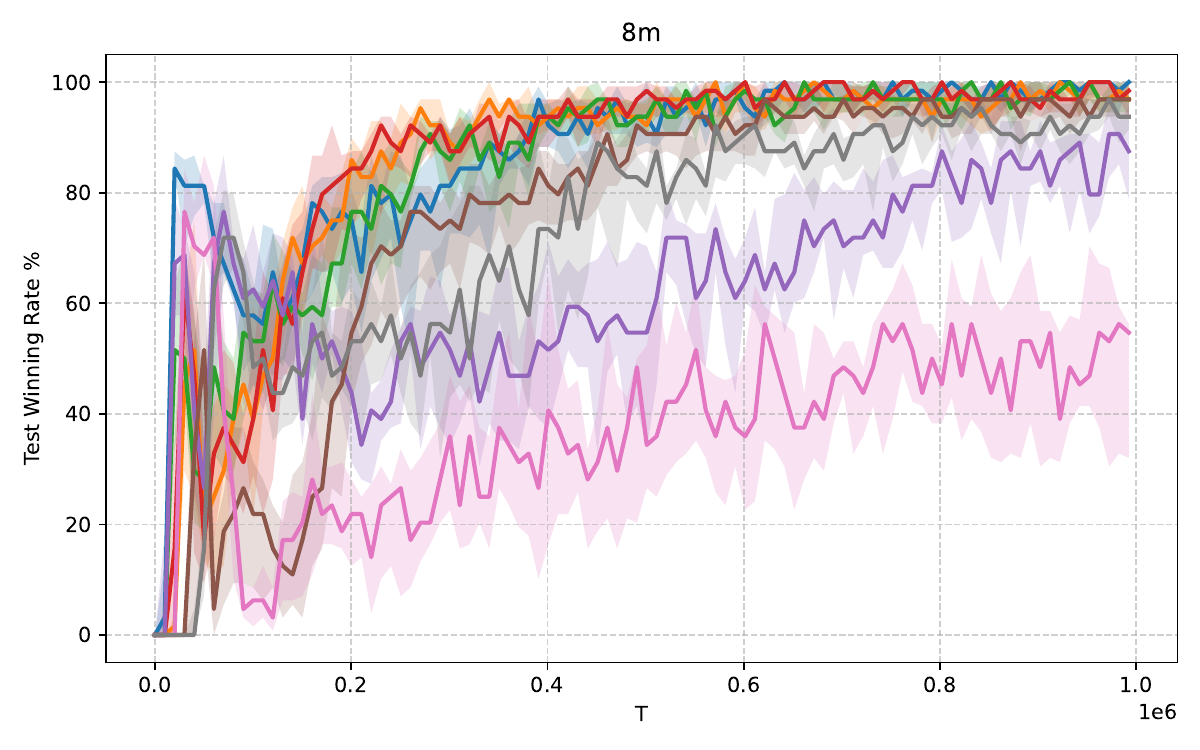}
        \vspace*{-8mm}
        \caption{8m}
        \label{fig:8m} 
    \end{subfigure}\hfill
    \begin{subfigure}[t]{0.3\textwidth}
        \centering
        \includegraphics[width=\linewidth]{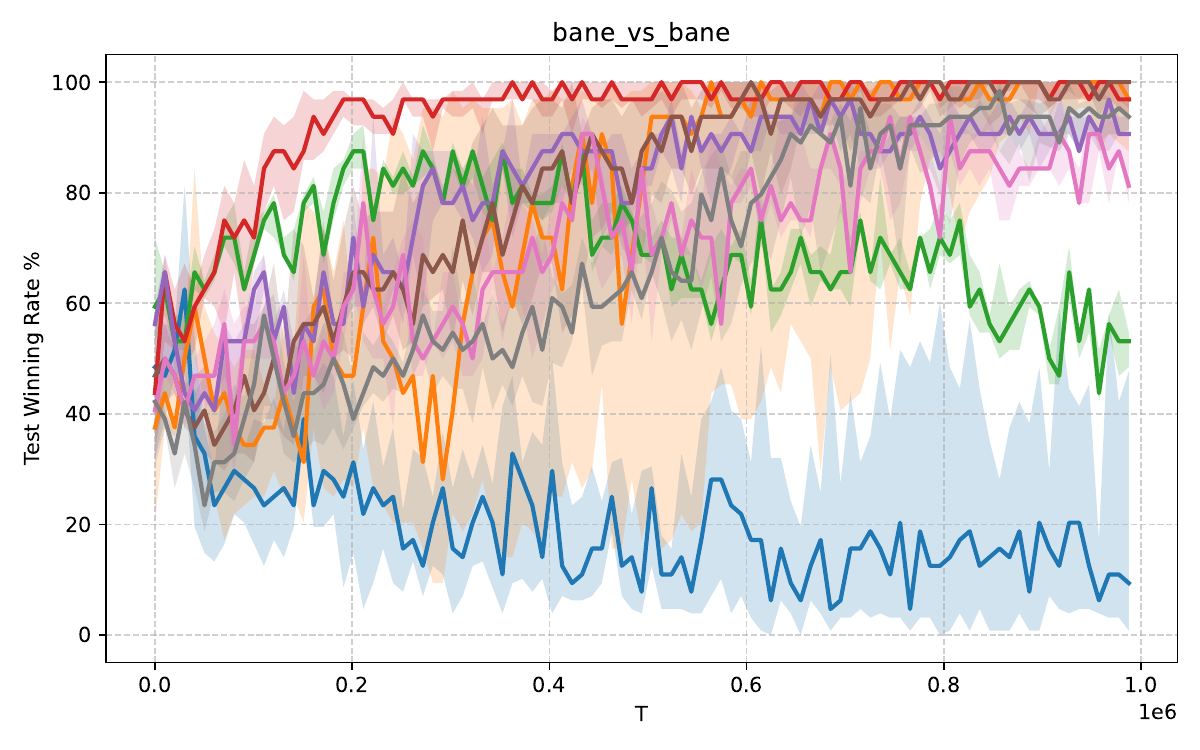}
        \vspace*{-8mm}
        \caption{bane\_vs\_bane}
        \label{fig:bane}
    \end{subfigure}
        \begin{subfigure}[t]{0.3\textwidth}
        \centering
        \includegraphics[width=\linewidth]{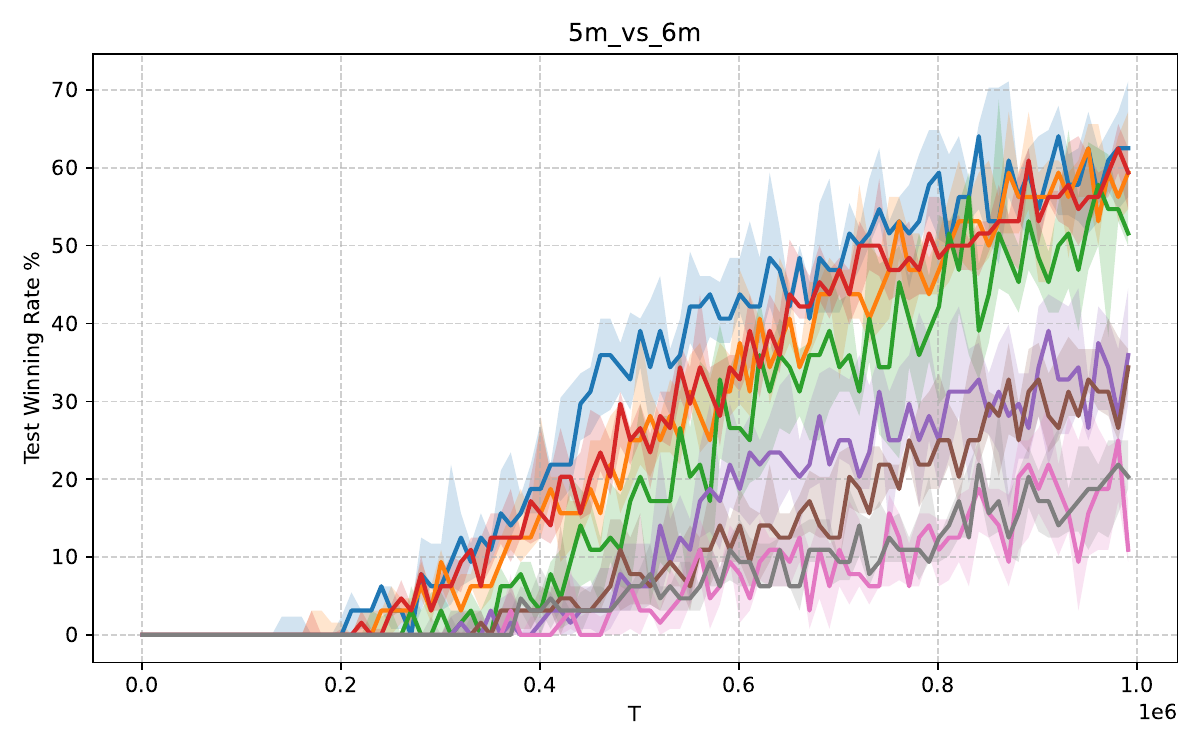}
        \vspace*{-8mm}
        \caption{5m\_vs\_6m}
        \label{fig:5m6m}
    \end{subfigure}\hfill
    \begin{subfigure}[t]{0.3\textwidth}
        \centering
        \includegraphics[width=\linewidth]{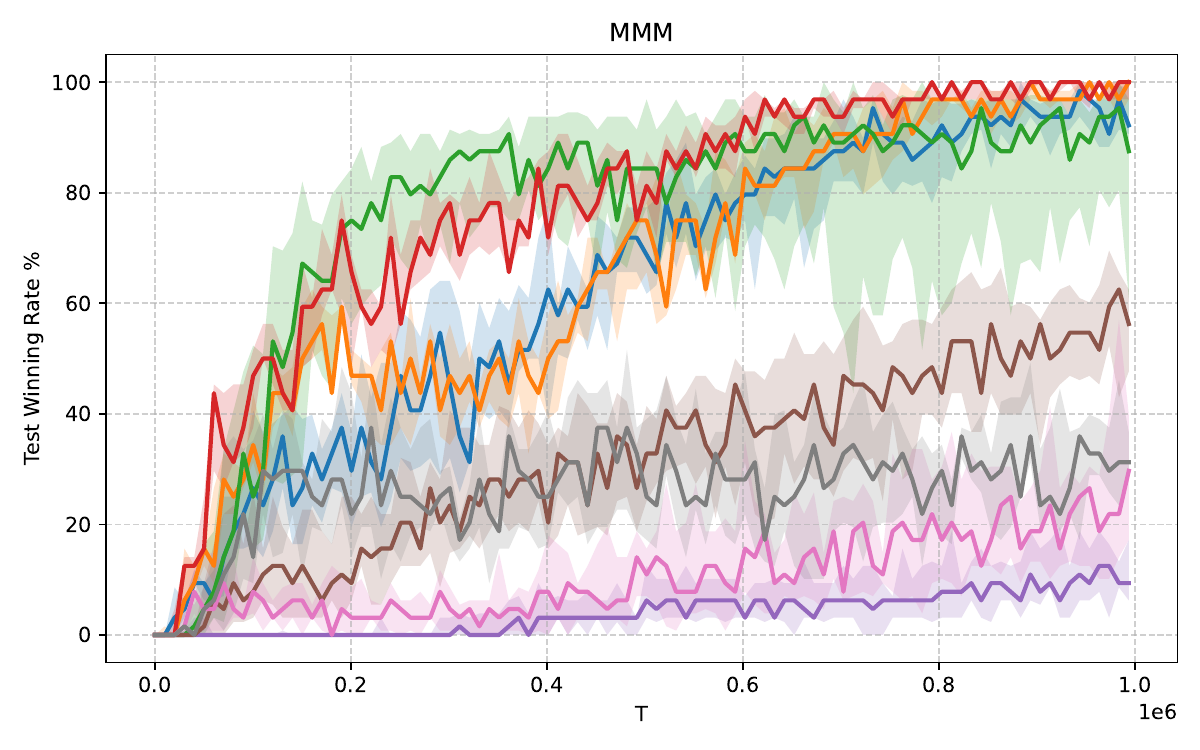}
        \vspace*{-8mm}
        \caption{MMM}
        \label{fig:MMM}
    \end{subfigure}\hfill
    \begin{subfigure}[t]{0.3\textwidth}
        \centering
        \includegraphics[width=\linewidth]{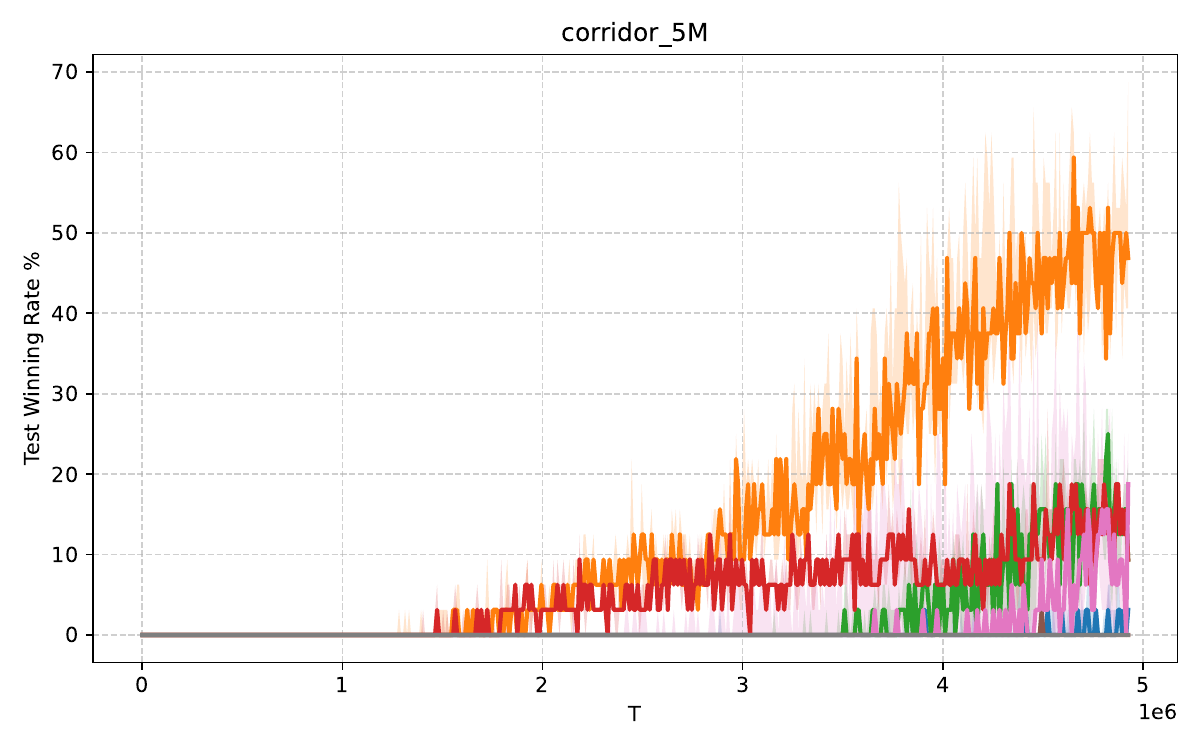}
        \vspace*{-8mm}
        \caption{corridor}
        \label{fig:corridor}
    \end{subfigure}

    \caption{Learning curves on different SMAC maps, showing median test win rate with 25-75\% percentile.}
    \vspace*{-4mm}
    \label{fig:winnig_main}
\end{figure*}

\paragraph{Implementation.} \method is implemented on top of \texttt{PyMARL} repository~\cite{samvelyan19smac}.
We start with a Skolemized \HyperLTL formula.
Using the quantitative semantics described in \cref{sec:quantsemantics}, we construct robustness functions and use the resulting robustness values as reward signals in a cooperative MARL environment.
Our objective is to synthesize policies $\policy_\ell$ for each agent $\ell \in \setof{1,\ldots,n}$, depending on the quantifier associated with its trace variable.
For universally quantified traces, i.e., policies $\tupleof{\policy_j}_{j \in \forallsof{}}$, each policy receives only the agent’s local history.
In contrast, for existentially quantified traces, i.e., policies $\tupleof{\policy_i}_{i \in \existsof{}}$, the policy input is the history induced by the output of the corresponding Skolem function $\skolemfunc_i$.

We train an \seqseq model to predict the traces of the preceding universally quantified variables.
During centralized training, mini-batches sampled from the replay buffer $\buffer$ are used to train the \seqseq model every $\steptrain$ steps.
The \seqseq model follows a feedforward autoencoder-style architecture.
We then employ value-based CTDE algorithms, namely VDN, QMIX, and QTRAN 
to construct the policies. 
In addition, we evaluate Independent Q-Learning (IQL) \cite{tan1993multi}.
For IQL, we consider only $\forall\forall$-form \HyperLTL specifications, which evaluates using local observations alone.
Additional details on the experimental setting are provided in~\Cref{appendix:exp-details}.

\vspace{-2mm}
\begin{figure*}[t!]
    \centering
    \vspace{-1mm}
    
    \begin{subfigure}[t]{\textwidth}
        \centering
        \includegraphics[width=0.99\linewidth]{img/legend_only.pdf}
    \end{subfigure}

\vspace{-5mm}

    \begin{subfigure}[t]{0.3\linewidth}
        \centering
        \includegraphics[width=\linewidth]{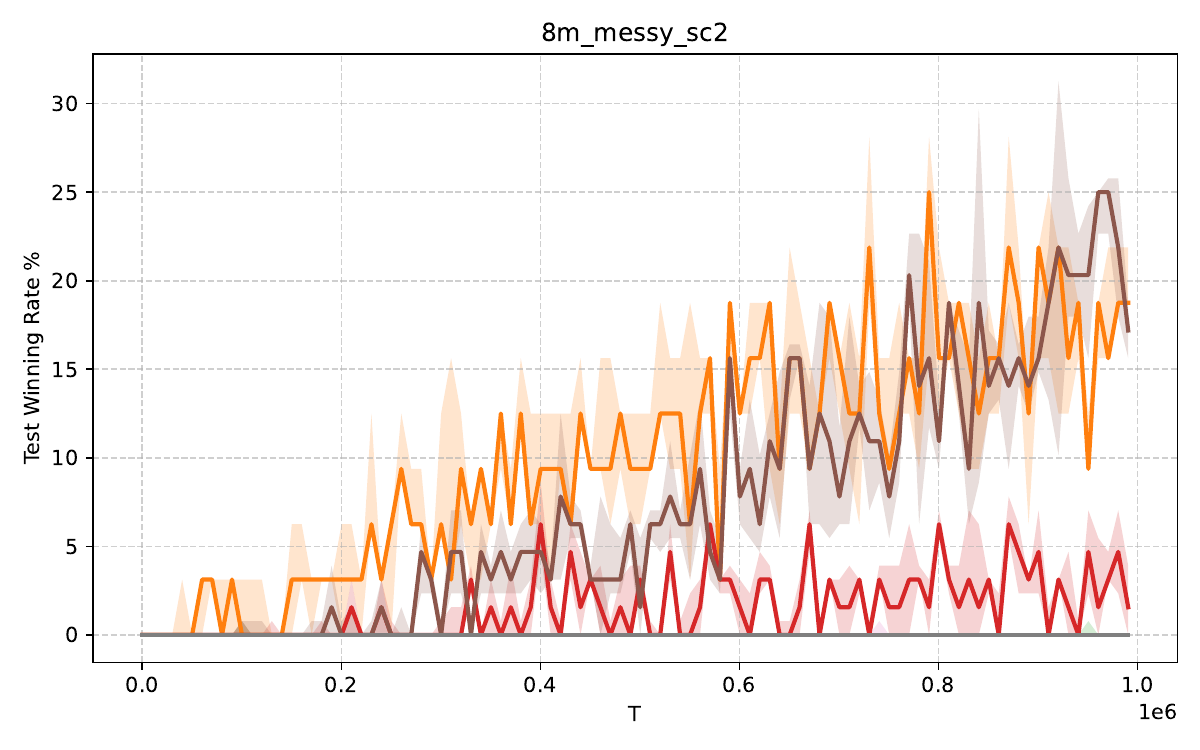}
        \caption{8m ($K=12, \phi = 0.15$)}
        \label{fig:8mmessy}
    \end{subfigure}\hfill
    \begin{subfigure}[t]{0.3\linewidth}
        \centering
        \includegraphics[width=\linewidth]{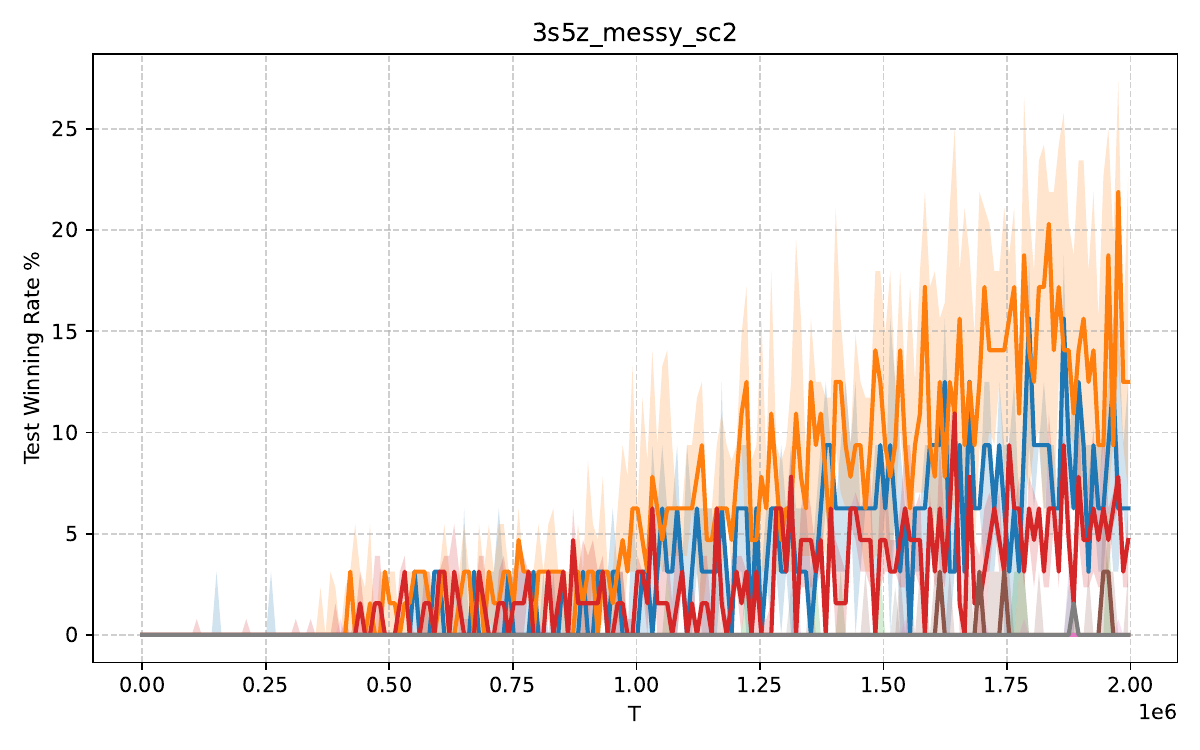}
        \caption{3s5z ($K=12, \phi = 0.15$)}
        \label{fig:3s5zmessy}
    \end{subfigure}\hfill
    \begin{subfigure}[t]{0.3\linewidth}
        \centering
        \includegraphics[width=\linewidth]{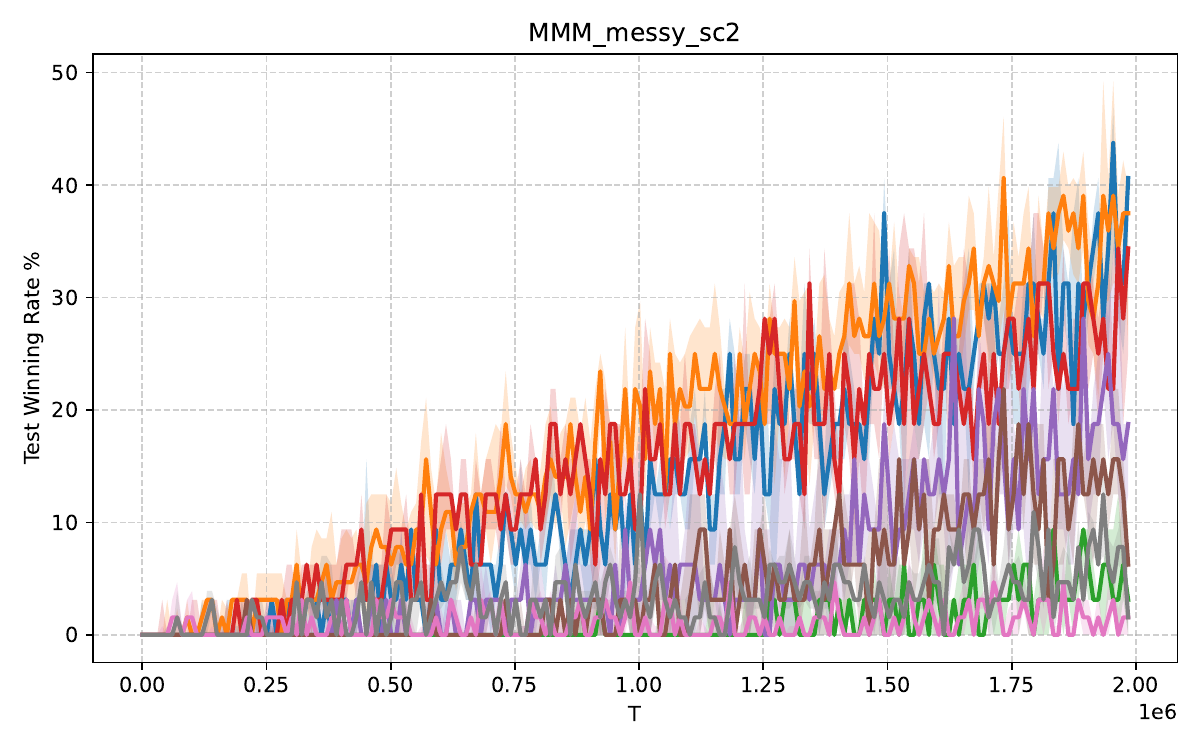}
        \caption{MMM ($K=12, \phi = 0.15$)}
        \label{fig:MMMmessy}
    \end{subfigure}

    \caption{Learning curves on different MessySMAC maps, showing median test win rate with 25-75\% percentile. }
    \vspace*{-2mm}
    \label{fig:smacmessy}
\end{figure*}

\paragraph{Case Studies.}
We conduct our experiments on scenarios from the StarCraft Multi-Agent Challenge (SMAC) ~\cite{samvelyan19smac}, which is based on the strategy game 
StarCraft~II and focuses on cooperative micromanagement tasks.
In addition, we use MessySMAC~\cite{pmlr-v202-phan23a}, which introduces observation stochasticity $\phi$ and randomized initialization with $K$ random steps before each episode starts. This case study adds significant challenges for finding optimal policies in SMAC.
Finally, we extend the WildFire benchmark introduced in~\cite{hsu2025hyprl} to a partially observable setting. 
More details on SMAC, MessySMAC, and WildFire are provided in~\Cref{app:smac_messy,appendix:wildfire}.

\vspace{-2mm}
\paragraph{Experimental Organization.}
We evaluate \method on nine SMAC scenarios, including both hard and super-hard scenarios (see \cref{tab:smac_maps} in the Appendix), five MessySMAC scenarios, and four WildFire scenarios (see \cref{tab:wild_maps} in the Appendix).
We (1) assess the effect of using robustness values from $\forall\forall$-\HyperLTL specifications as 
reward signals in MARL, compared to baseline MARL algorithms using standard shaped rewards;
(2) compare $\forall\exists$-\HyperLTL specifications with $\forall\forall$-\HyperLTL specifications 
in terms of learning efficiency and performance;
(3) investigate how \HyperLTL enables expressing distinct combat tactics, and
(4) discuss \HyperLTL specifications that fail to provide effective guidance for learning.
All specifications used in our experiments are presented in \Cref{sec:formula}.

\begin{figure*}[t]
    \centering

    \vspace{-2mm}
    \begin{subfigure}[t]{\textwidth}
        \centering
        \includegraphics[width=0.99\linewidth]{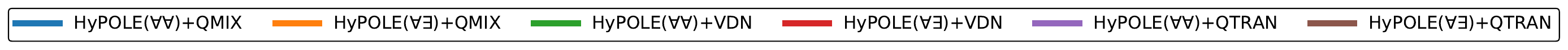}
    \end{subfigure}
\vspace{-5mm}

    \begin{subfigure}[t]{0.3\linewidth}
        \centering
        \includegraphics[width=\linewidth]{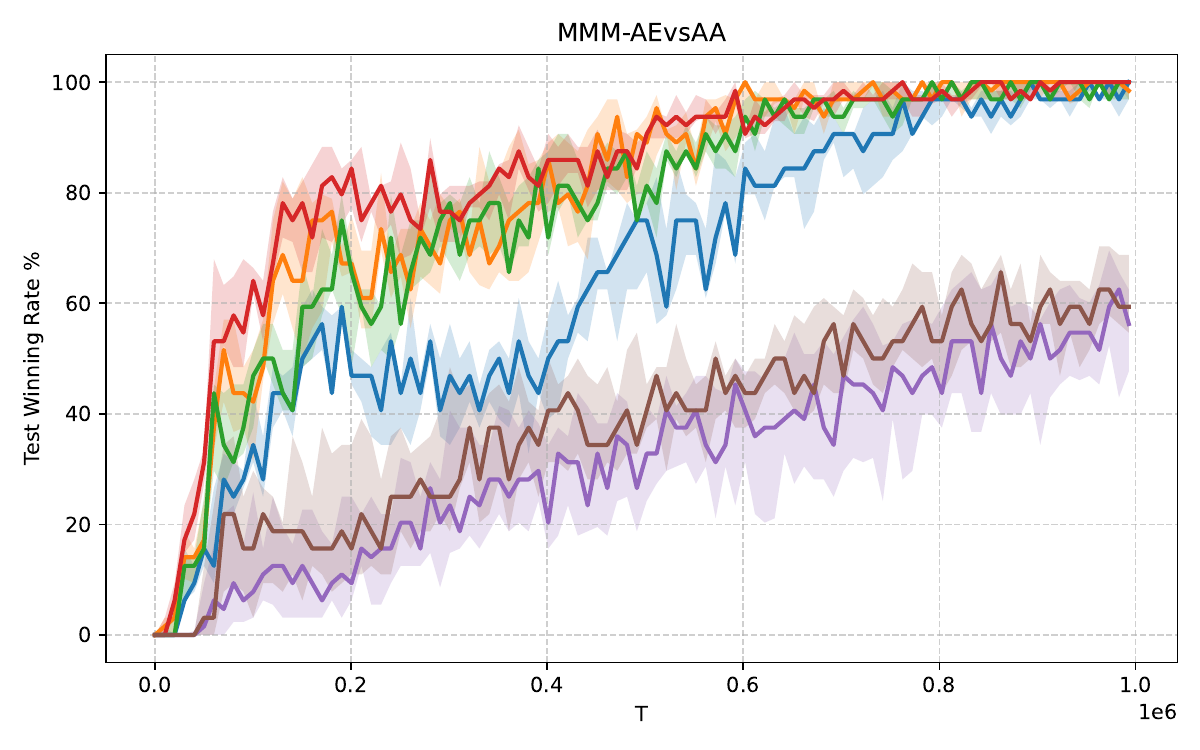}
        \caption{MMM (SMAC)}
        \label{fig:quantMMM}
    \end{subfigure}\hfill
    \begin{subfigure}[t]{0.3\linewidth}
        \centering
        \includegraphics[width=\linewidth]{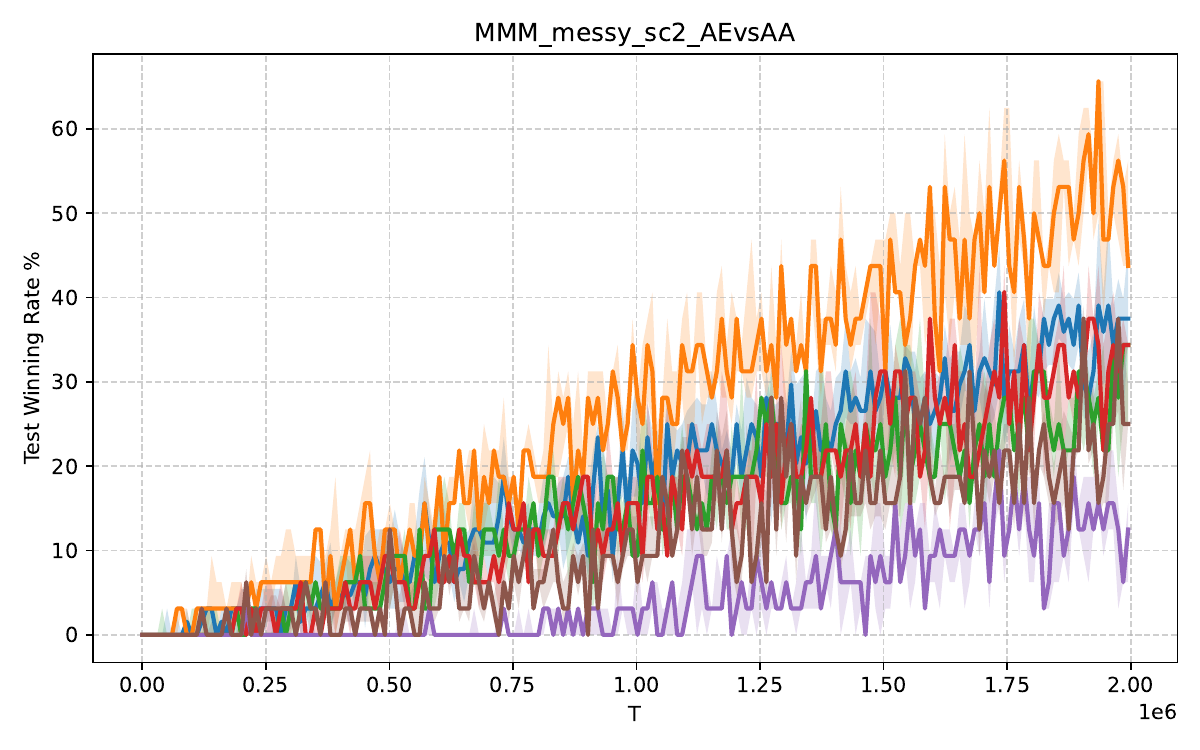}
        \caption{MMM (MessySMAC)}
        \label{fig:quantMMMmessy}
    \end{subfigure}\hfill
    \begin{subfigure}[t]{0.3\linewidth}
        \centering
        \includegraphics[width=\linewidth]{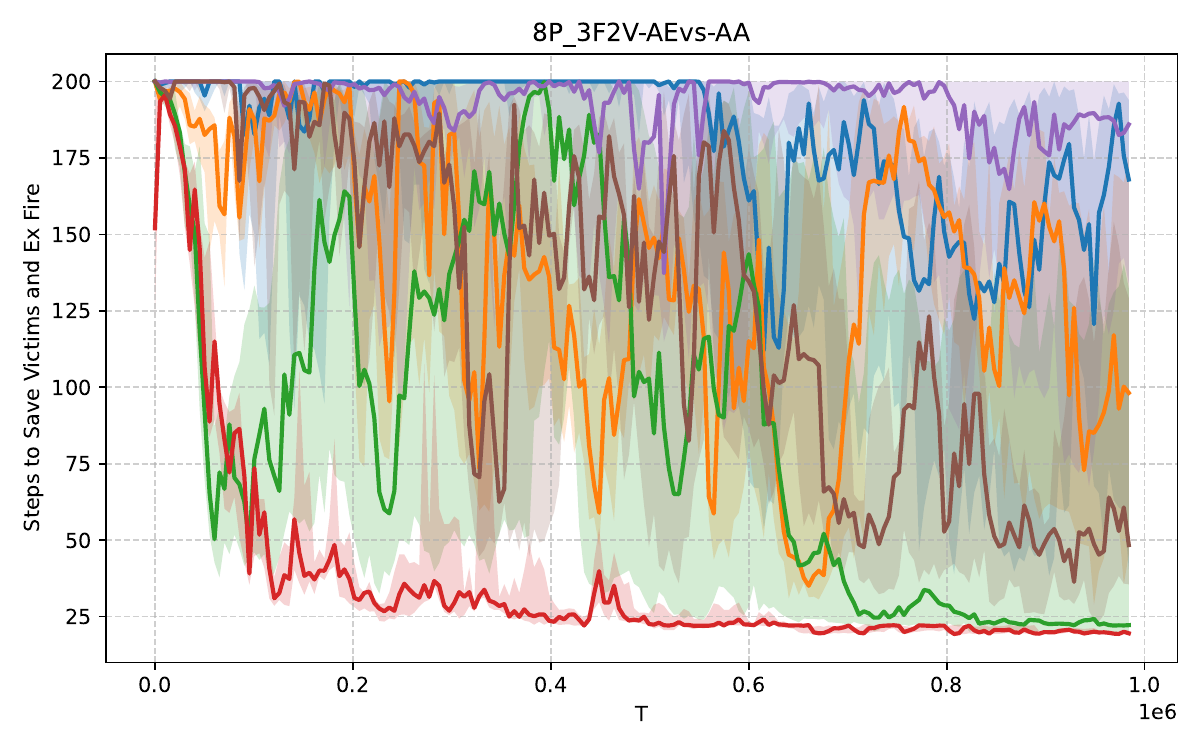}
        \caption{8P\_3F2V (WildFire)}
        \label{fig:quant8pwild}
    \end{subfigure}

    \caption{$\forall\forall$ vs. $\forall\exists$ learning curves across MessySMAC, SMAC, and WildFire, showing median test win rate with 25--75\% percentiles.}
    \vspace*{-2mm}
    \label{fig:quantefiers}
\end{figure*}

\begin{figure*}[t!]

    \centering
    \begin{subfigure}[t]{0.3\linewidth}
        \centering
        \includegraphics[width=\linewidth]{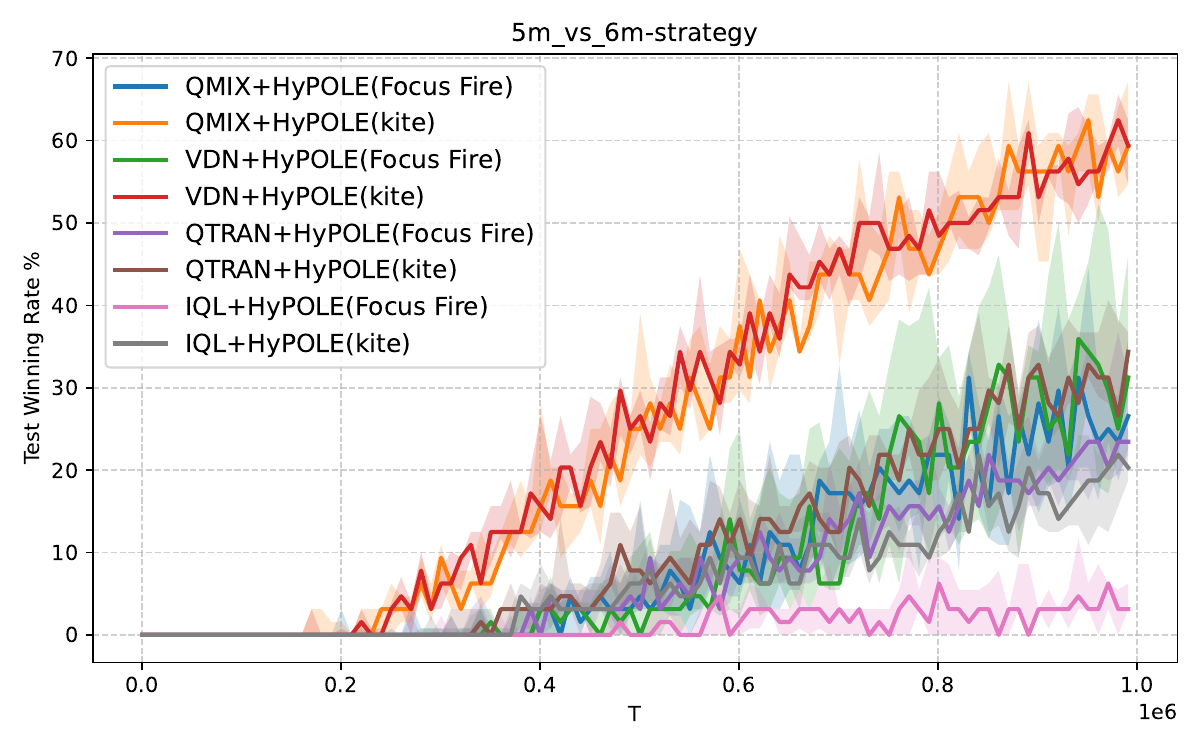}
        \caption{$\kite$ vs.\ $\focus$ in 5m\_vs\_6m}
        \label{fig:strategy_top}
    \end{subfigure}\hfill
    \begin{subfigure}[t]{0.3\linewidth}
        \centering
        \includegraphics[width=\linewidth]{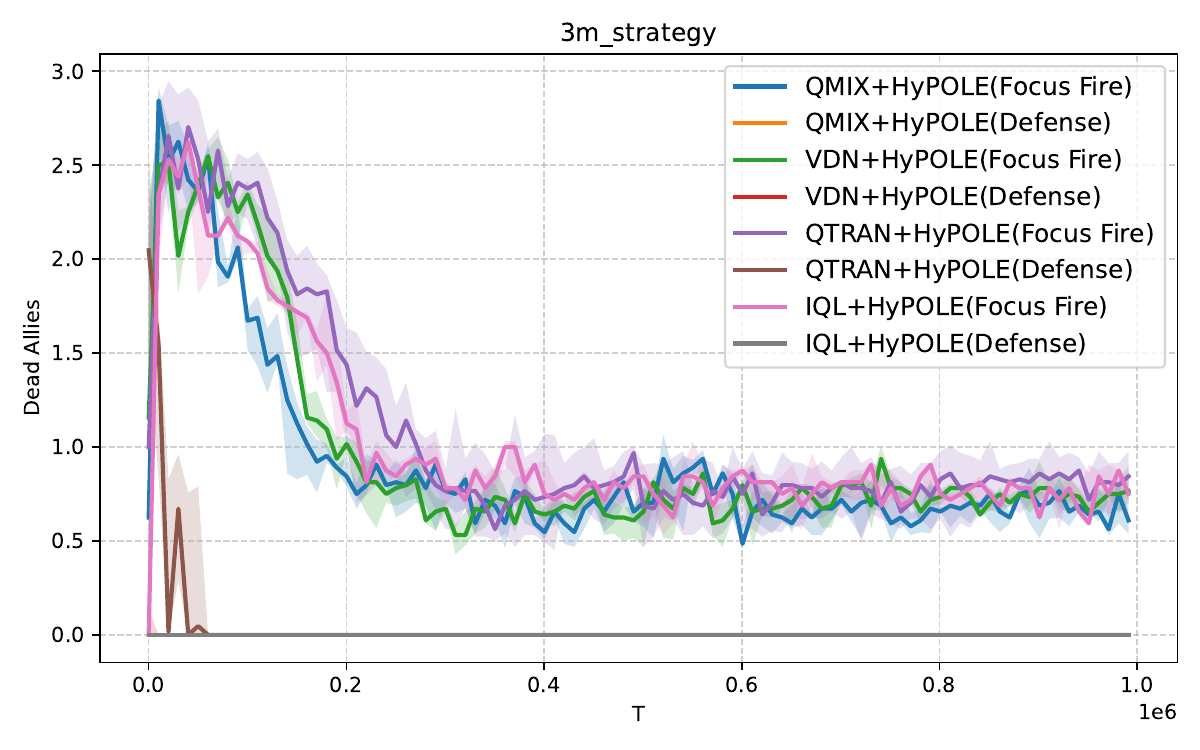}
        \caption{$\defense$ vs.\ $\focus$ in 3m}
        \label{fig:strategy_bot}
    \end{subfigure}\hfill
    \begin{subfigure}[t]{0.3\linewidth}
        \centering
        \includegraphics[width=\linewidth]{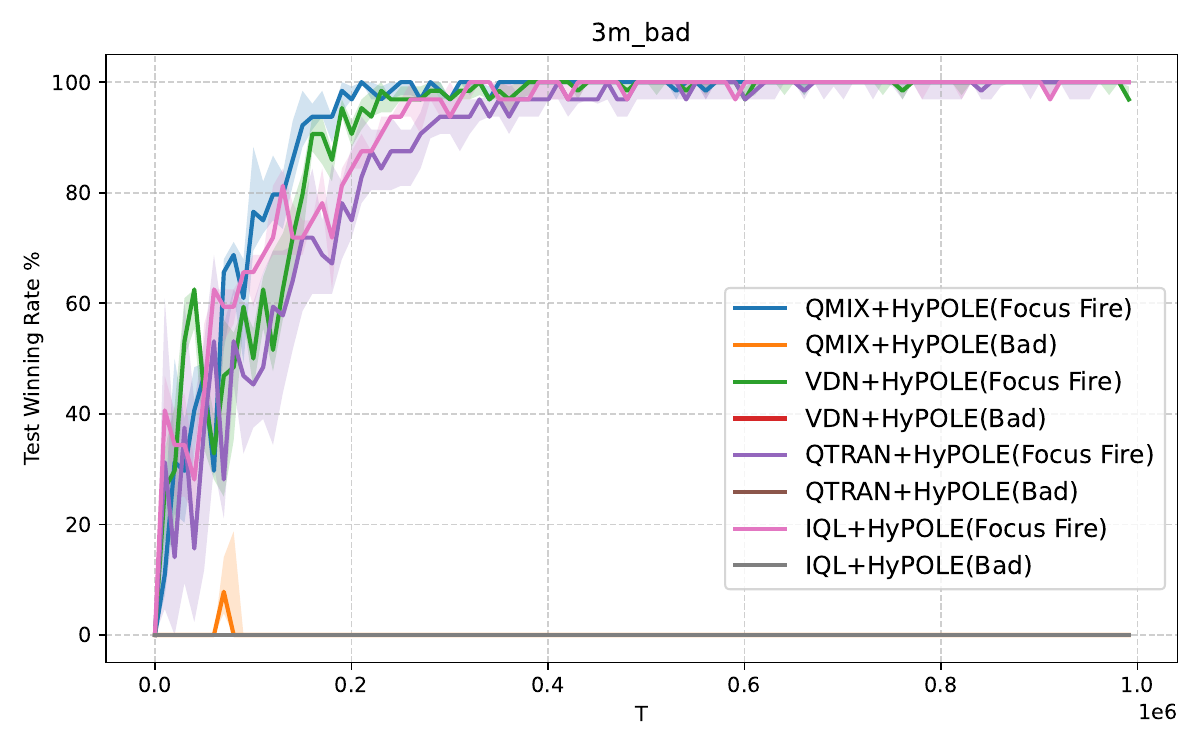}
        \caption{Results on Bad \HyperLTL Formula.}
        \label{fig:bad}
    \end{subfigure}

    \caption{Expressing tactics using \method, and results on a bad \HyperLTL formula.}
    \vspace*{-2mm}
    \label{fig:strategy_bad_row}
\end{figure*}

\vspace{-2mm}
\paragraph{\method vs. Shaped Rewards.}
We compare the robustness values generated by \method using $\forall\forall$-\HyperLTL specifications with the shaped reward functions introduced in~\cite{samvelyan19smac}.
\Cref{fig:winnig_main} reports the median test win rate in SMAC (i.e., elimination of all enemy agents) of 
\method+MARL, where $\text{MARL} \in \{\text{VDN}, \text{QMIX}, \text{QTRAN}, \text{IQL}\}$, compared against vanilla MARL using shaped rewards.
In \cref{fig:3s5z}, we present our results for the $\focus$ on the \texttt{3s5z} map, where $\focus$ encourages allied agents to focus fire on enemies.
Across all settings for this map, \method+MARL outperforms the vanilla MARL baselines.
Notably, \method+QMIX \orangeline initially underperforms vanilla QMIX \blueline, but after $\approx700$K environment steps the agents learn the focus fire tactic and subsequently surpass vanilla QMIX.
Similarly, on the \texttt{8m} map~\cref{fig:8m}, \method+MARL using $\focus$ outperforms the vanilla MARL baselines in all cases.
In particular, \method+IQL \grayline not only surpasses vanilla IQL \pinkline, but also outperforms vanilla QTRAN \purpleline, despite QTRAN being a more complex CTDE method.
Moreover, \method+IQL remains competitive with vanilla VDN \greenline and QMIX \blueline.
We also evaluate $\focus$ on \texttt{bane\_vs\_bane} map~\cref{fig:bane}, where \method+QMIX \orangeline and \method+VDN \redline significantly outperform vanilla QMIX \blueline and VDN \greenline.

\Cref{fig:5m6m} evaluates \method+MARL using $\kite$ formula on \texttt{5m\_vs\_6m}, where $\kite$ encourages agents to attack only when their weapon is ready.
For all cases on this map, \method+MARL competes with the vanilla MARL baselines.
In \cref{fig:MMM}, we evaluates the $\medivac$ on \texttt{MMM} map, where $\medivac$ encourages Medivac agents to move toward low-health allies to heal them.
In all cases, \method+MARL outperforms the vanilla baselines. In particular, \method+QTRAN \brownline and \method+IQL \grayline substantially improve over vanilla QTRAN \purpleline and IQL \pinkline, respectively.
We find \method particularly promising on the \texttt{corridor} map (\cref{fig:corridor}), which is widely regarded as one of the hardest SMAC maps.
We use $\corridor$ formula, which encourages agents to hold the choke point and damage enemies from that position.
Using this formula, \method+QMIX \orangeline achieves a median win rate of up to $\approx 60\%$, substantially outperforming vanilla QMIX \blueline, which achieves a win rate of less than $5\%$.
To illustrate this, \Cref{fig:corridor_tac} shows snapshots of agent behavior after 5M training steps.
Under \method+QMIX, agents form a defensive line at the choke point. 
In contrast, vanilla QMIX agents lose formation and split into three groups. 
One group of three agents, marked by \raisebox{-0.15em}{\includegraphics[height=1.0em]{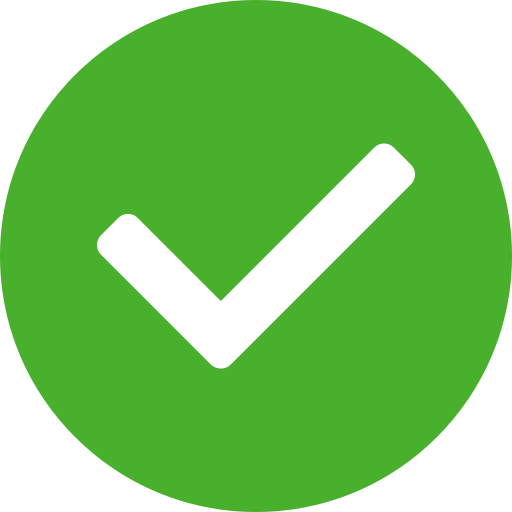}}, is surrounded by five enemies and may still survive because \texttt{Zealots} are stronger than \texttt{Zerglings}. 
However, the single agent marked by \raisebox{-0.15em}{\includegraphics[height=1.0em]{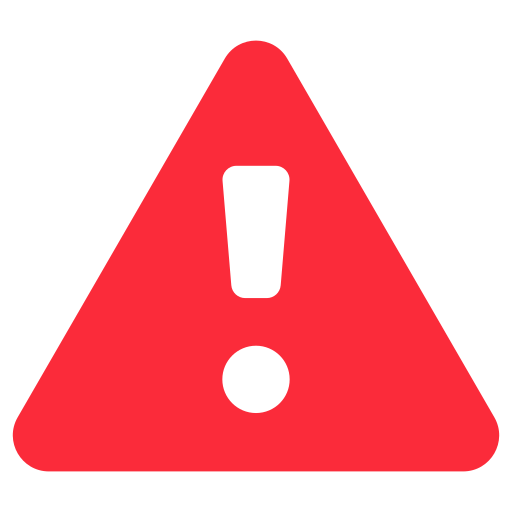}} is surrounded by more than ten \texttt{Zerglings} and is likely to be defeated. 
This then exposes the two agents marked by \raisebox{-0.15em}{\includegraphics[height=1.0em]{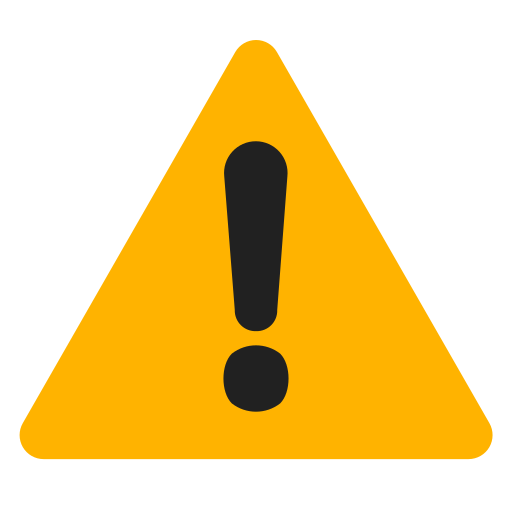}}, leading to the loss of three allies and eventually the remaining agents, who are heavily outnumbered.

\vspace{3mm}
We also compare \method+MARL against vanilla MARL on MessySMAC, with results shown in~\Cref{fig:smacmessy}. 
In \texttt{8m} map~\Cref{fig:8mmessy}, we use the $\focus$ formula.
Across all cases, \method+CTDE outperforms vanilla CTDE. 
In particular, \method+QTRAN \brownline and \method+QMIX \orangeline achieve win rates of up to $25\%$, whereas the vanilla baselines fail to achieve any wins.
Similarly, we use $\focus$ in \texttt{3s5z} map~\Cref{fig:3s5zmessy}.
In this scenario, \method+VDN \redline outperforms vanilla VDN, which fails to achieve any wins. 
Moreover, \method+QMIX \orangeline performs slightly better than vanilla QMIX \blueline.
In \texttt{MMM} map~\Cref{fig:MMMmessy}, we use the $\medivac$ formula. 
Here, \method+VDN \redline achieves a win rate of up to $\approx 35\%$, substantially outperforming vanilla VDN \greenline, which remains below $10\%$.
The remaining SMAC and MessySMAC maps, together with the full WildFire comparison against hand-built rewards, are provided in ~\Cref{sec:shape_app}.

\vspace{-2mm}
\paragraph{$\boldsymbol{\forall\exists}$ vs.\ $\boldsymbol{\forall\forall}$ Specifications.}
We study the effect of using $\forall\exists$ variants of the $\medivac$ and $\wild$ specifications, and show that they can lead to improved policies by expanding the policy search space.
In the \texttt{MMM} scenario in SMAC and MessySMAC, there is an inter-agent dependency between the Medivac and the remaining agents.
To capture this, we change the quantifier associated with the Medivac from $\forall$ to $\exists$ in $\medivac$.
In \texttt{MMM} scenario in SMAC~\cref{fig:quantMMM}, \method($\forall\exists$)+QMIX  \orangeline and \method($\forall\exists$) +VDN \redline outperform \method($\forall\forall$)+QMIX \blueline and \method ($\forall\forall$)+VDN \greenline.
For the MessySMAC benchmark, \Cref{fig:MMMmessy} shows that \method($\forall\forall$)+QMIX has only a slight edge over the vanilla baseline. In contrast, \Cref{fig:quantMMMmessy} shows that \method+QMIX($\forall\exists$) \orangeline performs better than \method+QMIX($\forall\forall$) \blueline and, consequently, the vanilla QMIX baseline.
Similarly, in~\Cref{fig:MMMmessy}, \method+QTRAN($\forall\forall$) performs worse than the vanilla baseline. However, using \method+QTRAN($\forall\exists$) \brownline in~\Cref{fig:quantMMMmessy} improves the performance of \method.
In \Cref{fig:quant8pwild}, we observe that the $\forall\exists$ version of the $\wild$ formula helps \method achieve better results in the WildFire \texttt{8P\_3F2V} scenario.
Specifically, \method+VDN($\forall\exists$) \redline performs substantially better than \method+VDN($\forall\forall$) \greenline, converging after 500K training steps compared to 800K steps. 
Additional results comparing $\forall\exists$ and $\forall\forall$ are provided in \cref{sec:aaee-app}.

\vspace{-2mm}
\paragraph{Tactics as \HyperLTL Formulas.}
A key advantage of \method is learning policies that maximize satisfaction of \HyperLTL-encoded tactics.
In \cref{fig:strategy_top}, we study the \texttt{5m\_vs\_6m} scenario and show that the kiting ($\kite$) yields consistently higher win rates than the focus firing ($\focus$), and this trend is consistently evident in all cases. 
In \texttt{3m} map~\cref{fig:strategy_bot}, we further demonstrate the expressiveness of \method, by specifying a defensive tactic using $\defense$, which encourages allied agents to retreat from enemies and results in zero ally casualties.
This experiment shows that diverse tactics can be encoded in \HyperLTL and compiled into policies via \method.

\vspace{-2mm}
\paragraph{Quality of \HyperLTL Specifications.}
We also study the effect of weakening a \HyperLTL specification by removing a subformula from the $\focus$, resulting in the $\bad$. 
This modification eliminates the constraint that encourages agents to shoot enemies. 
In \Cref{fig:bad}, \method with $\bad$ formula achieves an almost zero win rate.
This shows that the performance of \method strongly depends on the quality and completeness of the underlying \HyperLTL specification.

%% file: conc.tex
\section{Conclusion} 
\label{sec:conclusion}

We presented \method, a specification-guided MARL framework that enables model-free synthesis of decentralized policies from \HyperLTL specifications over POMDPs.
\method effectively handles complex multi-agent objectives under partial observability by leveraging the semantics of \HyperLTL to guide the learning process by off-the-shelf MARL algorithms.
\method is fully implemented and we showed superior performance over baselines, namely, vanilla versions of VDN, QMIX, QTRAN, and IQL, across a set of diverse case studies.

\paragraph{Limitations.} Admittedly, the main challenge in using \method is writing precise \HyperLTL specifications.
Low-quality specifications may hinder the learning process.
We believe this limitation opens up exciting future work on AI-enabled specification generation.
Moreover, the current version of \method is limited to discrete-action settings because it relies on value-based CTDE methods, which we choose over actor-critic methods due to their IGM-like guarantees.
Finally, \method is not the method of choice when the task does not involve relational dependencies or temporal behaviors between agents and can instead be well captured by standard reward design, such as tasks with independent objectives or simple non-temporal requirements.

%% file: theory-appendix.tex
\section{Proofs}
\label{sec:proof}

\subsection{Proof of \cref{theorem:skolem}}
\label{proof:skolem}

\begin{proof}

Let $\pomdp$ be a POMDP and let $\skolemized{\varphi}$ denote the Skolemized form of the HyperLTL formula $\varphi$, written as
\[
\skolemized{\varphi} =
\exists \skolemfunc_i(\tau_{i_1}, \ldots, \tau_{i_{|\forallsof{i}|}})
\;\forall \tracevar_j.\;
\skolemized{\psi},
\]
for each $i \in \existsof{}$ and $j \in \forallsof{}$.
Assume that the tuple of policies
$
\tupleof{\maxpolicy_i}_{i\in\existsof{}}
\orderedunion
\tupleof{\maxpolicy_j}_{j\in\forallsof{}}
$
optimizes the robustness-based objective induced by $\skolemized{\varphi}$.

Since $\maptostate$ is path-consistent, for any history $h$ generated by a latent path $\zeta$,
$
\hat{\traj}
\;\definedas\;
\arg\max_{\traj \in \trajs^*} \maptostate(h,\traj)
$
coincides with the actual path $\zeta$ that generated $h$.
Consequently, for any $k \ge 0$,
$\traceof{\hat{\zeta}_{[0:k]}} = \traceof{\zeta_{[0:k]}}$.

Fix an arbitrary $k \ge 0$.
The zipped path induces a trace assignment for the variables of $\varphi$ as follows.

\begin{itemize}
    \item For each universally quantified trace variable $\tracevar_j$, where $j \in \forallsof{}$,
    \[
    \tracevar_j \;\mapsto\;
    \traceof{
        (
        \arg\max_{\traj \in \trajs^*}
        \maptostate(h_j,\traj)
        )_{[0:k]}
    }.
    \]

    \item For each existentially quantified trace variable $\tracevar_i$, where $i \in \existsof{}$,
    \[
    \tracevar_i \;\mapsto\;
    \skolemfunc_i\Big(
    \traceof{
        (\arg\max_{\traj \in \trajs^*}
        \maptostate(h_{i_1},\traj))_{[0:k]}
    },
    \ldots,
    \traceof{
        (\arg\max_{\traj \in \trajs^*}
        \maptostate(h_{i_{|\forallsof{i}|}},\traj))_{[0:k]}
    }
    \Big).
    \]
\end{itemize}

By construction of Skolemization, each Skolem function $\skolemfunc_i$ depends only on trace variables corresponding to universally quantified indices preceding $i$.
Moreover, if $i_\ell = j$ for some $\ell$, then the corresponding trace prefixes coincide.
Since the given policy tuple optimizes $\skolemized{\varphi}$, it maximizes the probability that the robustness value
\[
\rbvalue (
\zip(\cdot),
\skolemized{\psi}
)
\]
converges to $\rbvalue_{\mathit{max}}$.
By the quantitative semantics of LTL, convergence of robustness to $\rbvalue_{\mathit{max}}$ is equivalent to satisfaction of the inner LTL formula $\skolemized{\psi}$.
Therefore, the induced trace assignment maximizes the probability of satisfying $\skolemized{\varphi}$.

\paragraph{Lifting to the original \HyperLTL formula.}
Since Skolemization preserves satisfiability of \HyperLTL formulas, and existentially quantified traces are instantiated as Skolem witnesses depending only on preceding
universal traces, maximizing the robustness-based objective induced by
$\skolemized{\varphi}$ yields a tuple of policies that maximizes the probability
of satisfying the original \HyperLTL formula $\varphi$ in $\pomdp$.
Hence, the tuple of policies
$
\tupleof{\maxpolicy_i}_{i\in\existsof{}}
\orderedunion
\tupleof{\maxpolicy_j}_{j\in\forallsof{}}
$
optimizes the probability of satisfying $\varphi$ in $\pomdp$.
\end{proof}

\subsection{Proof of \cref{theorem:Qlearning}}
\label{proof:qlearning}

\begin{proof}
Given a POMDP $\pomdp$ and a \HyperLTL formula $\varphi$,
let $\skolemized{\varphi}$ be the Skolemized form of $\varphi$,
and let the robustness values
$\rbvalue\!\left(\zip(\cdot), \skolemized{\psi}\right)$
be used as the reward signal by the CTDE algorithm.
The CTDE algorithm optimizes a joint action--value function
$\jointqvalue(\jointhistory,\actiondec)$ that evaluates the expected discounted return under the
robustness-based reward signal.
Hence, an optimal joint policy $\jointpolicy^\star$ computed as:
$
\jointqvalue^{\jointpolicy^\star}(\jointhistory,\actiondec)
= \max_{\jointpolicy}\ \jointqvalue^{\jointpolicy}(\jointhistory,\actiondec),
$

We assume that $\jointqvalue$ satisfies the IGM property.
Then there exist individual action--value functions
$\tupleof{\Qvalue_i}_{\ell\in \{1,\ldots, n \}}$
such that for every joint history $\jointhistory=\tupleof{h_\ell}_{\ell\in \{1,\ldots, n \}}$,
\[
\arg\max_{a}\ \jointqvalue(\jointhistory,\actiondec)
=
\Big\langle
\arg\max_{a_\ell}\ \Qvalue_\ell(h_\ell,a_\ell)
\Big\rangle_{\ell\in\existsof{}\cup\forallsof{}}.
\]
Hence we get optimal decentralized policies from: 
\[
\Qvalue_\ell^{\policy_\ell^\star}(h_\ell,a_\ell)
= \max_{\policy}\ \Qvalue_\ell^{\policy_\ell}(h_\ell,a_\ell),
\quad\text{for each }\ell\in\existsof{}\cup\forallsof{}.
\]
By the IGM equality above, the induced joint greedy action
$\langle \maxpolicy_\ell(h_\ell)\rangle_\ell$ attains a joint maximizer of $\jointqvalue(\jointhistory,\cdot)$
for given $\jointhistory$; therefore the induced joint policy is optimal w.r.t.\ $\jointqvalue$.
It is important to note that we the input of the policies with the existential quantifier gets the history associated with 

By construction, the CTDE algorithm uses the robustness values
$\rbvalue\!\left(\zip(\cdot), \skolemized{\psi}\right)$ as the reward signal.
Hence, the learned joint action--value function $\jointqvalue$ optimizes the same objective as
\eqref{eq:opt}, i.e., it maximizes (over tuples of policies) the probability that the robustness-value
of the zipped traces w.r.t.\ $\skolemized{\psi}$ converges to $\rbvalue_{\max}$. 
Therefore, any policy tuple induced by the decentralized policies
$\tupleof{\maxpolicy_i}_{i\in\existsof{}}$ and $\tupleof{\maxpolicy_j}_{j\in\forallsof{}}$
is an optimizer of \eqref{eq:opt}, and thus optimizes the satisfaction probability of
$\skolemized{\varphi}$.
Note that our theoretical guarantees assume that, during decentralized execution, $f_{\texttt{enc-dec}}$ accurately reconstructs the relevant universally quantified trace prefixes required by the Skolem witnesses.
\end{proof}

\section{Continuation of Problem Statement Running Example}
\label{appendix:example}

Consider the POMDP in \cref{fig:pomdp}. 
We use the following \HyperLTL formula:
\[
\varphi_{\text{exp2}} \definedas \forall \tracevar_1 \forall \tracevar_2 . ~ \textsf{Dist}(\texttt{pos}_{\tracevar_1}, \texttt{pos}_{\tracevar_2})<3 ~~ \U ~~ (\texttt{pos}_{\tracevar_1} = i )
\]
$\ff$, corresponding to $\tracevar_1$, samples histories using policy $\policy_{1}$,
$\actobshist_{\tracevar_1}= \{ h_{1}^1, h_1^2 \}$:
\begin{align}
    h_1^1:  
\tupleof{a}\overset{\text{R}}{\rightarrow}\tupleof{b} 
\overset{\text{R}}{\rightarrow}
\tupleof{c, \textcolor{red}{c}}\overset{\text{U}}{\rightarrow} 
\tupleof{f, \textcolor{blue}{f} , \textcolor{red}{f}}\overset{\text{U}}{\rightarrow} \tupleof{i, \textcolor{red}{i}} 
\notag \\
    h_1^2:  
\tupleof{a}\overset{\text{R}}{\rightarrow}
\tupleof{b}\overset{\text{R}}{\rightarrow}
\tupleof{e}\overset{\text{U}}{\rightarrow}
\tupleof{h}\overset{\text{R}}{\rightarrow} \tupleof{i, \textcolor{red}{i}}  \notag ~~~~~~~~
\end{align}
$\med$, corresponding to $\tracevar_2$, samples histories using policy $\policy_{2}$,
$\actobshist_{\tracevar_2}= \{ h_{2}^1, h_2^2 \}$:
\begin{align}
    h_2^1:
\tupleof{a}\overset{\text{U}}{\rightarrow} 
\tupleof{d}\overset{\text{U}}{\rightarrow}
\tupleof{g, \textcolor{blue}{g}}\overset{\text{D}}{\rightarrow}
\tupleof{d}\overset{\text{R}}{\rightarrow}\tupleof{d} \notag ~~~~~\\ 
    h_2^2:
\tupleof{a}\overset{\text{U}}{\rightarrow}
\tupleof{b}\overset{\text{L}}{\rightarrow} \tupleof{e} 
\overset{\text{R}}{\rightarrow} \tupleof{f, \textcolor{blue}{f} , \textcolor{red}{f}}
\overset{\text{U}}{\rightarrow}\tupleof{i, \textcolor{red}{i}}  \notag
\end{align}
Assume that $\maptostate$ is path-consistent. 
Then we can construct the paths associated with $\actobshist_{\tracevar_1}$ and $\actobshist_{\tracevar_2}$ as
$\trajs_{\tracevar_1}=\{\arg\max_{\traj\in\trajs^*}\maptostate(h_1^1,\traj),\,\arg\max_{\traj\in\trajs^*}\maptostate(h_1^2,\traj)\}$ and
$\trajs_{\tracevar_2}=\{\arg\max_{\traj\in\trajs^*}\maptostate(h_2^1,\traj),\,\arg\max_{\traj\in\trajs^*}\maptostate(h_2^2,\traj)\}$, respectively.
We now compute the probability of satisfying $\varphi_{\text{exp2}}$ using $\trajs_{\tracevar_1}$ and $\trajs_{\tracevar_2}$; the evaluation ranges over all pairwise combinations of paths in $\trajs_{\tracevar_1}\times \trajs_{\tracevar_2}$:

\[
\tracesof{\tupleof{\{\arg\max_{\traj\in\trajs^*}\maptostate(h_1^1,\traj)\}, \{\arg\max_{\traj\in\trajs^*}\maptostate(h_2^1,\traj)\}}} \not\models \varphi_{\text{exp2}} ~~ \tracesof{\tupleof{\{\arg\max_{\traj\in\trajs^*}\maptostate(h_1^1,\traj)\}, \{\arg\max_{\traj\in\trajs^*}\maptostate(h_2^2,\traj)\}}} \models \varphi_{\text{exp2}}
\]
\[
\tracesof{\tupleof{\{\arg\max_{\traj\in\trajs^*}\maptostate(h_1^2,\traj)\}, \{\arg\max_{\traj\in\trajs^*}\maptostate(h_2^1,\traj)\}}} \models \varphi_{\text{exp2}} ~~~~~ \tracesof{\tupleof{\{\arg\max_{\traj\in\trajs^*}\maptostate(h_1^2,\traj)\}, \{\arg\max_{\traj\in\trajs^*}\maptostate(h_2^2,\traj)\}}} \models \varphi_{\text{exp2}}
\]
Out of the four possible pairs in $\trajs_{\tracevar_1}\times \trajs_{\tracevar_2}$, only three satisfy $\varphi_{\text{exp2}}$, yielding a satisfaction probability of $0.75$.

\section{More Technical Details for \cref{sec:algorithm}}
\label{sec:neuripsdetails}

\subsection{Details on \cref{sec:skolemization}}
\label{appnedix:image}
Let
$
\varphi =
\quant_1 \tracevar_1 \ldots \quant_n \tracevar_n.~ 
\psi(\tracevar_1,\ldots,\tracevar_n)
$
be a \HyperLTL formula. The Skolemized formula has the form:
{
\begin{align*}
\skolemized{\varphi} = 
\underbrace{\exists \skolemfunc_{i}(\tau_{i_1}, \ldots, \tau_{i_{|\forallsof{i}|}})}_{\text{for each } i \in 
\existsof{}}.
\underbrace{\spvertund{1.5ex}\forall \tracevar_{j}.}_{\text{for each }j \in \forallsof{}}\skolemized{\psi} \\[-7mm]
\end{align*}
}

Using the Transformation , we reformulate the problem in~\Cref{fig:hypole}.
We begin by defining the \emph{image} of a Skolem function $\skolemfunc_i$ as:
\begin{align}
	\imageof{\skolemfunc_i}&
	\definedas 
	\{ \skolemfunc_i(\trace_{i_1}, \ldots, \trace_{i_{|\forallsof{i}|}}) \mid \notag 
	t_{i_j} \in \tracesof{\hspace{-4mm}\bigcup_{h \in (\actobshist_{i_j} \sim 
	\distributionof{\policy_{i_j}})}
		\hspace{-4mm} \{ \arg\max_{\traj \in \trajs^*}
		\maptostate(h, \traj) \} },\ 
	j \in \forallsof{i} \}
	\notag
\end{align}

That is, $\imageof{\skolemfunc_i}$ is the set of all traces obtained by mapping the preceding 
universally quantified traces $\tau_{i_j}$, where each $\trace_{i_j}$ ranges over the trace set 
derived from the corresponding history set $\actobshist_{i_j}$.
Now, let us use $\orderedsets$ to indicate that the collection of trace sets is ordered with respect to 
their path indices.
That is, given two tuples of sets of traces $\Tr_1$ and $\Tr_2$, applying $\orderedsets$ yields an 
ordered tuple $\Tr_1 \orderedsets \Tr_2 \definedas 
\tupleof{\tracesof{\trajs_{\tracevar_x}}}_{x\in\{1,\ldots,n\}}$, where each 
$\tracesof{\trajs_{\tracevar_x}}$ belongs to either $\Tr_1$ or $\Tr_2$.
Returning to the problem in~\Cref{fig:hypole}, given a POMDP $\pomdp$ and a \HyperLTL formula 
$\varphi$, our goal is to compute: a tuple of Skolem witnesses 
$\tupleof{\skolemfunc_i}_{i\in\existsof{}}$, and a tuple of optimal policies 
$\tupleof{\maxpolicy_j}_{j\in\forallsof{}}$, such that:

\[
\tupleof{ \maxpolicy_j}_j
\in 
\bigg[
\underset{
\tupleof{
    {
        {\policy_{{j}}}
    }
}
}
{{\arg\max}} ~\prob 
\Big[ 
\tupleof{ 
\imageof{\skolemfunc_i}
}
\orderedsets	
\tupleof{ 
\tracesof{\hspace{-5mm}\bigcup_{h \in (\actobshist_{i_j} \sim \distributionof{\policy_{i_j}})}
\{ \arg\max_{\traj \in \trajs^*} \maptostate(h, \traj) \} }
}
\models
\skolemized{\varphi}
\Big]
\bigg]_{{i \in \existsof{}}, {j \in \forallsof{}}} 
\]

The meaning of above \modif{formula} is that the policy tuple $\tupleof{\maxpolicy_j}$ maximizes the 
probability that the ordered collection consisting of (1) the generated traces of the universal 
quantifiers $\tupleof{\tracesof{\trajs_{\tau_j}}}_{j\in\forallsof{}}$ and (2) the Skolem witness for all 
existential quantifiers $\tupleof{\imageof{\skolemfunc_i}}_{i\in\existsof{}}$ jointly satisfies 
$\skolemized{\varphi}$.
In the reformulated problem, policies are synthesized solely for universally quantified traces, 
whereas Skolem functions are learned for existentially quantified traces to serve as witnesses of 
optimality.

\subsection{Details on \cref{sec:quantsemantics}}
\label{appendix:quant}

We reformulate the \HyperLTL satisfaction problem as an optimization problem under its quantitative semantics, where the Skolemized formula $\skolemized{\varphi}$ is evaluated on tuples of histories $\tupleof{h_1,\ldots,h_n}$ sampled from the POMDP $\pomdp$.

\paragraph{Robustness for a Single Trace.}
Let $\realnum$ denote the set of real numbers and let $\Psi$ denote the set of all \LTL formulas.
Let $\predicatefunc:2^\AP \rightarrow \mathbb{R}$ be a valuation function that assigns real values 
to sets of atomic propositions.
Let $\mathbb{R}$ be the set of real numbers, $\Psi$ the set of all \LTL formulas, and $\predicatefunc : 2^{\AP} \rightarrow \mathbb{R}$ be a \emph{valuation function} assigning real values to sets of atomic propositions.
For a state $\state \in \states$ of $\pomdp$, quantitative semantics are defined by predicates of the form $\predicatefuncof{\labelsof{\state}} < \constant$, where $\constant$ is a user-specified threshold (see~\Cref{fig:quantsemantics}).
Different values of $\constant$ can be assigned to different segments of the inner \LTL formula to emphasize the relative priority of those segments.
The robustness function $\rbvalue:\tracesof{\trajs^\finite}\times\Psi \rightarrow \mathbb{R}$ assigns a real-valued score to a finite trace and an \LTL formula, bounded by $\rbvalue_{\mathit{max}}$ and $\rbvalue_{\mathit{min}}$. 
For a given trace, higher $\rbvalue$ values correspond to greater robustness in satisfying $\psi$, while lower values indicate weaker satisfaction or possible violation.
Formally, given an \LTL formula $\psi$ and a POMDP $\pomdp$, we formulate the problem of synthesizing a policy $\maxpolicy$ as:
Formally, given an \LTL formula $\psi$ and a POMDP $\pomdp$, we formulate the policy synthesis problem as:

\[
    \maxpolicy \in 
\underset{
	\policy
}
{{\arg\max}} 
\underset{h \sim \distributionof{\policy}}
{\prob}
\Big[ \notag 
\rbvalueof{\traceoff{(\arg\max_{\traj \in \trajs^*} \maptostate(h, \traj))_{[0:k]}}, \psi} \converge \rbvalue_{\mathit{max}} 
\Big],
\]
\hspace*{-1mm}where $\converge$ means convergence. 
I.e., $\maxpolicy$ maximizes the probability that paths induced by histories under $\policy$ satisfy $\psi$.

\paragraph{Robustness for a Tuple of Traces.}

Next, to compute robustness over multiple (universally quantified) traces, we define a $\zip$ operator as follows:
that performs pointwise aggregation over a tuple of traces.
Given a tuple of finite traces
$\tupleof{\trace_\ell}_{\ell\in\setof{1,\ldots,n}}$, where each
$\trace_\ell = \trace_\ell(0)\trace_\ell(1)\cdots\trace_\ell(m_\ell)$,
we define their \emph{zipped trace} as:
\[
\zipof{\tupleof{\trace_\ell}_{\ell\in\setof{1,\ldots,n}}}
\definedas
\tupleof{\trace_\ell(0)}_{\ell\in\setof{1,\ldots,n}} \cdots
\tupleof{\trace_\ell(k)}_{\ell\in\setof{1,\ldots,n}}
\]
Given an \LTL formula $\psi$,
a tuple of paths $\tupleof{\traj_1, \traj_2, \ldots, \traj_n}$ is more likely to satisfy $\psi$ if the robustness value of
$
\zip\big(\tupleof{\traceof{\trajfromto{\traj_1}{0}{k_1}}, \traceof{\trajfromto{\traj_2}{0}{k_2}}, \ldots, \traceof{\trajfromto{\traj_n}{0}{k_n}}}\big)
$
converges to $\rbvalue_{\mathit{max}}$ with respect to $\psi$ for some $k_1, \ldots, k_n$, where
$0 \le k_\ell \le \sizeof{\traj_\ell}$ for each $1 \le \ell \le n$.

Thus, the optimization problem of computing a tuple of policies
$\tupleof{\maxpolicy_1, \maxpolicy_2, \ldots, \maxpolicy_n}$
that maximizes robustness can be stated as:
$$\tupleof{\maxpolicy_\ell}_{\ell}
\in 
\bigg[ 
\underset{
\tupleof{
\policy_{{\ell}}
}
}
{{\arg\max}}~
{\prob}
\Big[ 
\rho \big(
\zip 
\tupleof{
\traceof{(\arg\max_{\traj \in \trajs^*} \maptostate(h_{\ell} \sim \distributionof{\policy_{\ell}} , \traj))_{[0:k_l]}}
}
\big), 
\psi \big)
\converge 
\rbvalue_{\mathit{max}}
\Big]
\bigg]_{\ell \in \setof{1,...n}}$$
This transformation highlights that \LTL specifications are implicitly universally quantified in the single-trace setting.

\paragraph{Robustness for Skolemized HyperLTL.}

We now address the optimization problem induced by an alternating \HyperLTL formula.
Let $\tupleof{\trace_i}_{i\in I}$ and $\tupleof{\trace_j}_{j\in J}$ be two trace tuples, where
$I \cup J = \setof{1,\ldots,n}$ and $I \cap J = \emptyset$.
To preserve index order when combining them, we use $\orderedunion$, which produces a single tuple ordered by path indices.
In particular, applying $\orderedunion$ produces a tuple
$\traceof{\traj_1} < \traceof{\traj_2} < \cdots < \traceof{\traj_n}$,
where $\traceof{\traj_x} < \traceof{\traj_y}$ denotes $x<y$.
Based on this construction, satisfaction of the inner \LTL body $\psi$ is defined over tuples of histories.
For each $i\in\existsof{}$ and $j\in\forallsof{}$, a tuple $\tupleof{h_1,\ldots,h_n}$ satisfies $\psi$ if and only if:

\begin{align*}
\rbvalue
\Big(
	\zip
	\big(
	\tupleof{\traceof{\arg\max_{\traj \in \trajs^*} \maptostate(h_{i}, \traj))_{[0:k_i]}}} \;
	\orderedunion 
	\tupleof{\traceof{\arg\max_{\traj \in \trajs^*} \maptostate(h_{j}, \traj))_{[0:k_j]}}}
	\big), \notag  
	\skolemized{\psi}
\Big)
\converge \rbvalue_{\mathit{max}}
\notag
\end{align*}

Therefore, we formulate the optimization problem for the Skolemized \HyperLTL formula as follows:

\begin{align*}
    \tupleof{\maxpolicy_i}
\orderedunion
\tupleof{\maxpolicy_j}
\in &
\Bigg[
\underset{
	\tupleof{\policy_i}
	\orderedunion
	\tupleof{\policy_j}
}
{{\arg\max}} 
~{\prob}
\Big[
\rbvalue
\Big(
\zip
\big(
\tupleof{\traceoff{(\arg\max_{\traj \in \trajs^*} \maptostate(h_i \sim \distributionof{\policy_{i}} , 
\traj))_{[0:k_i]}}\big} \;
\orderedunion \, 
\label{eq:opt}
\\
& ~~~~~~~~~~~~~~~~~~~~~~~~~~~~~~~~~~
\tupleof{\traceoff{(\arg\max_{\traj \in \trajs^*} \maptostate(h_j \sim \distributionof{\policy_{j}} , \traj))_{[0:k_j]}}\big}
\big),
\skolemized{\psi}
\Big)
\converge 
\rbvalue_{\mathit{max}}
\Big]\Bigg]_{{i \in \existsof{}}, {j \in \forallsof{}}}
\end{align*}

\begin{figure*}[t]
	\begin{center}
		\renewcommand{\arraystretch}{1.5}
		\begin{bluebox2}
				\input{robustnessLTL}
		\end{bluebox2}	
	\end{center}

	\caption{Quantitative semantics for \LTL, adapted from~\cite{hsu2025hyprl, li2017reinforcement}.}
	\label{fig:quantsemantics}
\end{figure*}

%% file: robustnessltl.tex
\renewcommand{\arraystretch}{1}
\[
	\begin{array}{llll}
	\centering
	&\rbvalueof{\traceof{\trajfromto{\traj}{\ell}{k}}, \psi} 			
	&=& 
		\rbvalue_{\mathit{min}} \text{ if } \traceof{ \trajfromto{\traj}{\ell}{..} } = \epsilon \text{ and }
		\rbvalueof{\traceof{\trajfromto{\traj}{\ell}{k}}, \psi} \text{ otherwise.}
	\\
	&
	\rbvalueof{\traceof{\trajfromto{\traj}{\ell}{k}}, \tru} 			
	&=& \rbvalue_{\mathit{max}}
	\\
	&\rbvalueof{\traceof{\trajfromto{\traj}{\ell}{k}}, 
	\predicatefuncof{ { \labelsof{\state_\ell}<\constant}} } 			
	&=& \constant - \predicatefuncof{\labelsof{\state_\ell}}\\ 
	&
	\rbvalueof{\traceof{\trajfromto{\traj}{\ell}{k}}, \neg\psi} 	 
	&=& - \rbvalueof{\traceof{\trajfromto{\traj}{\ell}{k}}, \psi}\\
	&
	\rbvalueof{\traceof{\trajfromto{\traj}{\ell}{k}}, \X\psi} 			
	&=& \rbvalueof{\traceof{\trajfromto{\traj}{\ell+1}{k}}, \psi} ~~\text{if} ~~ (k > \ell).\\
	&\rbvalueof{\traceof{\trajfromto{\traj}{\ell}{k}}, \G\psi} 			
	&=& \minamong{i\in[\ell,k) }\rbvalueof{\traceof{\trajfromto{\traj}{i}{k}}, \psi}\\  
	&
	\rbvalueof{\traceof{\trajfromto{\traj}{\ell}{k}}, \F\psi} 			
	&=& \maxamong{i \in[\ell,k) }\rbvalueof{\traceof{\trajfromto{\traj}{i }{k}}\psi}  \\
	&\rbvalueof{\traceof{\trajfromto{\traj}{\ell}{k}}, \psi_1 \land \psi_2} 
	&=& ~\minimum\big(\rbvalueof{\traceof{\trajfromto{\traj}{\ell}{k}},  \psi_1} 
	\rbvalueof{\traceof{\trajfromto{\traj}{\ell}{k}}, \psi_2}\big) \\
	&\rbvalueof{\traceof{\trajfromto{\traj}{\ell}{k}}, \psi_1 \lor \psi_2} 
	&=& ~\maximum\big(\rbvalueof{\traceof{\trajfromto{\traj}{\ell}{k}}, \psi_1} 
	\rbvalueof{\traceof{\trajfromto{\traj}{\ell}{k}}, \psi_2}\big) \\
	&\rbvalueof{\traceof{\trajfromto{\traj}{\ell}{k}}, \psi_1 \until \psi_2} 
	&=& 
	\maxamong{ i \in[\ell,k) }
	\Big(
	\minimum
	\Big(
	\rbvalueof{\traceof{\trajfromto{\traj}{i}{k}}, \psi_2}, 
	\minamong{ j \in[\ell, i ) }\rbvalueof{\traceof{\trajfromto{\traj}{j}{i}}, \psi_1}
	\Big)
	\Big) 
\end{array}
\]

%% file: exp-appendix.tex
\section{Experimental Setup and Implementation Details}
\label{appendix:exp-details}
In this section, we detail the implementation, architecture, and hyperparameters used in our learning process.
\subsection{Details on MARL used in \method}
In our experiments, we use the implementations of QMIX, VDN, QTRAN, and IQL provided by \href{https://github.com/oxwhirl/pymarl}{\texttt{https://github.com/oxwhirl/pymarl}}.
For QMIX, we set \texttt{epsilon\_start}=1.0, \texttt{epsilon\_finish}=0.05, \texttt{epsilon\_anneal\_time}=50000, \texttt{buffer\_size}=5000, \texttt{target\_update\_interval}=200, \texttt{mixing\_embed\_dim}=64, \texttt{opt\_loss}=1, and \texttt{nopt\_min\_loss}=0.1.
For VDN, we set \texttt{epsilon\_start}=1.0, \texttt{epsilon\_finish}=0.05, \texttt{epsilon\_anneal\_time}=50000, \texttt{buffer\_size}=5000, and \texttt{target\_update\_interval}=200.
For QTRAN, we set \texttt{epsilon\_start}=1.0, \texttt{epsilon\_finish}=0.05, \texttt{epsilon\_anneal\_time}=50000, \texttt{buffer\_size}=5000, \texttt{target\_update\_interval}=200, \texttt{mixing\_embed\_dim}=64, \texttt{opt\_loss}=1, and \texttt{nopt\_min\_loss}=0.1.
Finally, for IQL, we set \texttt{epsilon\_start}=1.0, \texttt{epsilon\_finish}=0.05, \texttt{epsilon\_anneal\_time}=50000, \texttt{buffer\_size}=5000, and \texttt{target\_update\_interval}=200.

\subsection{Details on $\fseq$, $\skolemfunc$, $\traceToHist$ used in \method}
$\fseq$ is a feedforward \texttt{encoder--decoder} (autoencoder) model implemented in \textsf{PyTorch}.
Both the encoder and decoder are three-layer MLPs composed of \texttt{Linear} layers.
We use \texttt{ReLU} activations after each hidden layer, and apply a final \texttt{Tanh} activation at the decoder output to bound predictions to $[-1,1]$, matching the normalized target range.
The first hidden layer has width equal to the observation dimension, and the second hidden layer has 256 units; the remaining layer sizes follow this encoder--decoder symmetry.
We optimize the model using Adam with learning rate $10^{-3}$ and train it by minimizing mean squared error (MSE).

To train $\fseq$ model, we maintain a buffer $\buffer$ containing the $10{,}000$ most recent agent with existential quantifier observations and their corresponding global states at the same timestep.
After each episode, we uniformly sample a mini-batch of size $32$ from the pair buffer and perform one gradient update step.
To reduce unnecessary computation, we apply a decaying update schedule based $\beta$ on the current training loss.
When the loss exceeds $9\times 10^{-5}$, we update $\fseq$ every $50$ environment steps.
When the loss is between $9\times 10^{-5}$ and $9\times 10^{-6}$, we update every $100$ steps.
Once the loss drops below $9\times 10^{-6}$, we update every $1000$ steps.

Next, for each agent with an existential quantifier $i \in \existsof{}$, we use an \texttt{RNN} to approximate
$\traceToHist(\skolemfunc_i(\cdot))$ and obtain a history representation $\hat{h}_i$.
Basically, for agents associated with existential quantifiers, instead of feeding only their local observations $o_i$ into the \texttt{RNN} module, we first predict the global state sequences of the preceding universally quantified agents, denoted by $(\hat{\state}_{i_1[0:k]}, \ldots, \hat{\state}_{i_{|\forallsof{i}|}[0:k]})$. This predicted information is then incorporated into the input of the \texttt{RNN}, producing a history representation $\hat{h}_i$. The resulting representation contains information about both agent $i$ and the preceding universally quantified agents.
We then use $\hat{h}_i$ to construct the agent's utility function and decentralized policy $\pi_i$.
For the remaining components of CTDE, we follow the standard \texttt{PyMARL} implementation.
It is important to note that the \texttt{RNN} approximating $\traceToHist(\skolemfunc_i(\cdot))$ is trained jointly with CTDE.

\subsection{Details on Experimental Setup and Training}
We ran training on three machines. 
The first machine has an AMD EPYC 7742 64-core CPU and three NVIDIA RTX A6000 GPUs. 
The second is a shared server with an Intel Xeon Platinum 8268 CPU and four NVIDIA Quadro RTX 8000 GPUs; on this machine, CPU resources are limited due to sharing (GPU resources are not).
The third machine has an AMD EPYC 7742 64-core CPU and three NVIDIA RTX PRO 6000 Blackwell Max-Q Workstation Edition GPUs.
For SMAC and MessySMAC, we train each agent for $10$k environment steps and evaluate every checkpoint over 32 test episodes. 
All reported SMAC and MessySMAC results are obtained in \texttt{test\_mode}.
For WildFire, we train each agent for $5$k environment steps and evaluate every checkpoint over 30 test episodes.
All SMAC and Wildfire experiments were conducted over ten independent runs with different random seeds, while MessySMAC experiments were conducted over five independent runs.

\begin{table*}[t]
\caption{SMAC maps used in evaluation, including unit configurations and \HyperLTL formulas used in our experiments.}
\label{tab:smac_maps}
\centering
\begin{small}
\begin{tabular}{l l l l c}
\toprule
Map Name & Ally Units & Enemy Units & Formula & Quantifiers \\
\midrule
3m & 3 Marines & 3 Marines  & $\focus$, $\defense$, $\bad$ & $\forall^*$ \\
8m & 3 Marines  & 3 Marines  & $\focus$ & $\forall^*$ \\
2s3z & 2 Stalkers \& 3 Zealots & 2 Stalkers \& 3 Zealots & $\focus$ & $\forall^*$ \\
3s5z & 3 Stalkers \& 5 Zealots & 3 Stalkers \& 5 Zealots & $\focus$ & $\forall^*$ \\
MMM & 1 Med \& 2 Marauders \& 7 Marines & 1 Med \& 2 Marauders \& 7 Marines & $\medivac$ & $\forall^*$, $\forall^*\exists$ \\
\midrule
5m\_vs\_6m & 5 Marines & 6 Marines & $\focus$,$\kite$ & $\forall^*$ \\
MMM2 & 1 Med, 2 Marauders \& 7 Marines & 1 Med, 2 Marauders \& 8 Marines & $\medivac$ & $\forall^*$, $\forall^*\exists$ \\
\midrule
bane\_vs\_bane & 20 Zerglings \& 4 Banelings & 20 Zerglings \& 4 Banelings & $\focus$ & $\forall^*$ \\
corridor & 6 Zealots & 24 Zerglings &  $\corridor$ & $\forall^*$ \\
\bottomrule
\end{tabular}
\end{small}
\end{table*}

\section{Details on SMAC and MessySMAC Benchmark}
\label{app:smac_messy}
We evaluate on the StarCraft~II Multi-Agent Challenge (SMAC) benchmark \href{https://github.com/oxwhirl/smac}{\texttt{https://github.com/oxwhirl/smac}}. 
\cref{tab:smac_maps} lists the scenarios used in our experiments, together with the \HyperLTL formulas we evaluate and their corresponding quantifier structures. 
For baseline comparisons, we use the shaped reward provided by SMAC, which combines hit-point damage, enemy kills, and an additional bonus for winning the battle. 
SMAC also includes a sparse reward ($+1$ for winning and $-1$ for losing an episode); however, in our preliminary experiments the sparse reward performed substantially worse than the shaped reward. 
Therefore, we use the shaped reward for all baseline results.

We mostly use the same configuration for MessySMAC. However, this benchmark adds observation stochasticity $\phi$ and randomized initialization with $K$ random steps before each episode starts. For the \texttt{3m} map, we use $(K=10, \phi=0.15)$. For the \texttt{8m}, \texttt{2s3z}, \texttt{3s5z}, and \texttt{MMM} maps, we use $(K=12, \phi=0.15)$.

\section{Details on Wildfire Benchmark}
\label{appendix:wildfire}
We extend the Wildfire scenario from~\citet{hsu2025hyprl} to a partially observable setting.
The environment contains two agent types: a firefighter drone $\ff$ \raisebox{-0.15em}{\includegraphics[height=1.2em]{img/ff.png}} and a medical drone $\med$ \raisebox{-0.15em}{\includegraphics[height=1.2em]{img/medic.png}} (see \cref{fig:pomdp}). 
The agents must extinguish fires and rescue victims within a fixed time while maintaining a safe separation distance.
In this setting, each agent observes only entities within a fixed sight range.
We use the maps listed in~\Cref{tab:wild_maps}.
The global state includes all agent features and object (fire/victim) features, the last action, and the time step. 
Agent and object features are represented by centered coordinates: $(x-\texttt{center}_x,\,y-\texttt{center}_y)$, normalized by the maximum horizontal and vertical distances, respectively. 
The last-action vector encodes each agent's previous action.
Each agent's observation consists of its move features, own features, the time step, and the features of agents and objects within its sight range.
Move features specify the set of available actions.
The action space is $\{\texttt{Up}, \texttt{Down}, \texttt{Left}, \texttt{Right}\}$.
For each observed entity, we include relative position (normalized by sight range), distance to the entity (normalized by sight range), the entity type ID, and their last action.

We introduce three reasonable baseline reward functions $R_{\text{Wild\_1}}$, $R_{\text{Wild\_2}}$, $R_{\text{Wild\_3}}$. The functions are as follows:
\[
R_{\text{Wild\_1}} = 
\begin{cases}
20 & \text{Victim Saved}\\
20 & \text{Fire Ext}\\
-10 & \text{Out of Range}\\
-10 & \text{\med in Fire}
\end{cases}
~~~~
R_{\text{Wild\_2}} = 
\begin{cases}
20 & \text{Victim Saved}\\
20 & \text{Fire Ext}\\
-10 & \text{Out of Range}\\
-10 & \text{\med in Fire} \\
+100 & \text{Mission Accomplish}
\end{cases}
~~~~
R_{\text{Wild\_3}} = 
\begin{cases}
20 & \text{Victim Saved}\\
20 & \text{Fire Ext}\\
-10 & \text{Out of Range}\\
-10 & \text{\med in Fire} \\
+100 & \text{Mission Accomplish}\\
-50 & \text{Mission failed}\\
\end{cases}
\]

\begin{table*}[t!]
\caption{WildFire maps used in evaluation, including agents configurations and \HyperLTL formulas used in our experiments.}
\label{tab:wild_maps}
\centering
\begin{tabular}{l l l l l}
\toprule
Map Name & Grid Size& Agents & Objects & Formula  \\
\midrule
5P\_1F1V & $5\times5$& 1 \ff, 1 \med & 1 Victim 1 Fire Zone  & $\wild$ \\
5P\_3F2V & $5\times5$& 1 \ff, 1 \med   & 2 Victim 3 Fire Zone  & $\wild$ \\ 
8P\_3F2V & $8\times8$& 1 \ff, 1 \med  & 2 Victim 3 Fire Zone  & $\wild$ \\
10P\_3F2V & $10\times10$ & 1 \ff, 1 \med  & 2 Victim 3 Fire Zone  & $\wild$ \\

\bottomrule
\end{tabular}
\end{table*}

\section{\HyperLTL Formulas}
\label{sec:formula}
We provide all \HyperLTL formulas used in the SMAC and MessySMAC experiments, namely $\focus$, $\medivac$, $\kite$, $\corridor$, $\defense$, and $\bad$, as well as $\wild$ for the WildFire benchmark.

\begin{align}
    \focus \definedas \overbrace{\forall \tracevar_1, \ldots, \tracevar_n}^{\text{Ally Agents}}  ~. ~\bigg[ \Big( \overbrace{\big( \big(\motifdist(\pos_{\tracevar_1},\posenm{1}) < \shootrate_{\tracevar_1} \wedge \ldots \land \motifdist(\pos_{\tracevar_n},\posenm{1}) < \shootrate_{\tracevar_n} \big)}^{\text{Focused Shooting}} \lor \notag \\ \cdots \lor\big(\motifdist(\pos_{\tracevar_1},\posenm{m}) < \shootrate_{\tracevar_1} \wedge \ldots \land \motifdist(\pos_{\tracevar_n},\posenm{m}) < \shootrate_{\tracevar_n} \big) \big)  \wedge \notag \\ 
    \big( \underbrace{(\healthenm{1} < \healthenmprev{1})}_\text{Damaging Enemies Health/Shield Bar} \vee \cdots \vee (\healthenm{m} < \healthenmprev{m}) \big)\Big) \notag \\ \U \underbrace{\noenem>\noenemprev}_{\text{Eliminating Enemies}} \bigg]  ~~\mathcal{U}~~ \win \notag 
\end{align}
\begin{align}
    \medivac \definedas \overbrace{\forall \tracevar_1, \ldots, \tracevar_{n-1}}^{\text{Ally Agents}}~\overbrace{\forall \tracevar_{n} / \exists \tracevar_{n} }^{\text{Medivac}}  ~.~~~~~~~~~~~~~~~~~~~~~~~~~~~~~~~~~~~~~~~~~~~~~~~~~~~~~~~~~~~~~~~~~~~~~~~~~~~~~~~~~~~~~~~~~~~~~~~~~~~~~~~~~~~~~~~~~~~~~~~~~~~~~~~~~~~~~~~~~
    \notag\\\bigg[ \Big( \overbrace{\big( \big(\motifdist(\pos_{\tracevar_1},\posenm{1}) < \shootrate_{\tracevar_1} \wedge \ldots \land \motifdist(\pos_{\tracevar_{n-1}},\posenm{1}) < \shootrate_{\tracevar_{n-1}} \big)}^{\text{Focused Shooting}} \lor \notag \\ \cdots \lor\big(\motifdist(\pos_{\tracevar_1},\posenm{m}) < \shootrate_{\tracevar_1} \wedge \ldots \land \motifdist(\pos_{\tracevar_{n-1}},\posenm{m}) < \shootrate_{\tracevar_{n-1}} \big) \big)  \wedge \notag \\ 
   \overbrace{\big( \motifdist(\pos_{\tracevar_n},\textsf{Min\_Health}(\tracevar_1,\ldots,\tracevar_{n-1})) < \texttt{h\_range}_{\tracevar_n}\big)}^{\text{Medivac Close to Agent With Min Health}} \land\notag \\
   \big( \underbrace{(  \sum_{i=1}^{n-1} \texttt{health\_prev}_{\tracevar_i} - \sum_{i=1}^{n-1} \texttt{health}_{\tracevar_i}) < ( \sum_{i=1}^{m} \healthenmprev{i}) - \sum_{i=1}^{m} \healthenm{i} }_\text{Damaging Enemies Health/Shield Bar While Preserving Allies Health/Shield Bar} \big) \Big) \notag \\ 
    \U \underbrace{\noenem>\noenemprev}_{\text{Eliminating Enemies}} \bigg]  ~~\mathcal{U}~~ \win \notag 
\end{align}
\begin{align}
    \kite \definedas \overbrace{\forall \tracevar_1, \ldots, \tracevar_n}^{\text{Ally Agents}}  ~. ~\bigg[ \Big( \overbrace{\big(  \neg \texttt{ready}_{\tracevar_1}\lor (\bigwedge_{i=1} ^m\motifdist(\pos_{\tracevar_1},\posenm{i})<\texttt{s\_range\_i}) \big)}^{\text{Kiting}} \land \notag  ~~~~~~~~~~~~~~~~~~~~~~~~~~~~~~~~~~~~~~~~\\ \ldots \land \big(  \neg \texttt{ready}_{\tracevar_n}\lor (\bigwedge_{i=1} ^m\motifdist(\pos_{\tracevar_n},\posenm{i})<\texttt{s\_range\_i}) \big) \land \notag ~~~~~~~~~~~~~~~~~~~~~~~~~~~~~~~~~~\\ 
    \big( \underbrace{(\healthenm{1} < \healthenmprev{1})}_\text{Damaging Enemies Health/Shield Bar} \vee \cdots \vee (\healthenm{m} < \healthenmprev{m}) \big)\Big) \notag \\ \U \underbrace{\noenem>\noenemprev}_{\text{Eliminating Enemies}} \bigg]  ~~\mathcal{U}~~ \win \notag 
\end{align}
\begin{align}
    \corridor \definedas \overbrace{\forall \tracevar_1, \ldots, \tracevar_n}^{\text{Ally Agents}}  ~. ~\bigg[ \Big( \overbrace{\big(\motifdist(\pos_{\tracevar_1}, \texttt{pos\_choke\_point})<\delta \land \ldots \land \motifdist(\pos_{\tracevar_n}, \texttt{pos\_choke\_point})<\delta \big) }^{\text{Defend in Choke Point}}  \U \notag \\
    \big( \underbrace{(\healthenm{1} < \healthenmprev{1})}_\text{Damaging Enemies Health/Shield Bar} \vee \cdots \vee (\healthenm{m} < \healthenmprev{m}) \big)\Big) \notag \\ \U \underbrace{\noenem>\noenemprev}_{\text{Eliminating Enemies}} \bigg]  ~~\mathcal{U}~~ \win \notag 
\end{align}
\begin{align}
    \defense \definedas \overbrace{\forall \tracevar_1, \ldots, \tracevar_n}^{\text{Ally Agents}}  ~. ~ \overbrace{ \G \big( \bigwedge_{i=1}^{n}\bigwedge_{j=1}^{m}(\motifdist(\pos_{\tracevar_i},\posenm{j}) > \obsrate{j}) \big)}^{\text{Running Away from Enemies}} \land \notag 
   \underbrace{ \G \big( \bigwedge_{i=1}^{n} (\texttt{health}_{\tracevar_i} > \delta \big)}_\text{Preserve Allies Health/Shield Bar } \notag 
\end{align}
\begin{align}
    \bad \definedas \overbrace{\forall \tracevar_1, \ldots, \tracevar_n}^{\text{Ally Agents}}  ~. ~\bigg[ \Big( \overbrace{\big( \big(\motifdist(\pos_{\tracevar_1},\posenm{1}) < \shootrate_{\tracevar_1} \wedge \ldots \land \motifdist(\pos_{\tracevar_n},\posenm{1}) < \shootrate_{\tracevar_n} \big)}^{\text{Focused Shooting}} \lor \notag \\ \cdots \lor\big(\motifdist(\pos_{\tracevar_1},\posenm{m}) < \shootrate_{\tracevar_1} \wedge \ldots \land \motifdist(\pos_{\tracevar_n},\posenm{m}) < \shootrate_{\tracevar_n} \big) \big)   \U \notag \\ 
    \underbrace{\noenem>\noenemprev}_{\text{Eliminating Enemies}} \bigg]  ~~\mathcal{U}~~ \win \notag 
\end{align}

 \begin{align}
    \wild \definedas \overbrace{\forall \tracevar_{1}}^{\med}. \underbrace{(\forall \tracevar_{2} /\exists \tracevar_{2} )}_{\ff}  ~. ~ \bigg[\bigg(
   \Big( \big(\overbrace{(\motifdist(\texttt{pos}_{\tracevar_2} , \texttt{pos\_fire\_1})< \texttt{prev\_dist\_FF\_fire\_1}_{\tracevar_2})}^{\text{FF Get Close to Fire}}~ \lor ~~~~~~~~~~~~~~~~~~~~~~~~~~~~~ \notag \\ 
   \ldots \lor  (\motifdist(\texttt{pos}_{\tracevar_2} , \texttt{pos\_fire\_n})< \texttt{prev\_dist\_fire\_n}_{\tracevar_2})\big) \U \texttt{Ex\_fire}_{\tracevar_2} \Big) ~\land \notag \\ 
   \Big( \big((\motifdist(\motifdist(\texttt{pos}_{\tracevar_1} , \texttt{pos\_Vic\_1})< \texttt{prev\_dist\_Vic\_1}_{\tracevar_1})~ \lor \notag \\ 
   \ldots \lor  (\motifdist(\motifdist(\texttt{pos}_{\tracevar_1} , \texttt{pos\_Vic\_m})< \texttt{prev\_dist\_Vic\_m}_{\tracevar_1})\big) \U \texttt{Save\_Vic}_{\tracevar_1}\Big)
\bigg) ~ \land  \notag\\  
\motifdist(\texttt{pos}_{\tracevar_1}, \texttt{pos}_{\tracevar_2})<\delta ~~\land ~~ \bigwedge_{i=0}^{n}\neg \texttt{pos\_Fire\_i}_{\tracevar_1} \bigg] \U \texttt{Mission\_Accomplished}  \notag
\end{align}

\section{Additional Results}

\subsection{\method vs. Shaped Rewards}
\label{sec:shape_app}
\begin{figure*}[]
    \centering
    \begin{subfigure}[t]{\textwidth}
        \centering
        \includegraphics[width=0.99\linewidth]{img/legend_only.pdf}
    \end{subfigure}

    \begin{subfigure}[t]{0.33\textwidth}
        \centering
        \includegraphics[width=\linewidth]{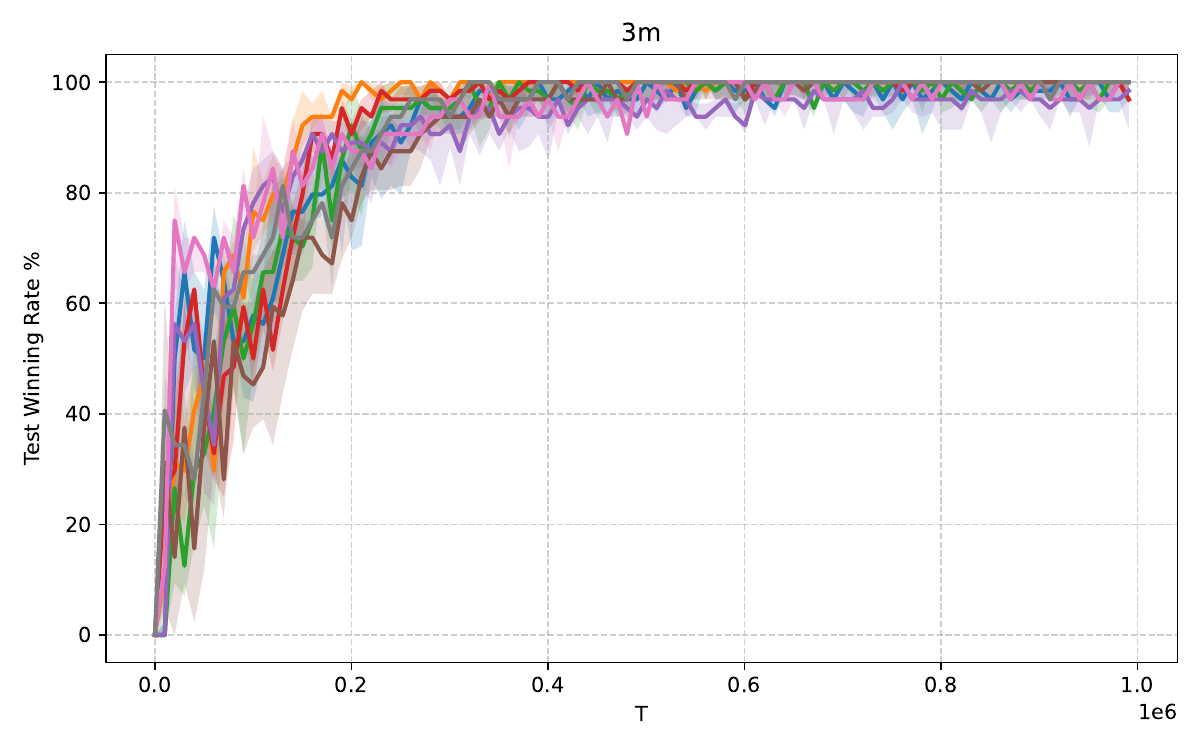}
        \caption{3m}
        \label{fig:3m}
    \end{subfigure}\hfill
    \begin{subfigure}[t]{0.33\textwidth}
        \centering
        \includegraphics[width=\linewidth]{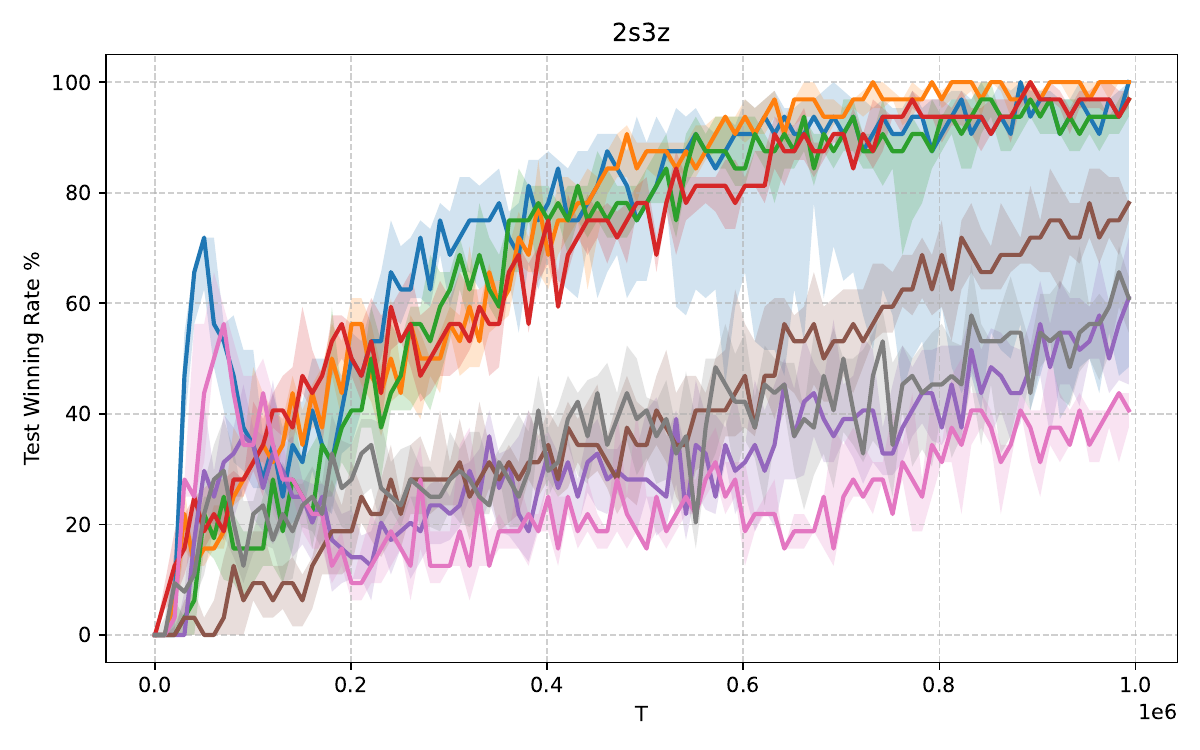}
        \caption{2s3z}
        \label{fig:2s3z}
    \end{subfigure}\hfill
    \begin{subfigure}[t]{0.33\textwidth}
        \centering
        \includegraphics[width=\linewidth]{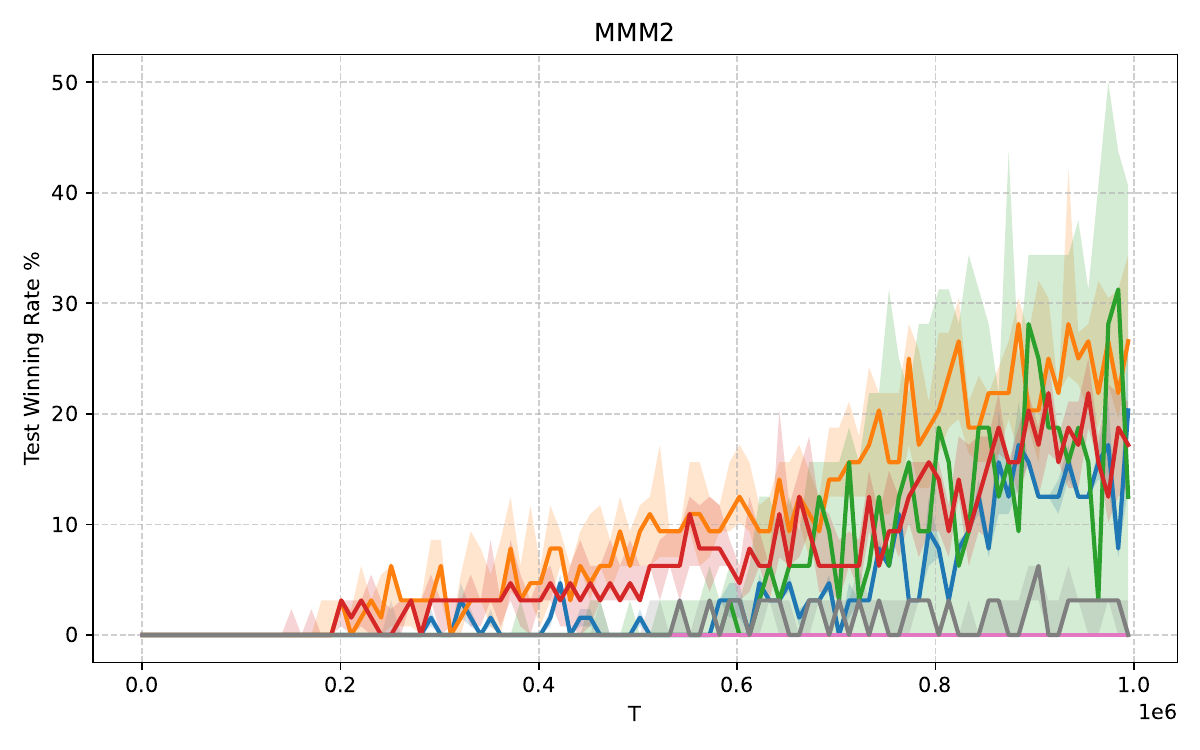}
        \caption{MMM2}
        \label{fig:MMM2}
    \end{subfigure}

    \caption{Learning curves on remaining SMAC maps, showing median test win rate with 25-75\% percentile.}
    \label{fig:winnig_main_rest}
\end{figure*}

\begin{figure*}[t!]

    \centering

        \begin{subfigure}[t]{\textwidth}
        \centering
        \includegraphics[width=0.99\linewidth]{img/legend_only.pdf}
    \end{subfigure}

    \begin{subfigure}[t]{0.32\textwidth}
        \centering
        \includegraphics[width=\linewidth]{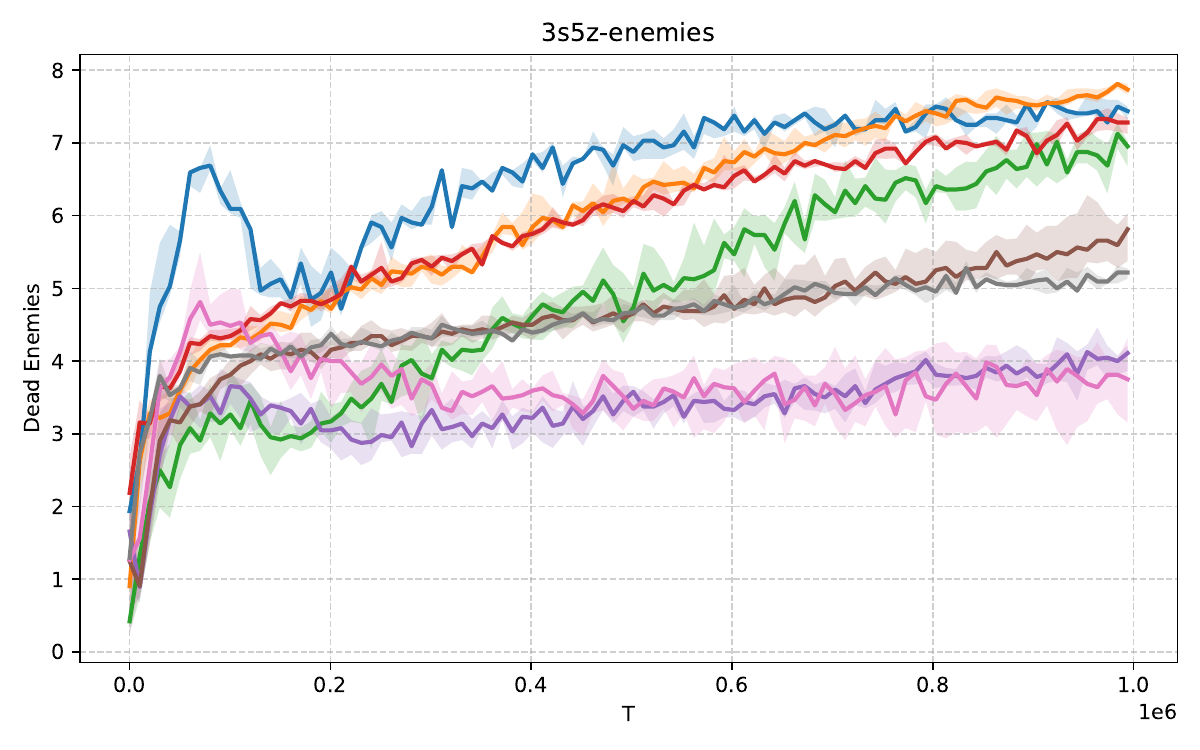}
        \caption{3s5z}
        \label{fig:3s5z-enemies}
    \end{subfigure}\hfill
    \begin{subfigure}[t]{0.32\textwidth}
        \centering
        \includegraphics[width=\linewidth]{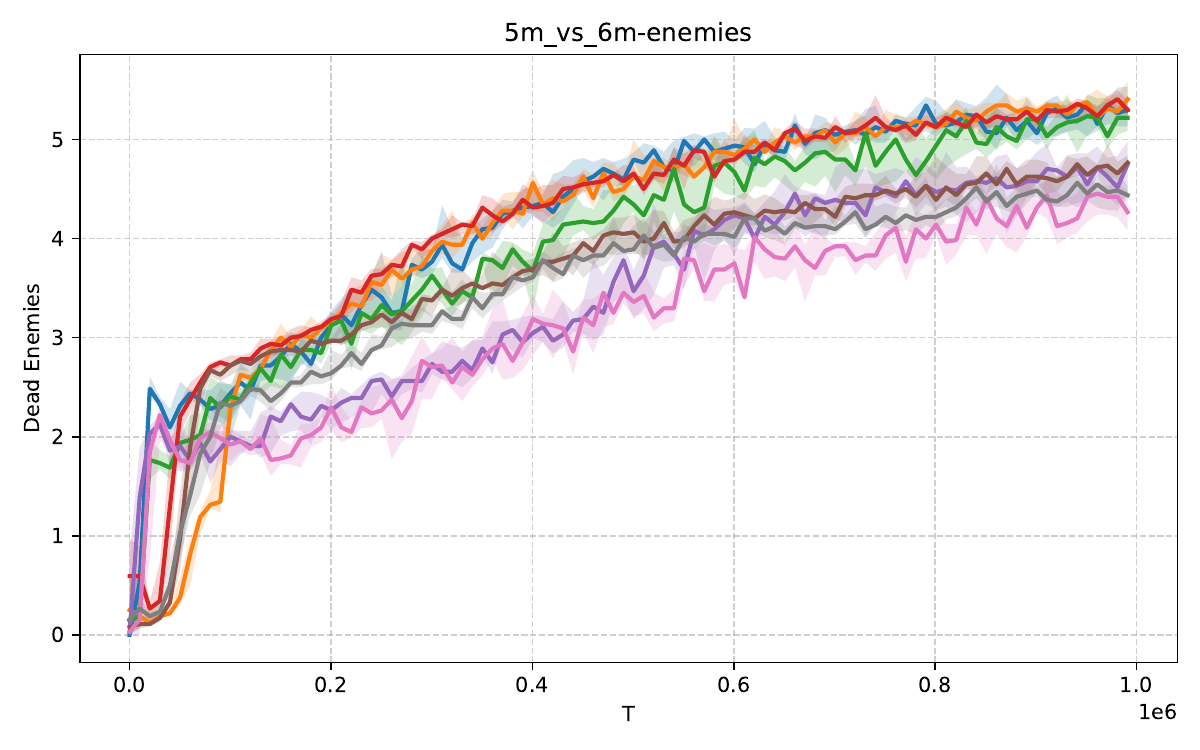}
        \caption{5m\_vs\_6m}
        \label{fig:5m_vs_6m-enemies}
    \end{subfigure}\hfill
    \begin{subfigure}[t]{0.32\textwidth}
        \centering
        \includegraphics[width=\linewidth]{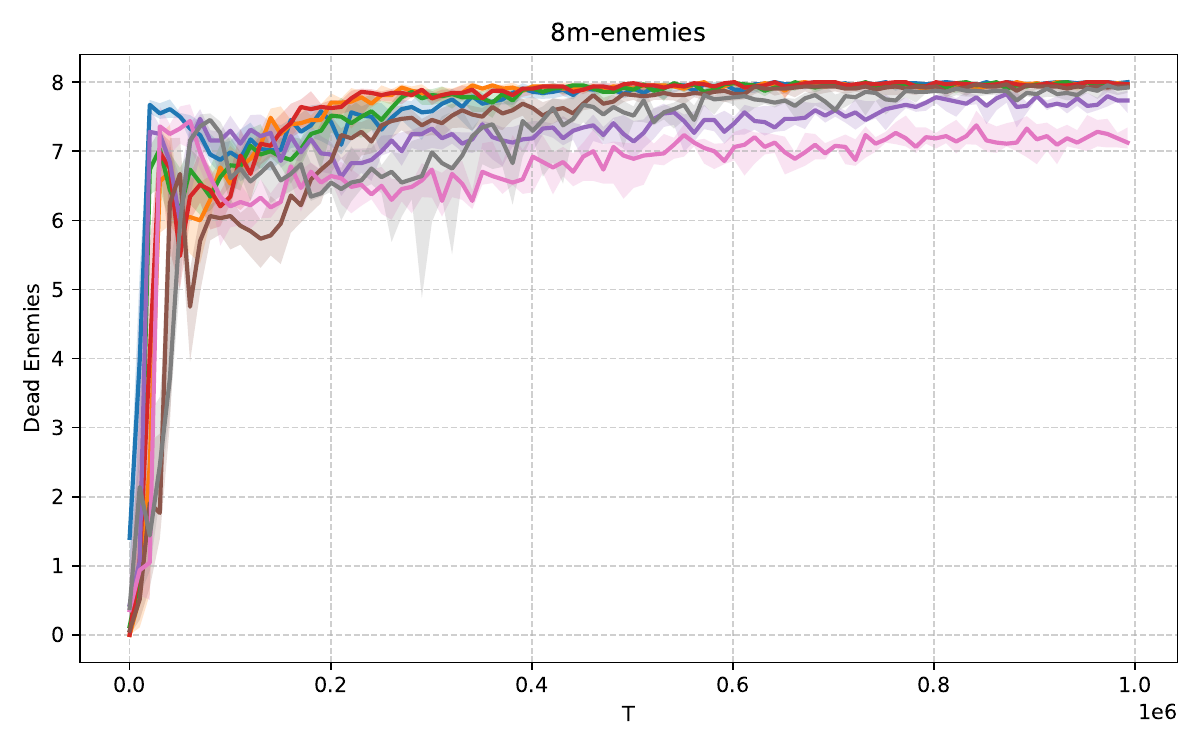}
        \caption{8m}
        \label{fig:8m-enemies}
    \end{subfigure}

    \begin{subfigure}[t]{0.32\textwidth}
        \centering
        \includegraphics[width=\linewidth]{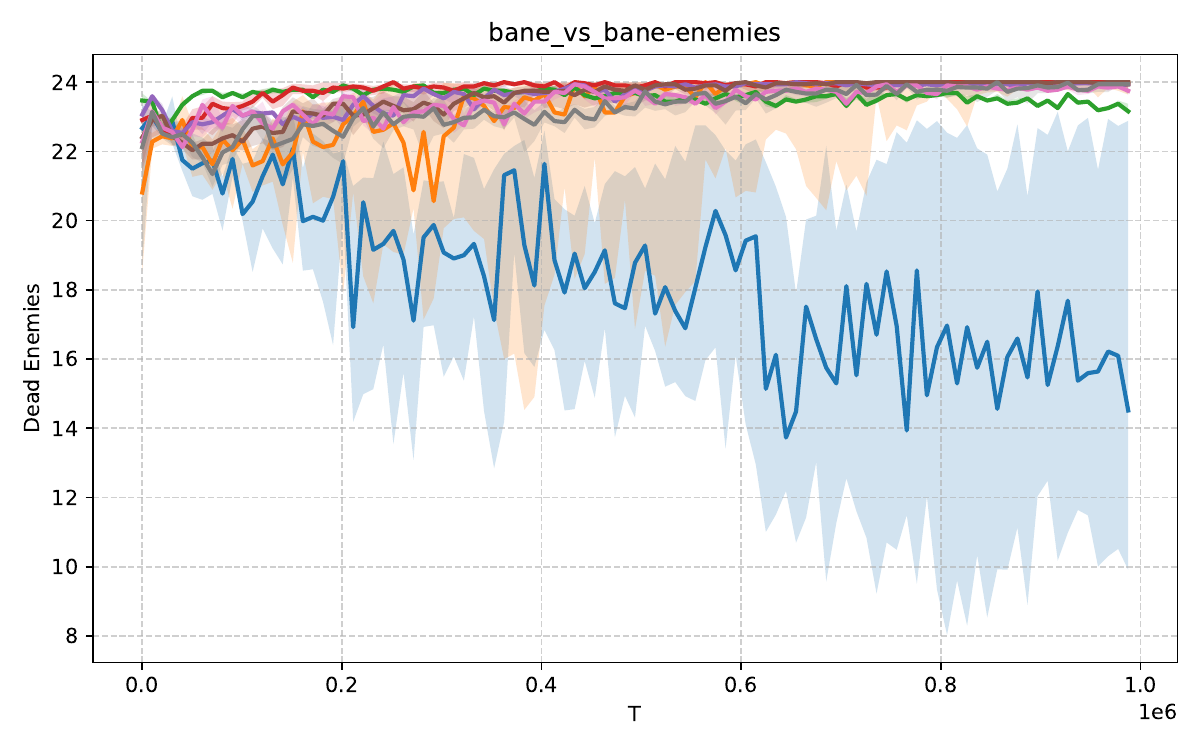}
        \caption{bane\_vs\_bane}
        \label{fig:bane_vs_bane-enemies}
    \end{subfigure}\hfill
    \begin{subfigure}[t]{0.32\textwidth}
        \centering
        \includegraphics[width=\linewidth]{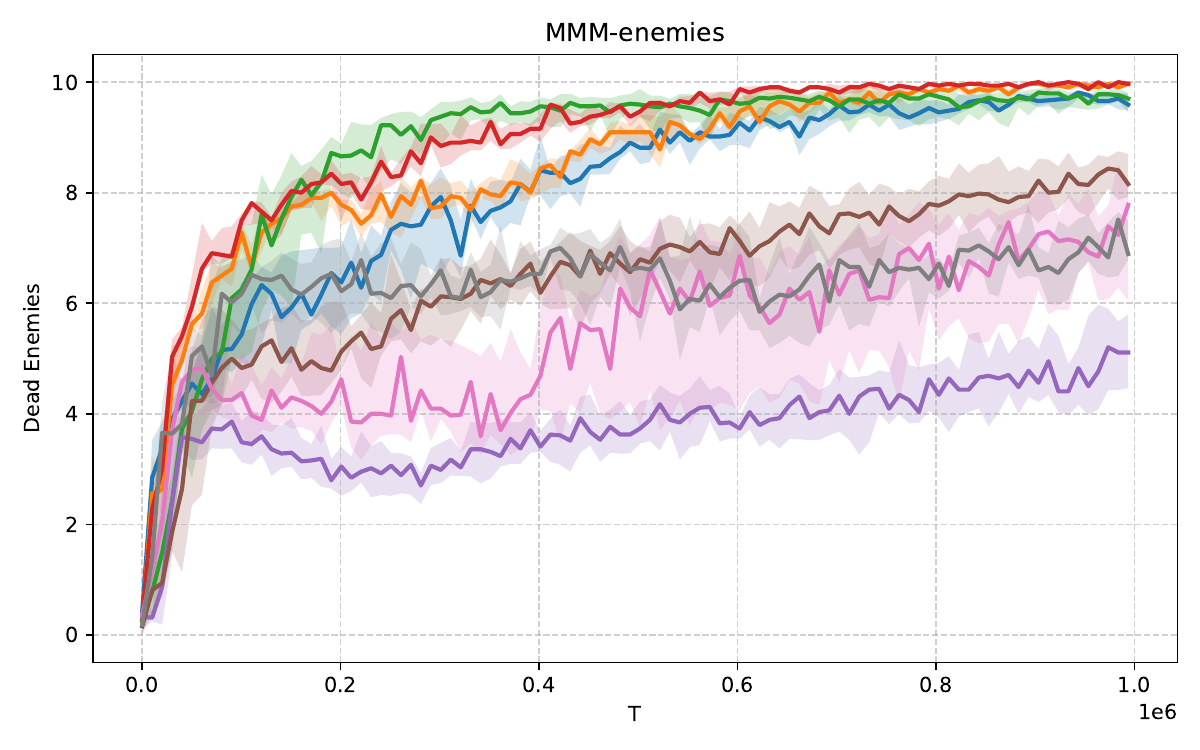}
        \caption{MMM}
        \label{fig:MMM-enemies}
    \end{subfigure}\hfill
    \begin{subfigure}[t]{0.32\textwidth}
        \centering
        \includegraphics[width=\linewidth]{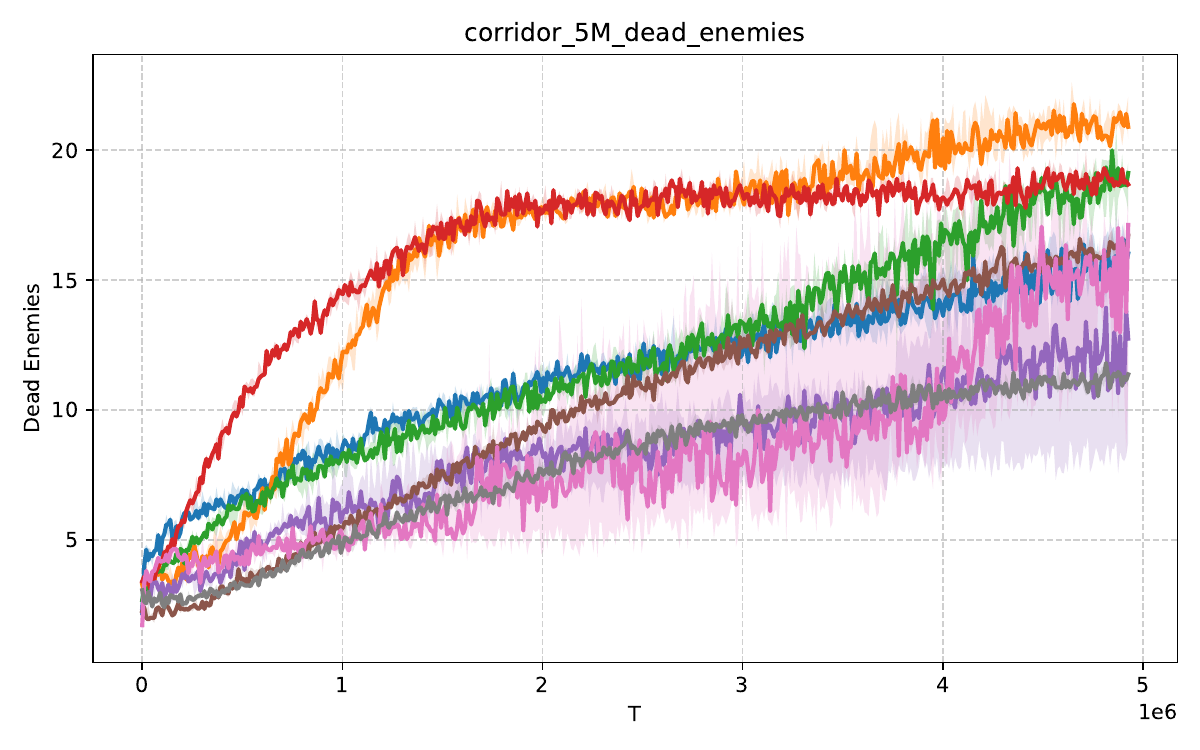}
        \caption{corridor}
        \label{fig:/corridor-enemies}
    \end{subfigure}

    \begin{subfigure}[t]{0.32\textwidth}
        \centering
        \includegraphics[width=\linewidth]{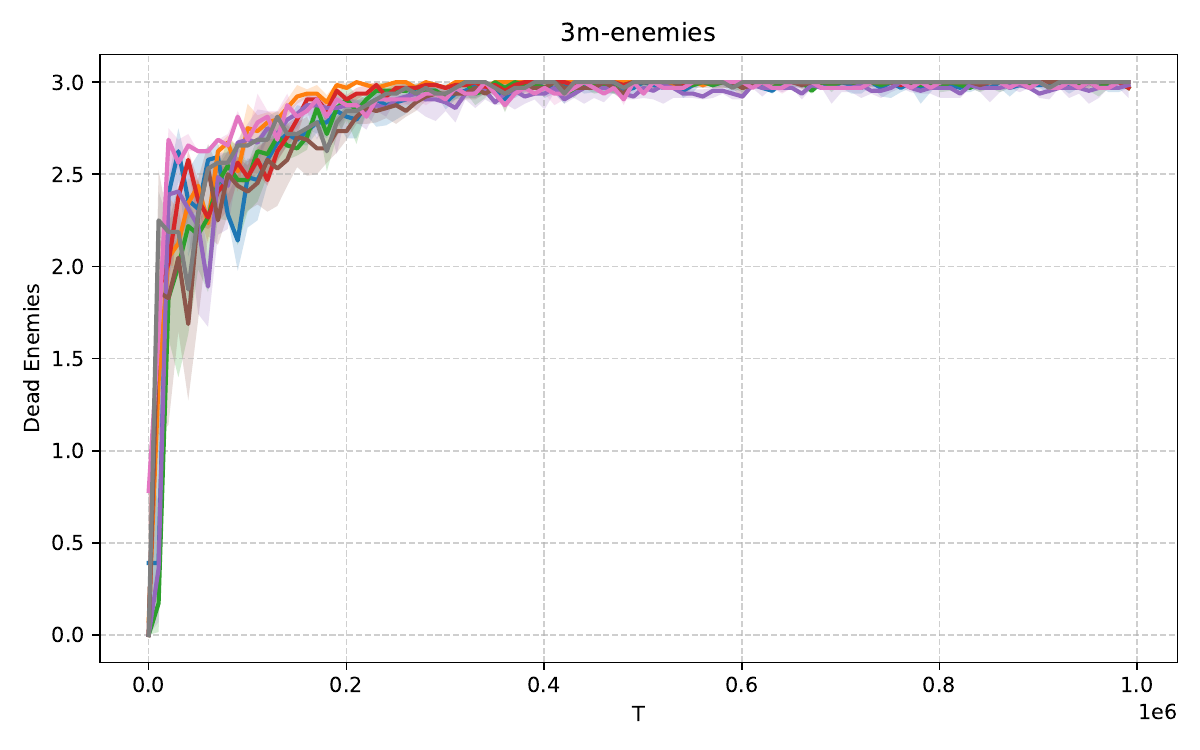}
        \caption{3m}
        \label{fig:3m-enemies}
    \end{subfigure}\hfill
    \begin{subfigure}[t]{0.32\textwidth}
        \centering
        \includegraphics[width=\linewidth]{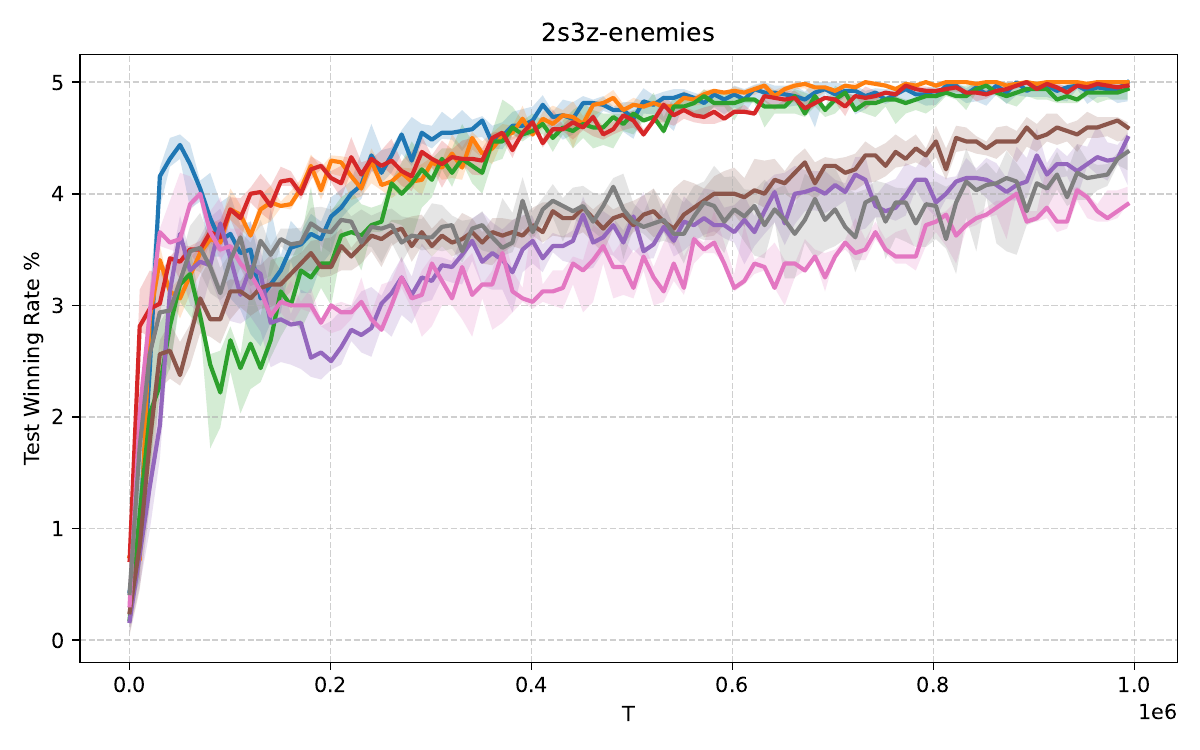}
        \caption{2s3z}
        \label{fig:2s3z-enemies}
    \end{subfigure}\hfill
    \begin{subfigure}[t]{0.32\textwidth}
        \centering
        \includegraphics[width=\linewidth]{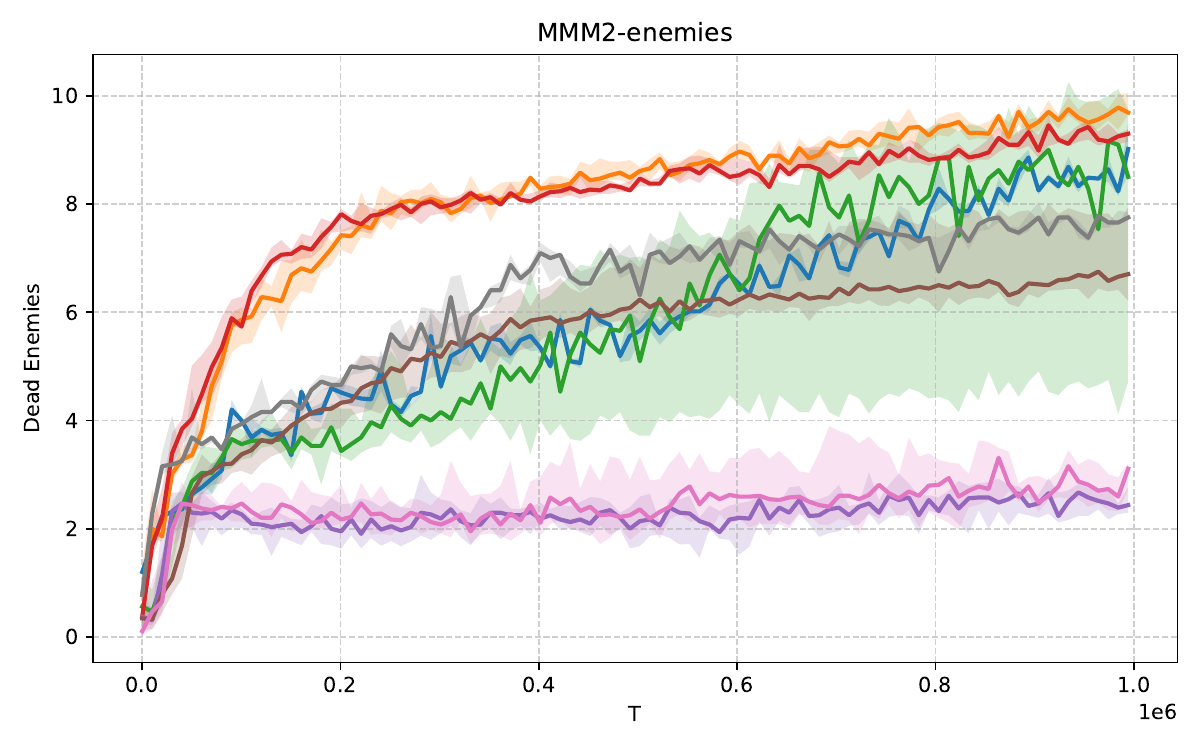}
        \caption{MMM2}
        \label{fig:MMM2-enemies}
    \end{subfigure}

    \caption{Learning curves on different SMAC combat maps, showing median dead enemy agents with 25-75\% percentile.}
    \label{fig:enemy_dead}
\end{figure*}

In addition to the maps discussed in \cref{sec:eval}, we provide results on additional SMAC scenarios. 
\cref{fig:winnig_main_rest} summarizes performance on \texttt{3m}, \texttt{2s3z}, and \texttt{MMM2}. 
As shown in \cref{fig:3m}, \texttt{3m} is relatively easy and all methods achieve similar performance. 
In contrast, on \texttt{2s3z} (\cref{fig:2s3z}) we observe clear gains for \method+QTRAN \brownline and \method+IQL \grayline over vanilla QTRAN \purpleline and IQL \pinkline. 
On \texttt{MMM2} (\cref{fig:MMM2}), where we apply $\medivac$ within \method, \method+QMIX \orangeline clearly outperforms vanilla QMIX \blueline. 
\method+IQL \grayline also achieves a non-trivial win rate (around $5\%$), whereas vanilla IQL \pinkline attains $0\%$. 
While both \method+QTRAN and vanilla QTRAN obtain $0\%$ win rate, \method+QTRAN \brownline eliminates more enemies than vanilla QTRAN (see \cref{fig:MMM2-enemies}).

\begin{figure*}[]
    \centering
    \begin{subfigure}[t]{\textwidth}
        \centering
        \includegraphics[width=0.99\linewidth]{img/legend_only.pdf}
    \end{subfigure}

    \begin{subfigure}[t]{0.4\textwidth}
        \centering
        \includegraphics[width=\linewidth]{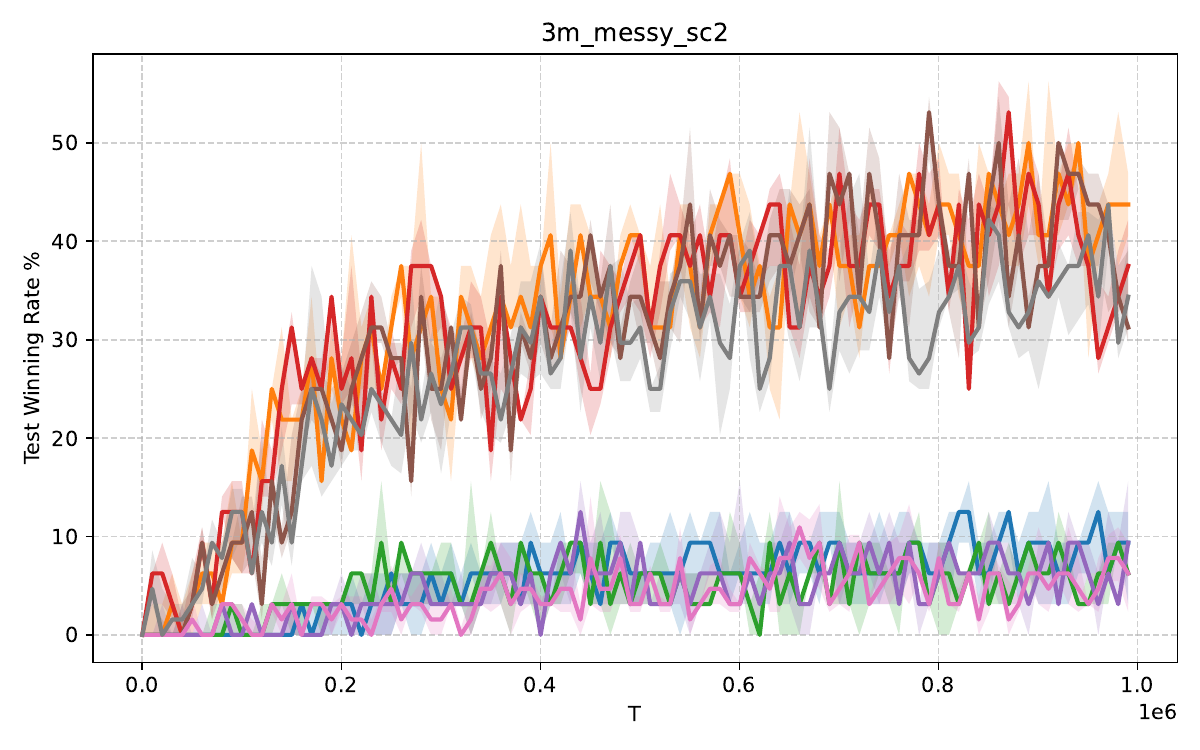}
        \caption{3m }
        \label{fig:3m_messy}
    \end{subfigure}\hfill
    \begin{subfigure}[t]{0.4\textwidth}
        \centering
        \includegraphics[width=\linewidth]{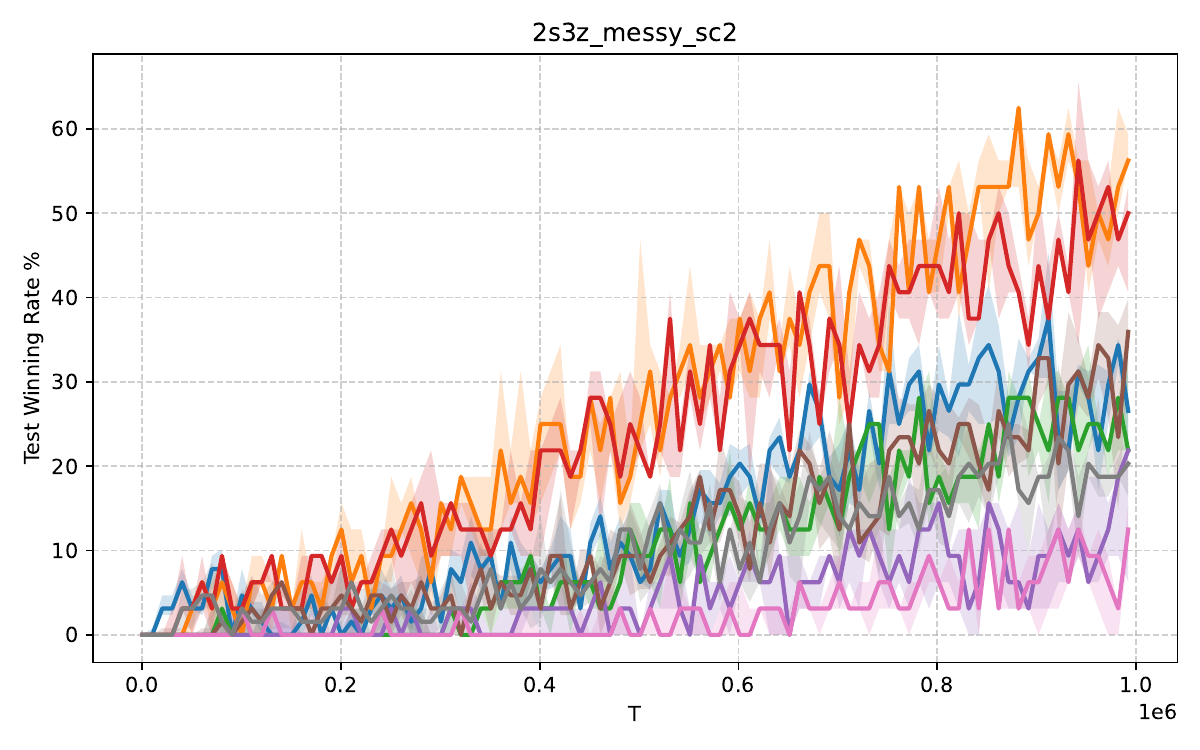}
        \caption{2s3z}
        \label{fig:2s3z_messy}
    \end{subfigure}\hfill

    \caption{Learning curves on remaining MessySMAC maps, showing median test win rate with 25-75\% percentile.}
    \label{fig:messy_rest}
\end{figure*}

The remaining MessySMAC maps are shown in~\Cref{fig:messy_rest}. 
In the \texttt{3m} map~\Cref{fig:3m_messy}, \method+MARL substantially outperforms the vanilla MARL baselines across all cases. 
In the \texttt{2s3z} map~\Cref{fig:2s3z_messy}, \method+QMIX \orangeline and \method+VDN \redline substantially outperform vanilla QMIX \blueline and vanilla VDN \greenline, respectively.

\begin{figure*}[]

    \centering
        \begin{subfigure}[t]{\textwidth}
        \centering
        \includegraphics[width=0.9\linewidth]{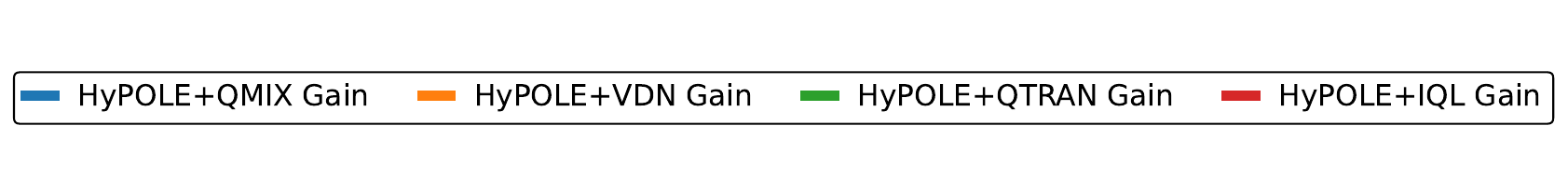}
    \end{subfigure}
    \begin{subfigure}[t]{0.32\textwidth}
        \centering
        \includegraphics[width=\linewidth]{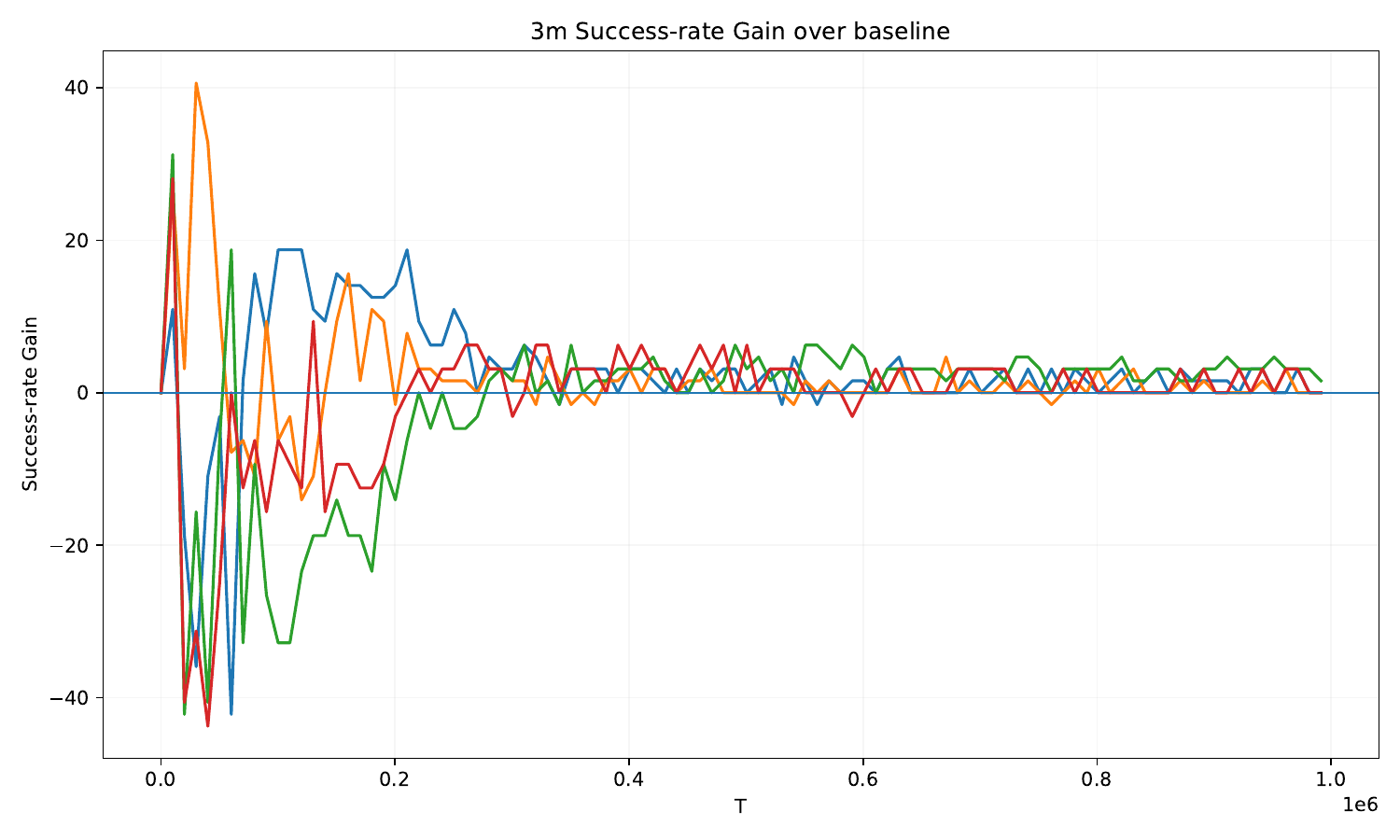}
        \caption{3m}
        \label{fig:3m_gain}
    \end{subfigure}\hfill
    \begin{subfigure}[t]{0.32\textwidth}
        \centering
        \includegraphics[width=\linewidth]{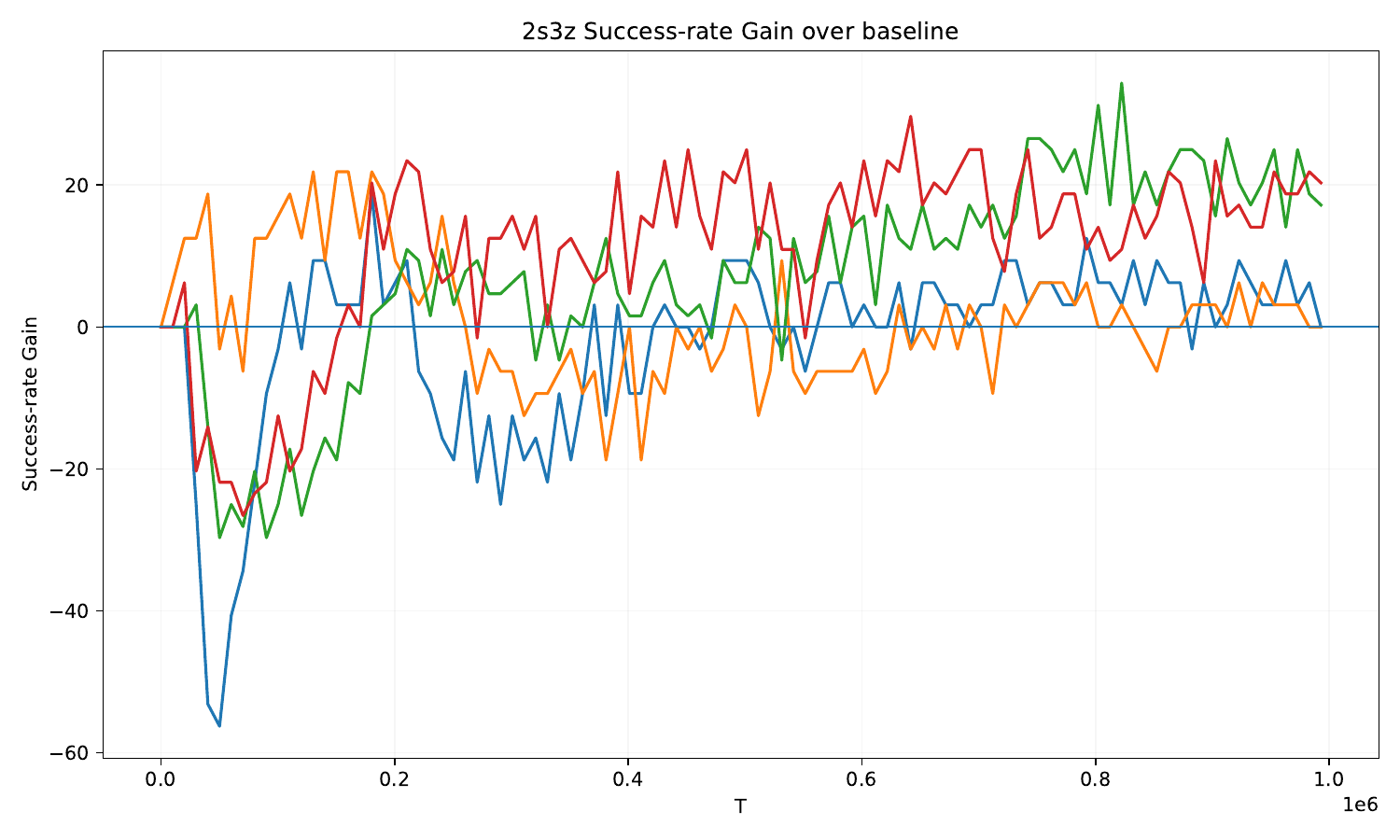}
        \caption{2s3z}
        \label{fig:2s3z_gain}
    \end{subfigure}\hfill
    \begin{subfigure}[t]{0.32\textwidth}
        \centering
        \includegraphics[width=\linewidth]{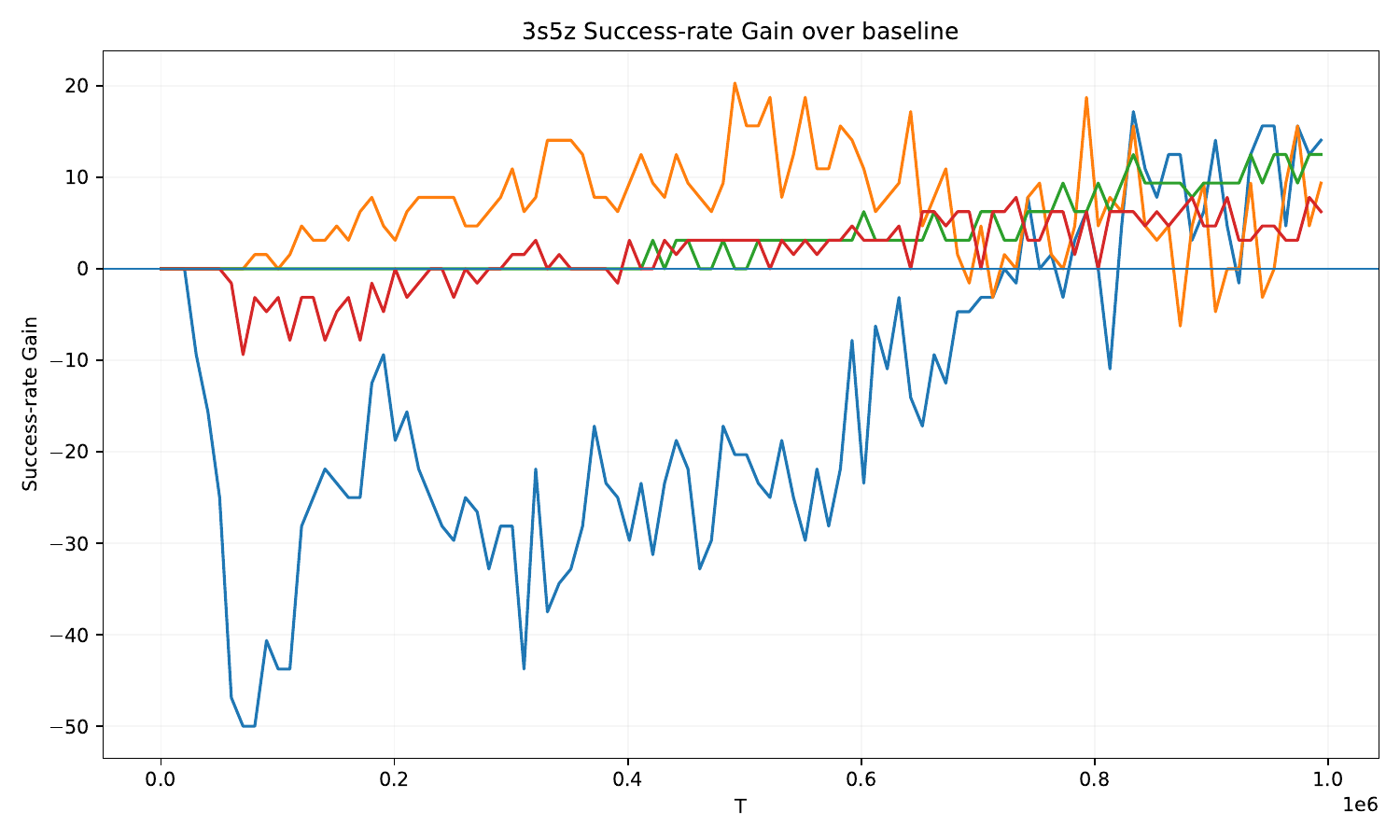}
        \caption{3s5z}
        \label{fig:3s5z_gain}
    \end{subfigure}\hfill
    \begin{subfigure}[t]{0.32\textwidth}
        \centering
        \includegraphics[width=\linewidth]{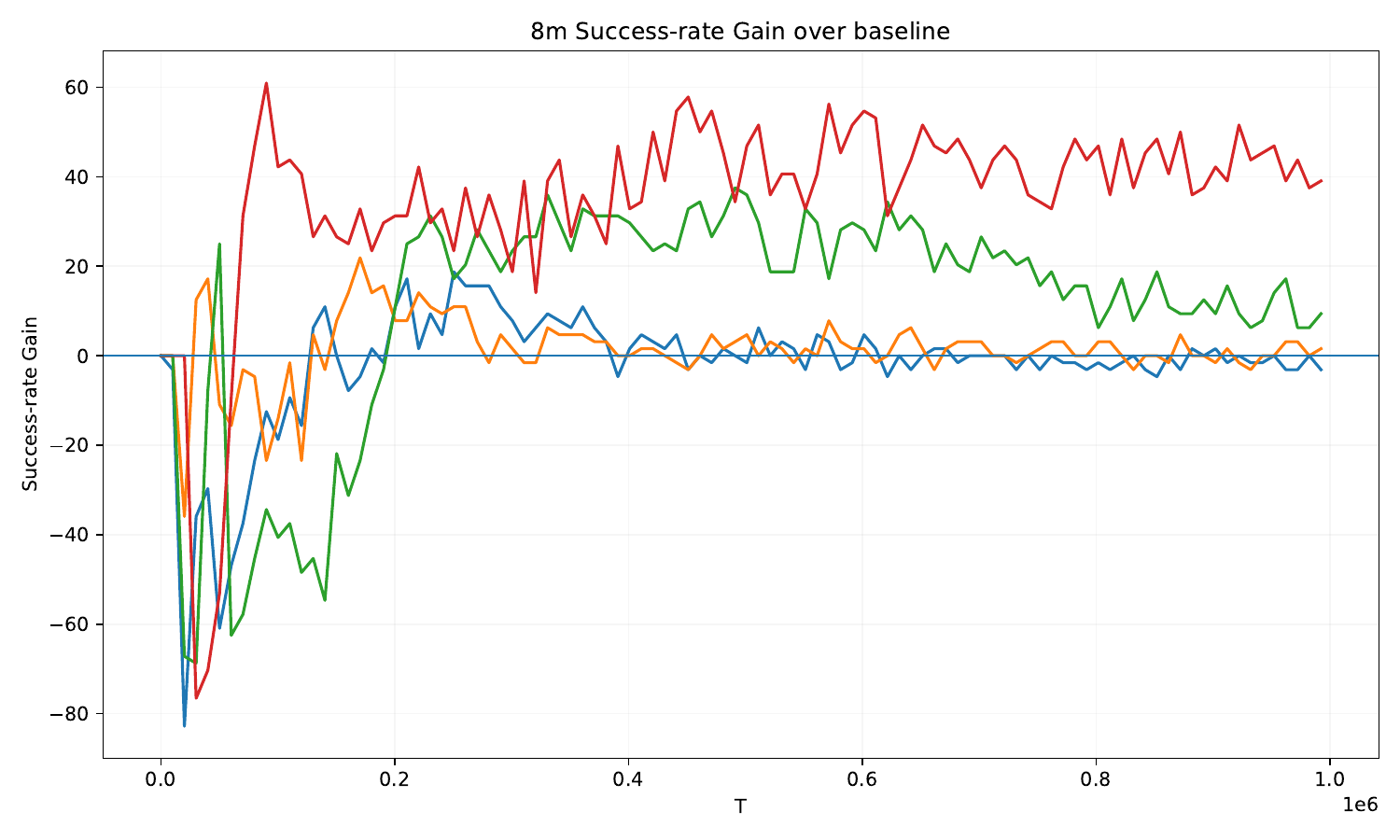}
        \caption{8m}
        \label{fig:8m_gain}
    \end{subfigure}
    \begin{subfigure}[t]{0.32\textwidth}
        \centering
        \includegraphics[width=\linewidth]{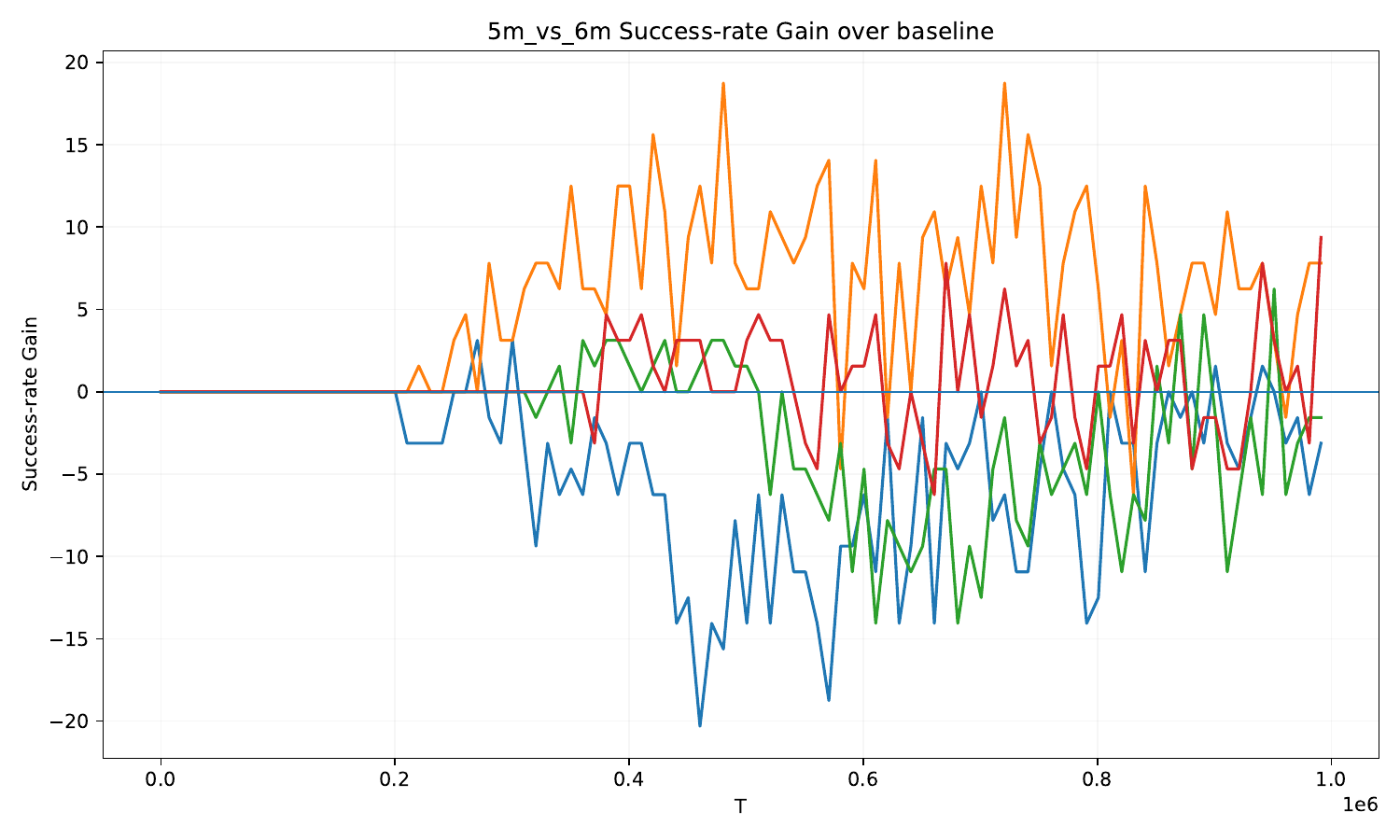}
        \caption{5m\_vs\_6m}
        \label{fig:5m_vs_6m_gain}
    \end{subfigure}\hfill
    \begin{subfigure}[t]{0.32\textwidth}
        \centering
        \includegraphics[width=\linewidth]{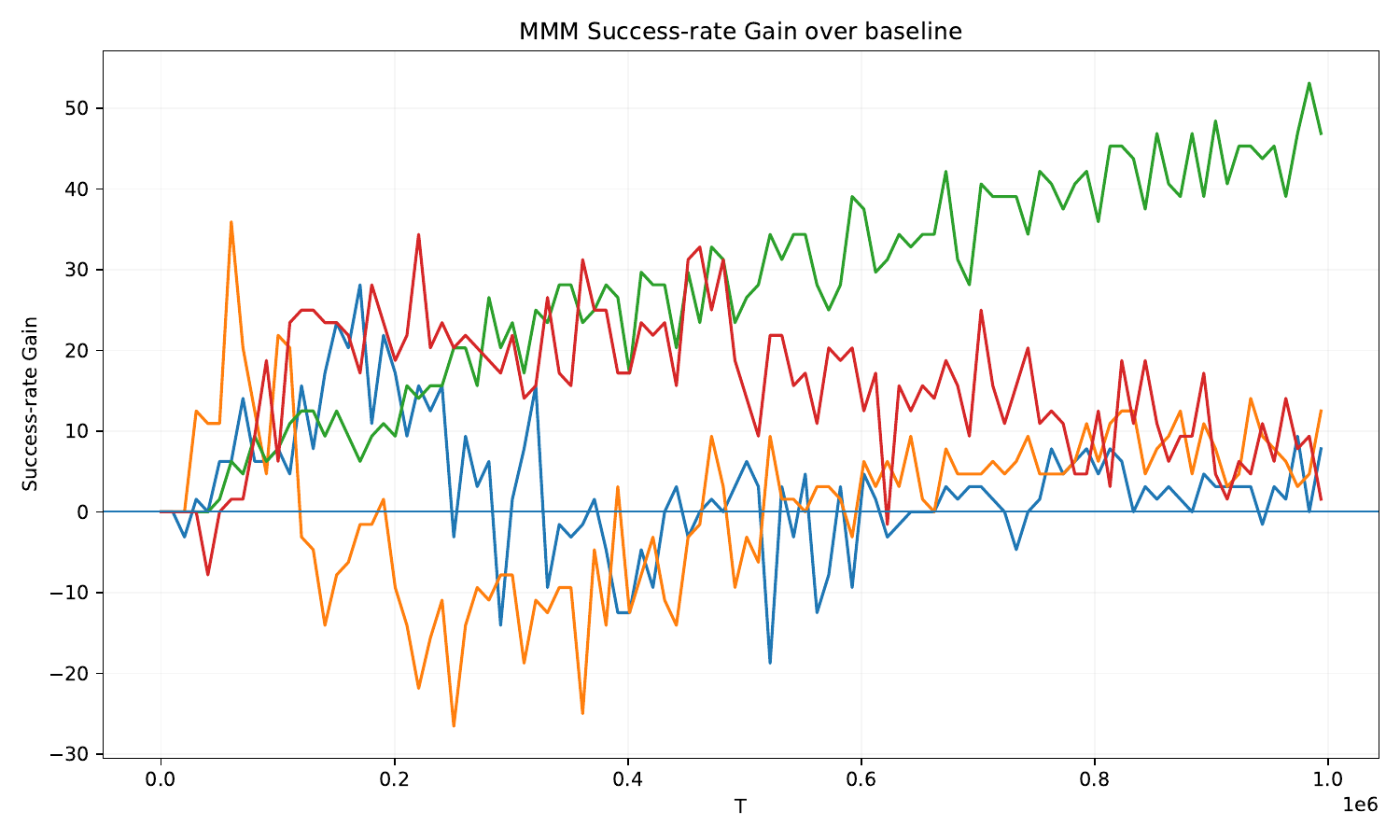}
        \caption{MMM}
        \label{fig:MMM_gain}
    \end{subfigure}\hfill
    \begin{subfigure}[t]{0.32\textwidth}
        \centering
        \includegraphics[width=\linewidth]{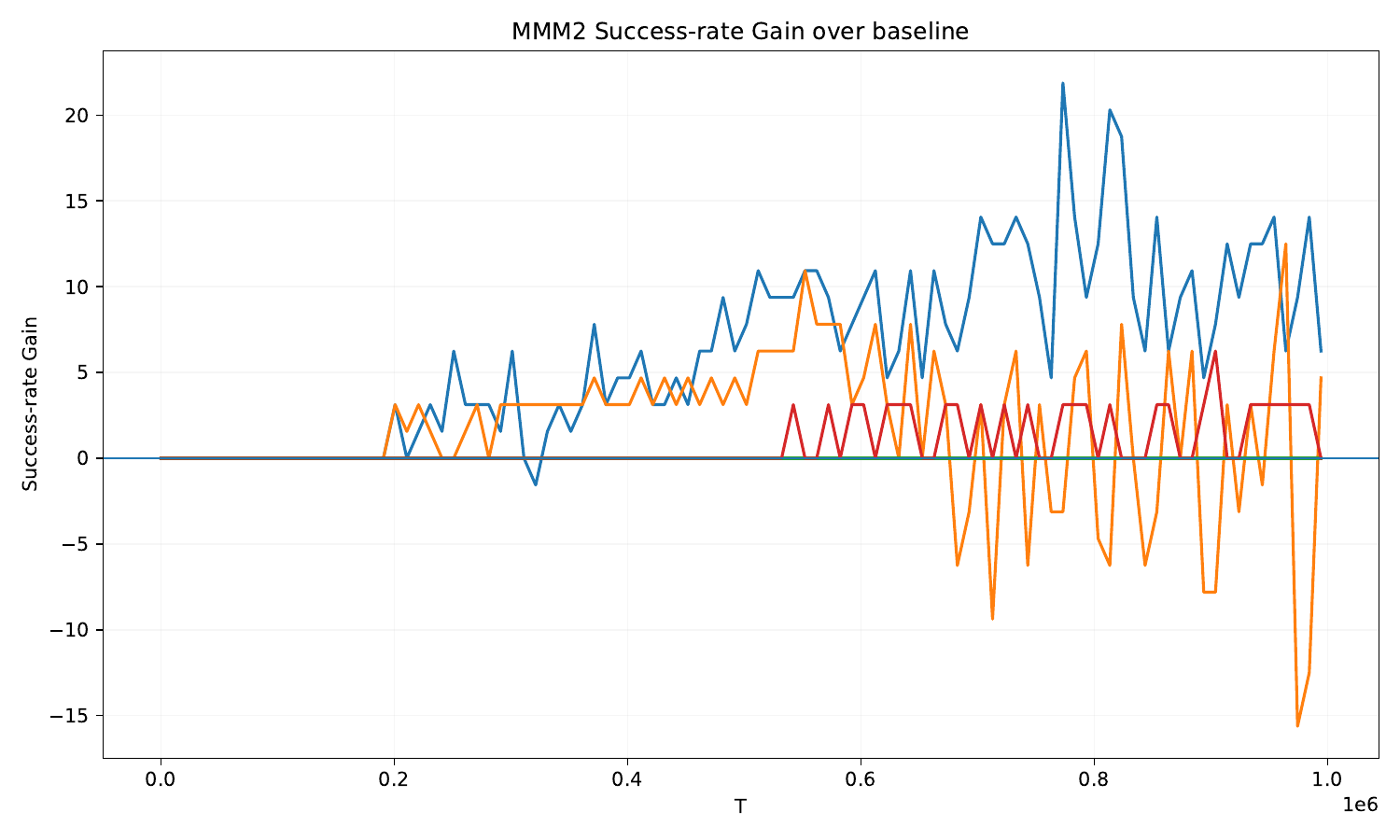}
        \caption{MMM2}
        \label{fig:MMM2-gain}
    \end{subfigure}
    \begin{subfigure}[t]{0.32\textwidth}
        \centering
        \includegraphics[width=\linewidth]{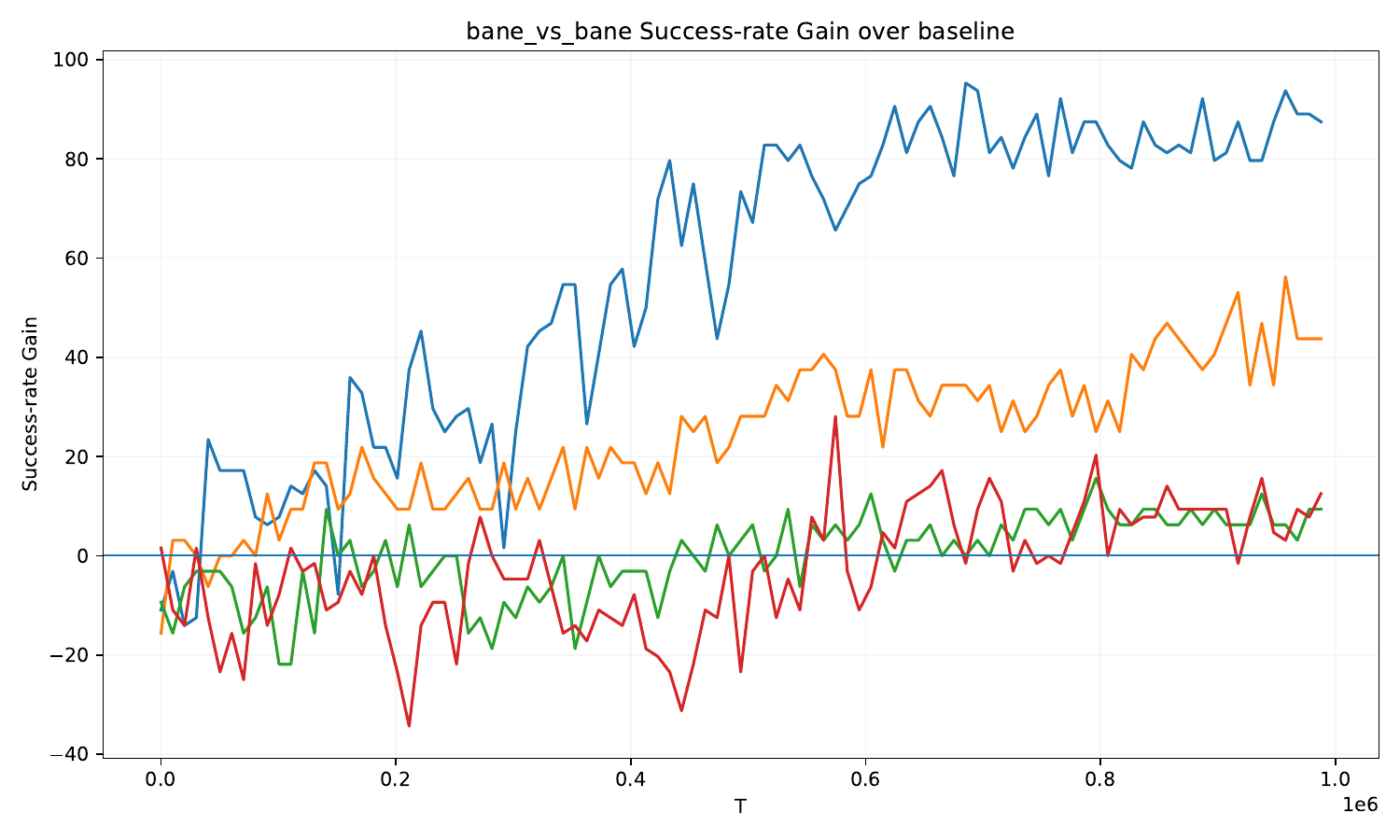}
        \caption{bane\_vs\_bane}
        \label{fig:bane_vs_bane_gain}
    \end{subfigure}\hfill
    \begin{subfigure}[t]{0.32\textwidth}
        \centering
        \includegraphics[width=\linewidth]{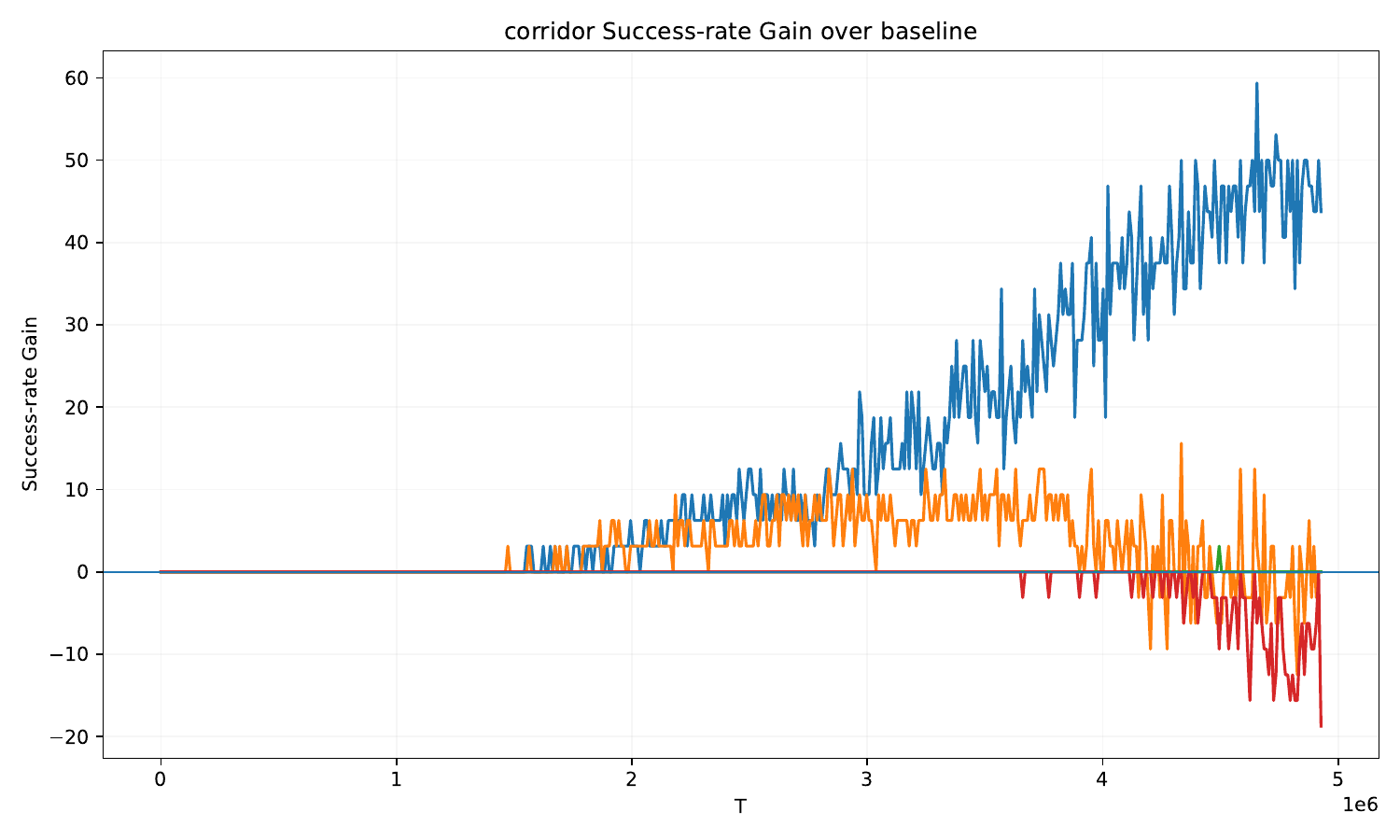}
        \caption{corridor}
        \label{fig:corridor_gain}
    \end{subfigure}\hfill

    \caption{Median Gain in Percentage of \method compared to their respective baselines in SMAC.}
    \label{fig:gain_sc2}
\end{figure*}

To better illustrate the significance of \method, we report its success-rate gains across all SMAC maps in~\Cref{fig:gain_sc2}.
\method+QMIX \blueline outperforms vanilla QMIX on almost all maps, except \texttt{5m\_vs\_6m} and \texttt{3s5z}; however, on \texttt{3s5z}, it still surpasses the baseline in the final training steps.
On hard scenarios such as \texttt{MMM2}~\Cref{fig:MMM2-gain}, \texttt{bane\_vs\_bane}~\Cref{fig:bane_vs_bane_gain}, and \texttt{corridor}~\Cref{fig:corridor_gain}, \method+QMIX achieves substantial improvements, reaching up to a $90\%$ gain in success rate on \texttt{bane\_vs\_bane}.
Similarly, \method+VDN \orangeline improves performance in most cases, except on \texttt{2s3z}, where the behavior is more unstable, while still achieving substantial gains on \texttt{bane\_vs\_bane}.
Finally, \method+QTRAN \greenline outperforms vanilla QTRAN in nearly all scenarios except \texttt{5m\_vs\_6m}, and achieves notable improvements on \texttt{MMM}, with up to a $50\%$ gain~\Cref{fig:MMM_gain}.
In addition to CTDE methods, \method+IQL \redline performs well on almost all maps, achieving about a $60\%$ gain on \texttt{8m}~\Cref{fig:8m_gain}.

\begin{figure*}[]

    \centering
        \begin{subfigure}[t]{\textwidth}
        \centering
        \includegraphics[width=0.9\linewidth]{img/legend_only-reb-gain.pdf}
    \end{subfigure}
    \begin{subfigure}[t]{0.32\textwidth}
        \centering
        \includegraphics[width=\linewidth]{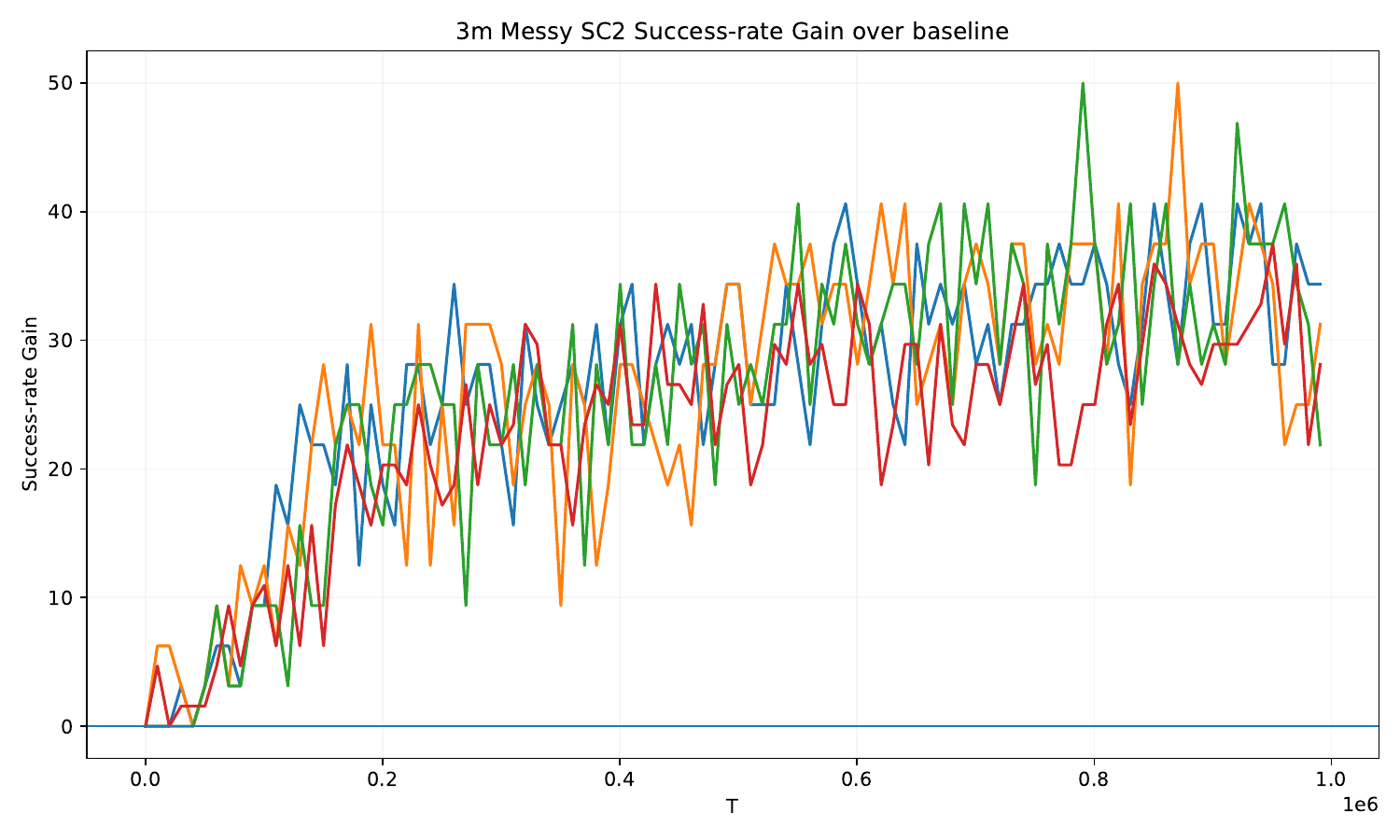}
        \caption{3m}
        \label{fig:3m_messy_sc2_gain}
    \end{subfigure}\hfill
    \begin{subfigure}[t]{0.32\textwidth}
        \centering
        \includegraphics[width=\linewidth]{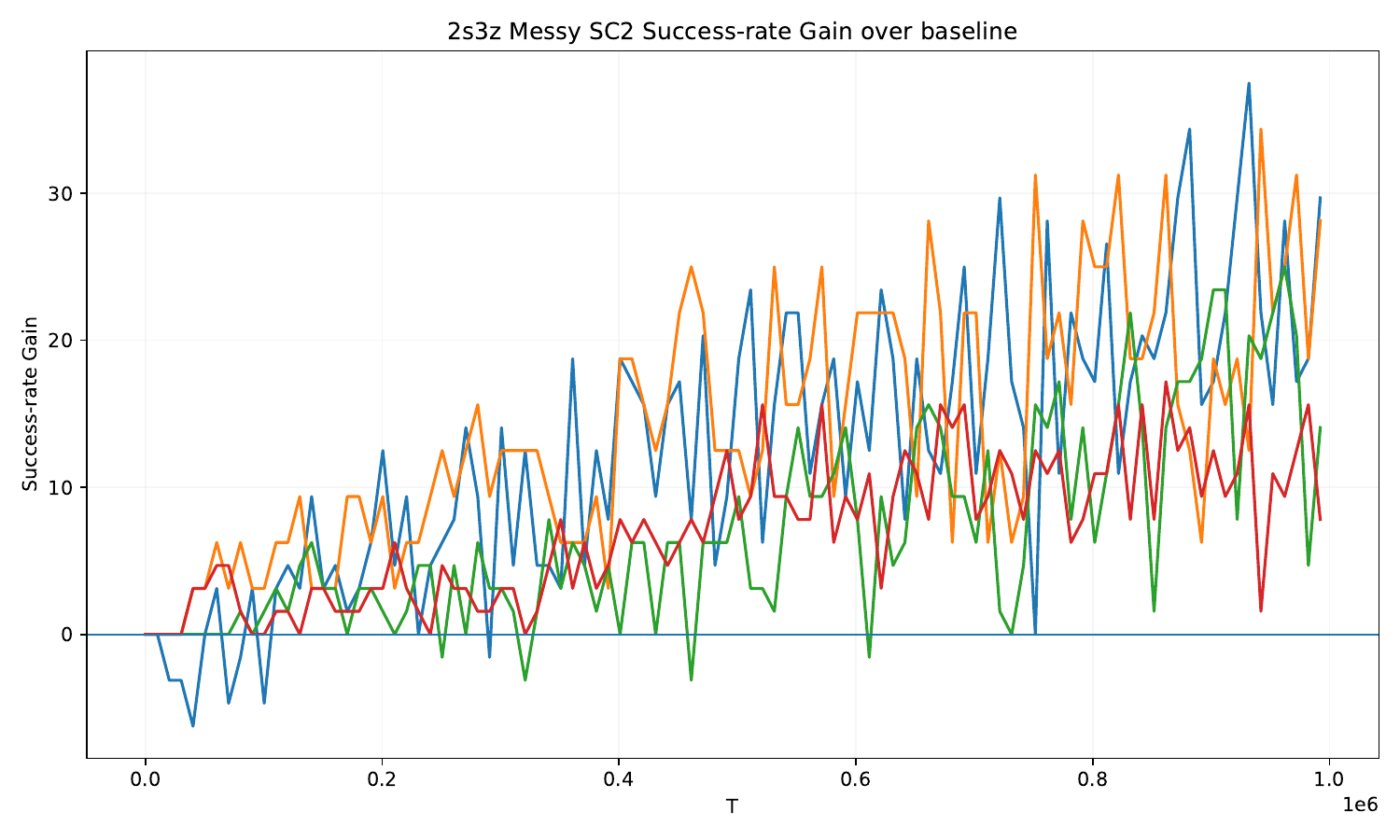}
        \caption{2s3z}
        \label{fig:2s3z_messy_sc2_gain}
    \end{subfigure}\hfill
    \begin{subfigure}[t]{0.32\textwidth}
        \centering
        \includegraphics[width=\linewidth]{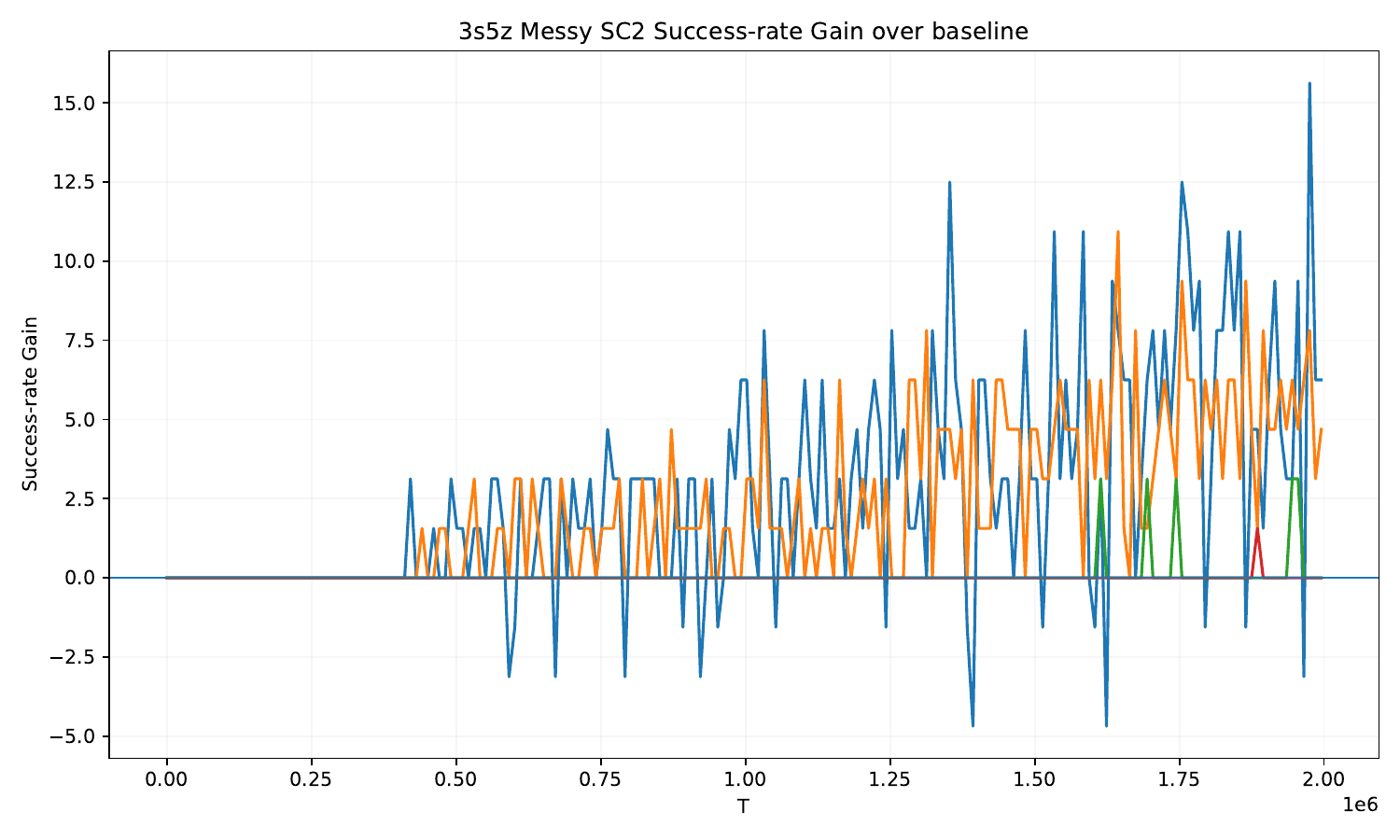}
        \caption{3s5z}
        \label{fig:3s5z_messy_sc2_gain}
    \end{subfigure}\hfill
    \begin{subfigure}[t]{0.32\textwidth}
        \centering
        \includegraphics[width=\linewidth]{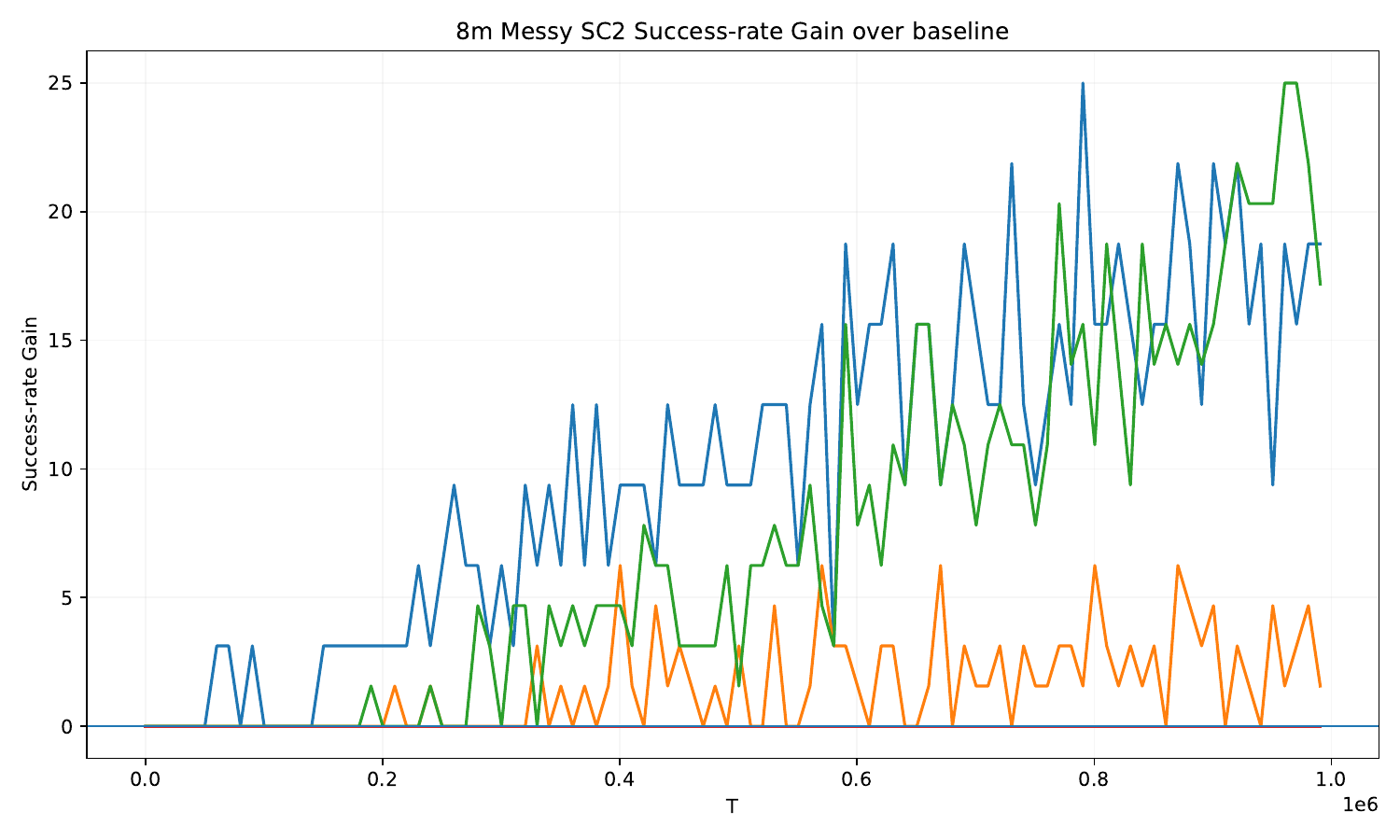}
        \caption{8m}
        \label{fig:8m_messy_sc2_gain}
    \end{subfigure}
    \begin{subfigure}[t]{0.32\textwidth}
        \centering
        \includegraphics[width=\linewidth]{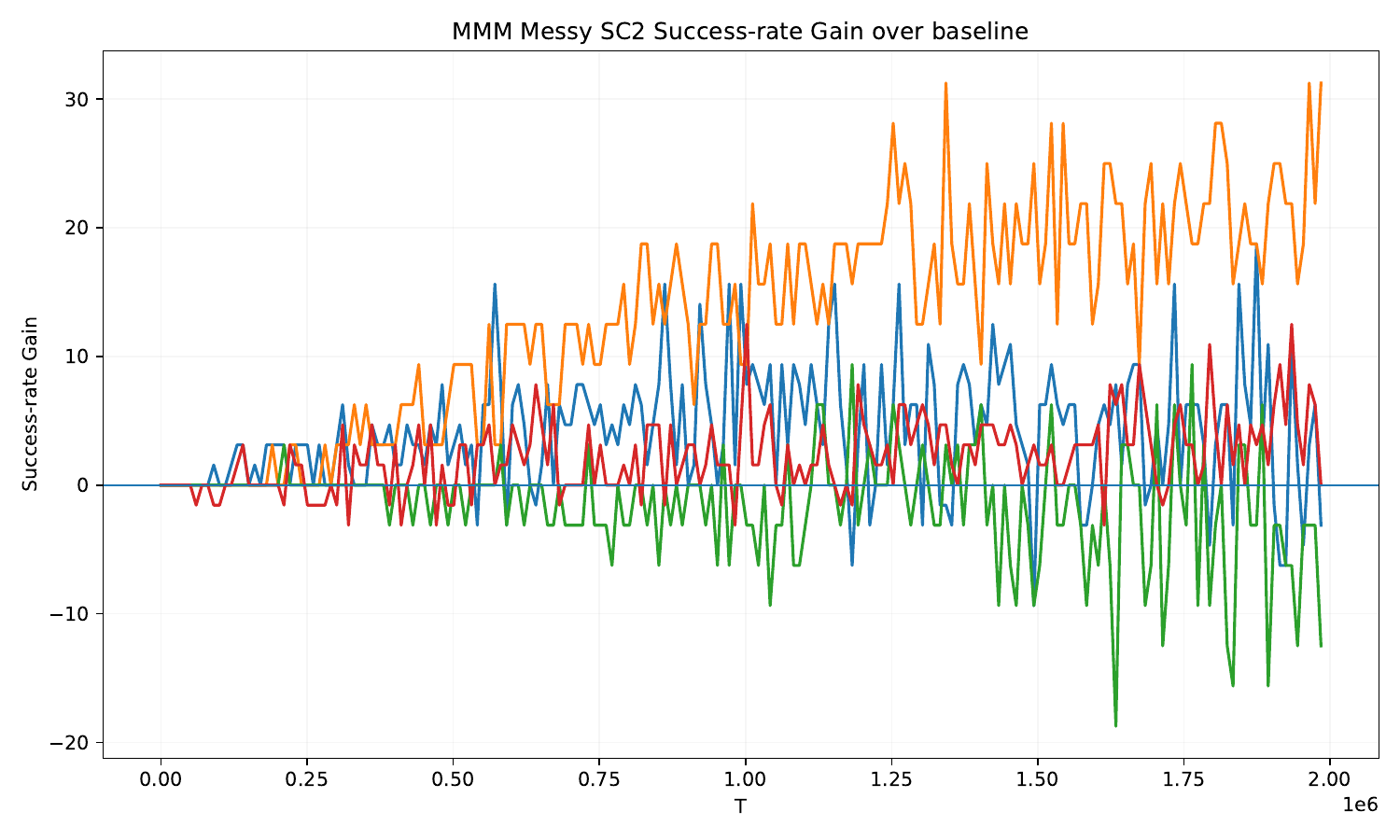}
        \caption{MMM}
        \label{fig:MMM_messy_sc2_gain}
    \end{subfigure}\hfill

    \caption{Median Gain in Percentage of \method compared to their respective baselines in MessySMAC.}
    \label{fig:gain_messy}
\end{figure*}

We perform the same analysis for MessySMAC to further illustrate the significance of \method, and report its success-rate gains across all MessySMAC maps in~\Cref{fig:gain_messy}.
\method+QMIX \blueline improves performance on most maps, especially on \texttt{3m}~\Cref{fig:3m_messy_sc2_gain}, where it achieves up to a $50\%$ gain.
\method+VDN \orangeline improves the win rate over vanilla VDN in all cases, and performs particularly well on the harder \texttt{MMM} map~\Cref{fig:MMM_messy_sc2_gain}, with up to a $30\%$ gain.
\method+QTRAN \greenline performs better in most cases, except on \texttt{MMM}~\Cref{fig:MMM_messy_sc2_gain}; however, it shows substantial improvement on \texttt{3m}~\Cref{fig:3m_messy_sc2_gain}, with up to a $50\%$ gain.
Similarly, \method+IQL \redline performs better on most maps, especially on \texttt{3m}~\Cref{fig:3m_messy_sc2_gain}, with up to a $\approx35\%$ gain, except on \texttt{8m}~\Cref{fig:8m_messy_sc2_gain}, where neither \method+IQL nor vanilla IQL achieves any wins.

\begin{figure*}[t!]
    \centering
        \begin{subfigure}[t]{\textwidth}
        \centering
        \includegraphics[width=0.99\linewidth]{img/legend_only.pdf}
    \end{subfigure}

    \begin{subfigure}[t]{0.32\textwidth}
        \centering
        \includegraphics[width=\linewidth]{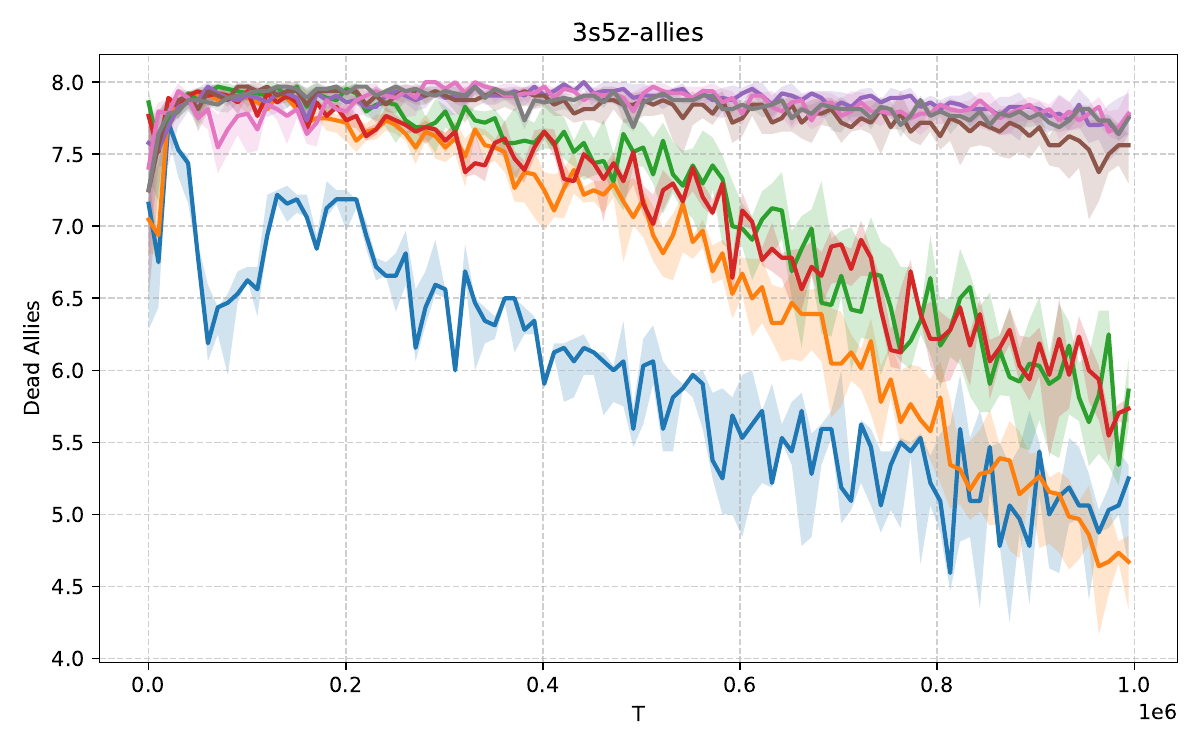}
        \caption{3s5z}
        \label{fig:3s5z-allies}
    \end{subfigure}\hfill
    \begin{subfigure}[t]{0.32\textwidth}
        \centering
        \includegraphics[width=\linewidth]{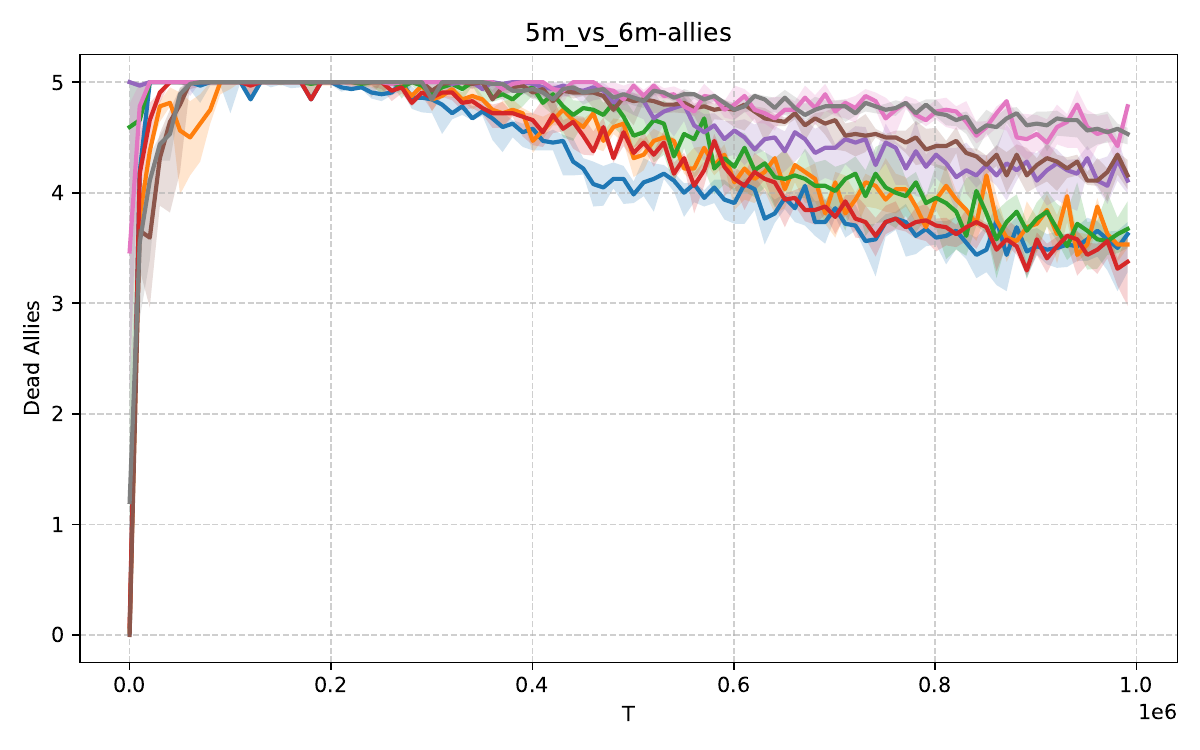}
        \caption{5m\_vs\_6m}
        \label{fig:5m_vs_6m-allies}
    \end{subfigure}\hfill
    \begin{subfigure}[t]{0.32\textwidth}
        \centering
        \includegraphics[width=\linewidth]{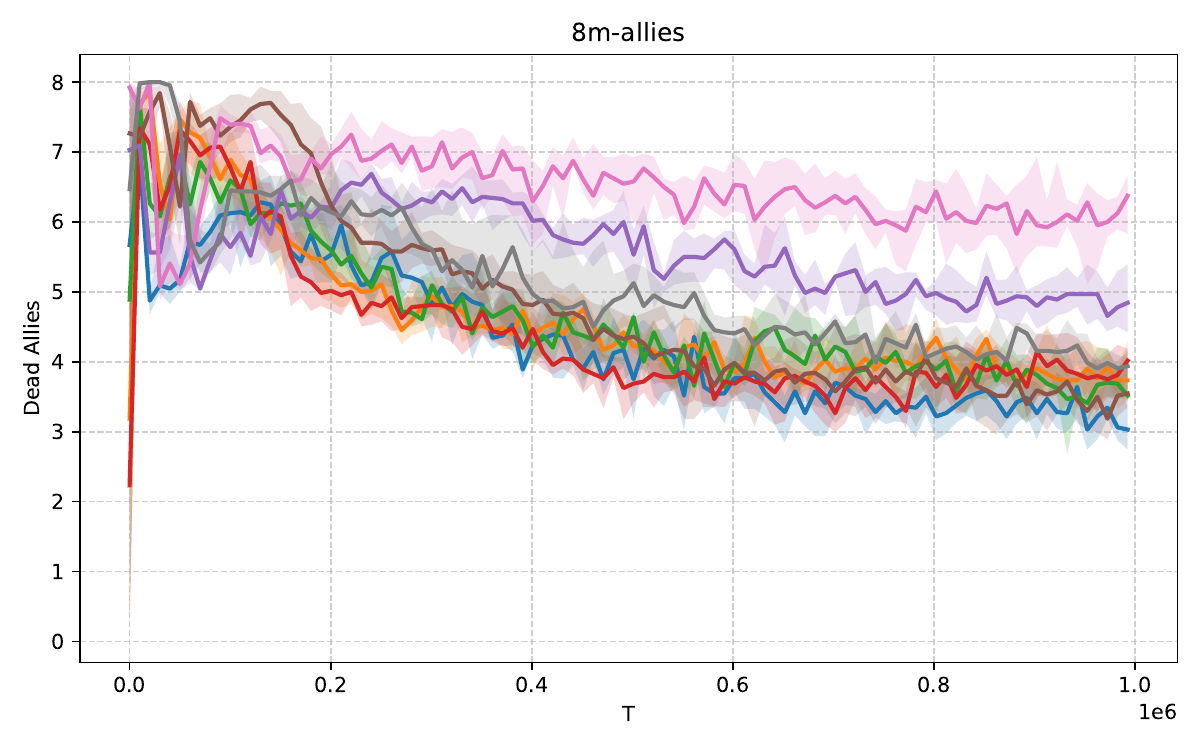}
        \caption{8m}
        \label{fig:8m-allies}
    \end{subfigure}

    \begin{subfigure}[t]{0.32\textwidth}
        \centering
        \includegraphics[width=\linewidth]{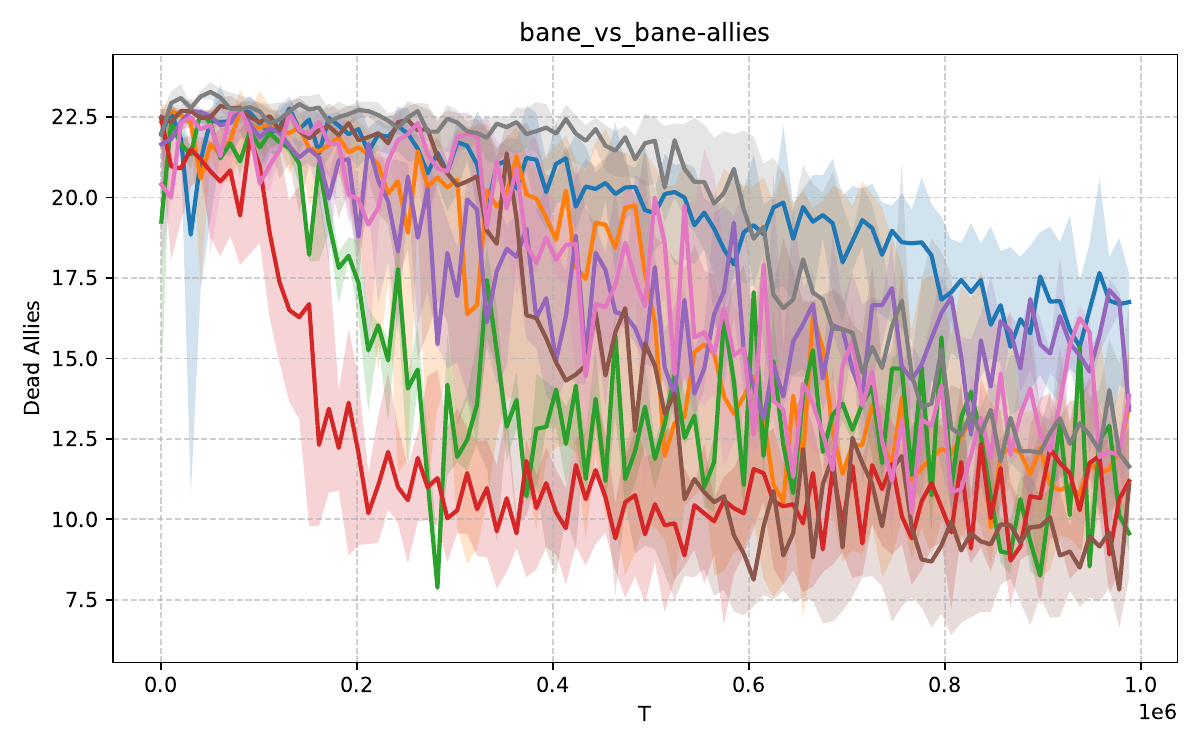}
        \caption{bane\_vs\_bane}
        \label{fig:bane_vs_bane-allies}
    \end{subfigure}\hfill
    \begin{subfigure}[t]{0.32\textwidth}
        \centering
        \includegraphics[width=\linewidth]{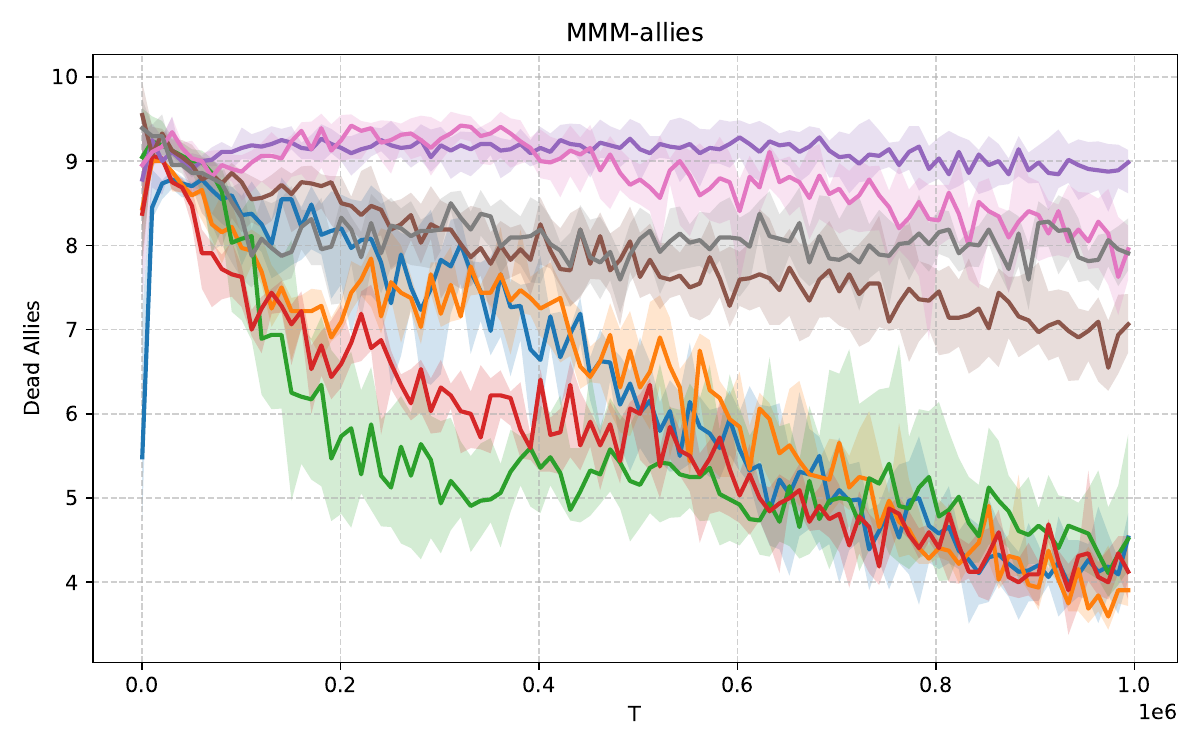}
        \caption{MMM}
        \label{fig:MMM-allies}
    \end{subfigure}\hfill
    \begin{subfigure}[t]{0.32\textwidth}
        \centering
        \includegraphics[width=\linewidth]{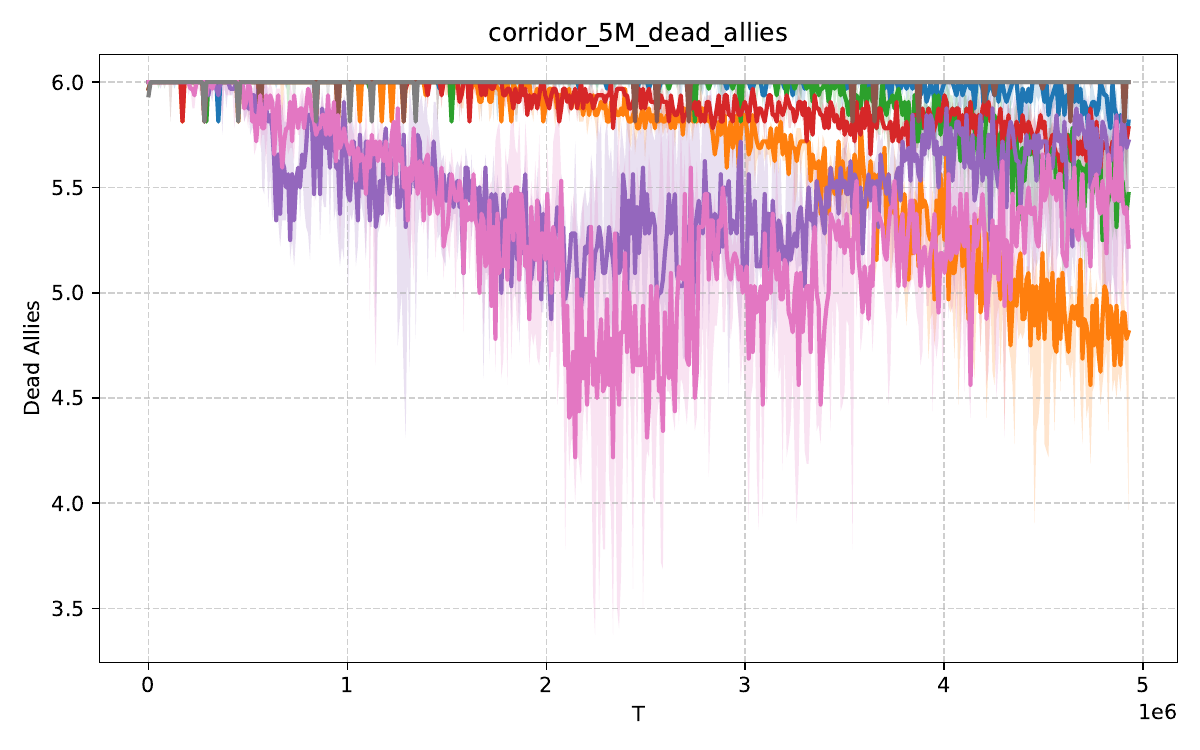}
        \caption{corridor}
        \label{fig:/corridor-allies}
    \end{subfigure}

    \begin{subfigure}[t]{0.32\textwidth}
        \centering
        \includegraphics[width=\linewidth]{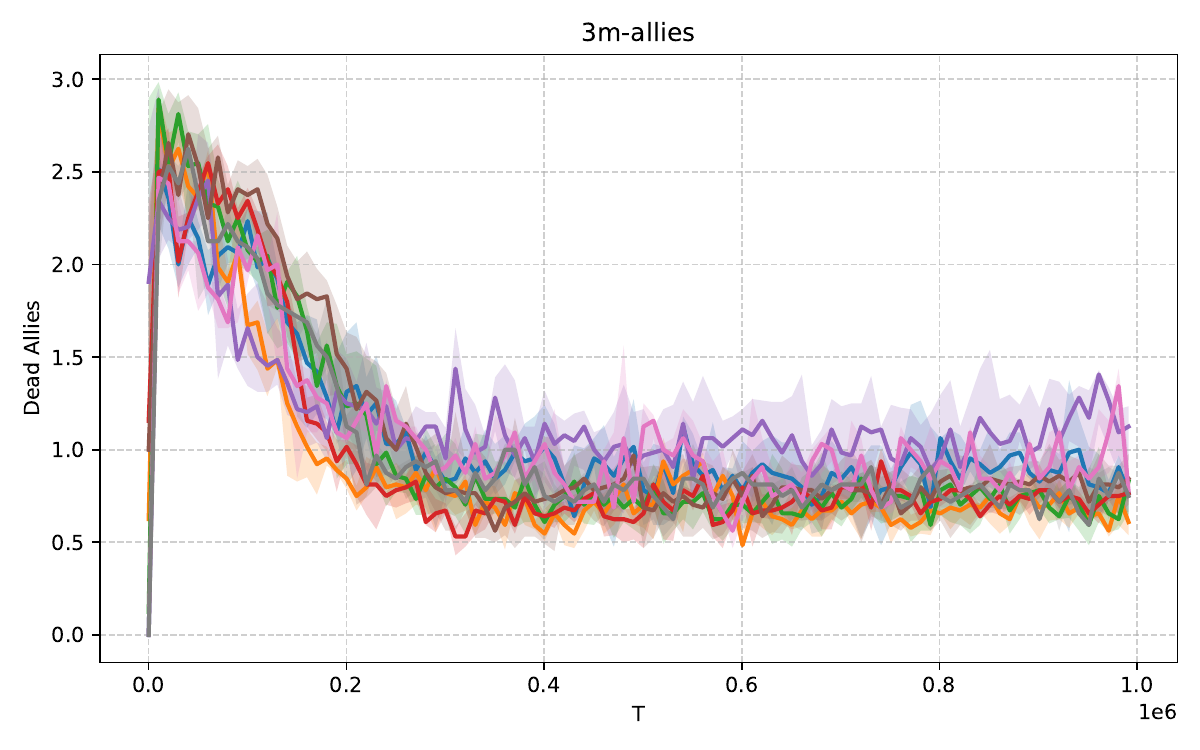}
        \caption{3m}
        \label{fig:3m-allies}
    \end{subfigure}\hfill
    \begin{subfigure}[t]{0.32\textwidth}
        \centering
        \includegraphics[width=\linewidth]{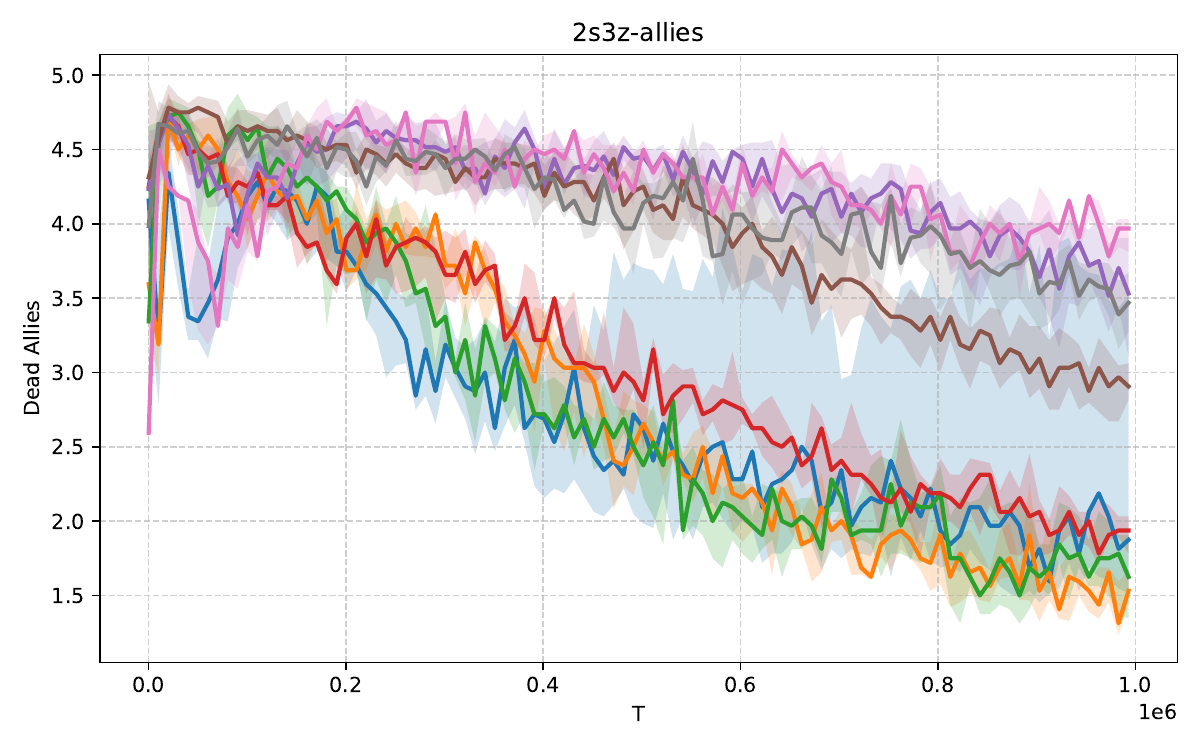}
        \caption{2s3z}
        \label{fig:2s3z-allies}
    \end{subfigure}\hfill
    \begin{subfigure}[t]{0.32\textwidth}
        \centering
        \includegraphics[width=\linewidth]{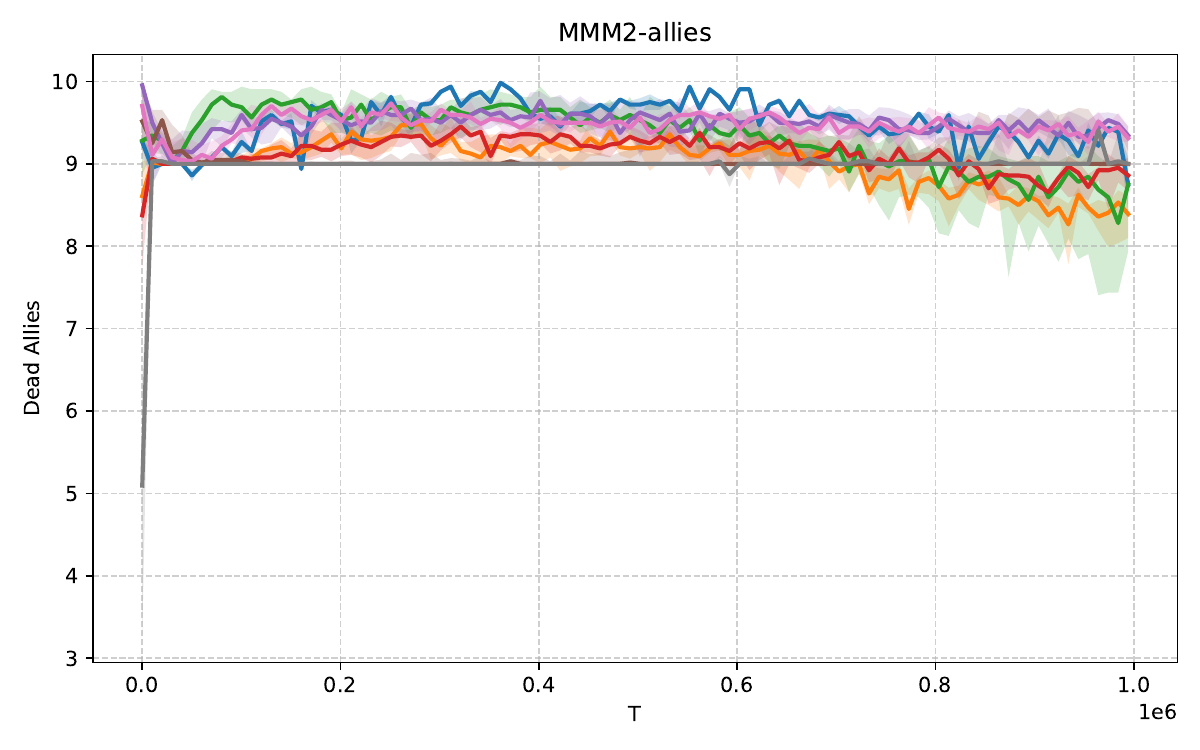}
        \caption{MMM2}
        \label{fig:MMM2-allies}
    \end{subfigure}

    \caption{Learning curves on different SMAC combat maps, showing median dead ally agents with 25-75\% percentile.}
    \label{fig:allied_dead}
\end{figure*}

\begin{figure*}[t!]
    \centering
        \begin{subfigure}[t]{\textwidth}
        \centering
        \includegraphics[width=0.99\linewidth]{img/legend_only.pdf}
    \end{subfigure}

    \begin{subfigure}[t]{0.32\textwidth}
        \includegraphics[width=\linewidth]{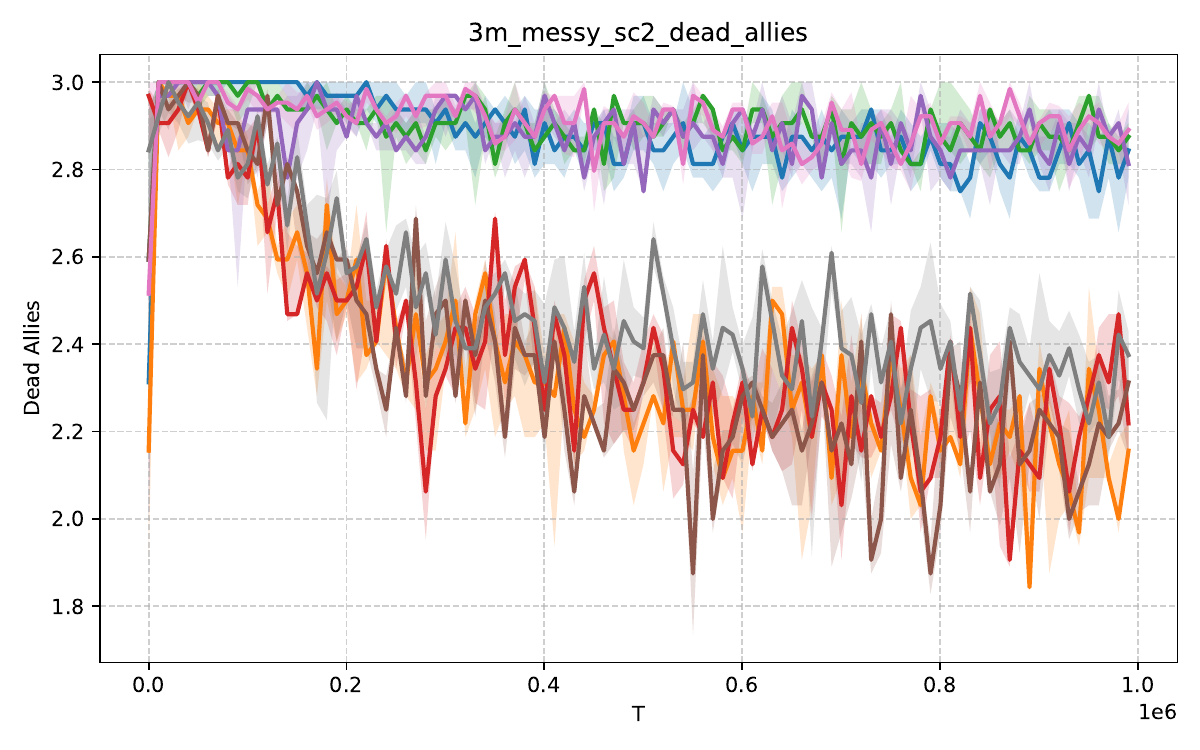}
        \caption{3m}
        \label{fig:3m-messy-allies}
    \end{subfigure}\hfill
    \begin{subfigure}[t]{0.32\textwidth}
        \centering
        \includegraphics[width=\linewidth]{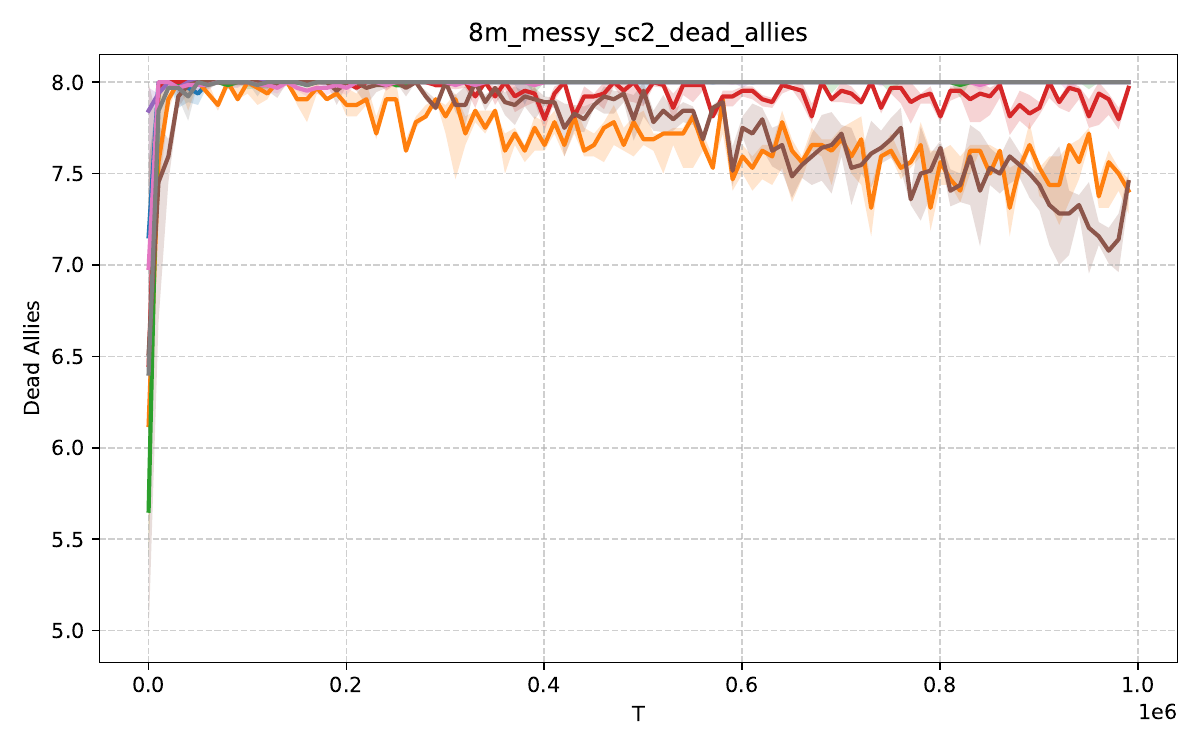}
        \caption{8m}
        \label{fig:8m-messy-allies}
    \end{subfigure}\hfill
    \begin{subfigure}[t]{0.32\textwidth}
        \includegraphics[width=\linewidth]{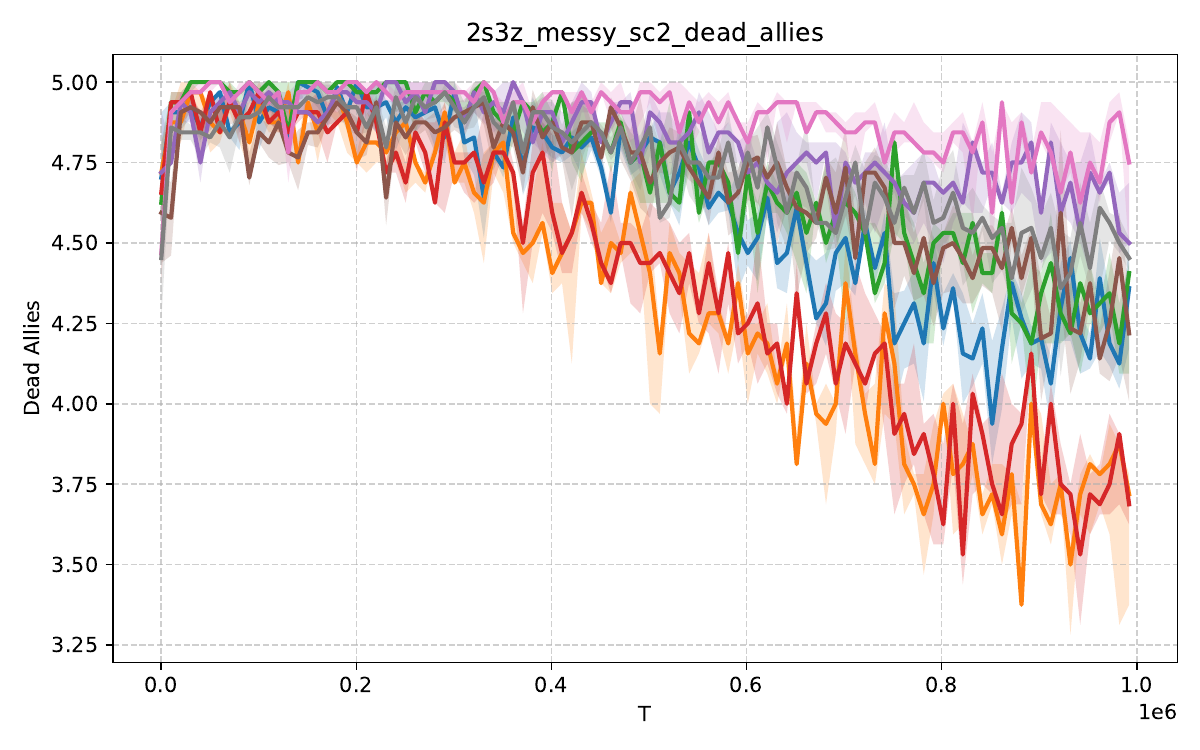}
        \caption{2s3z}
        \label{fig:2s3z-messy-allies}
    \end{subfigure}

    \begin{subfigure}[t]{0.32\textwidth}
        \includegraphics[width=\linewidth]{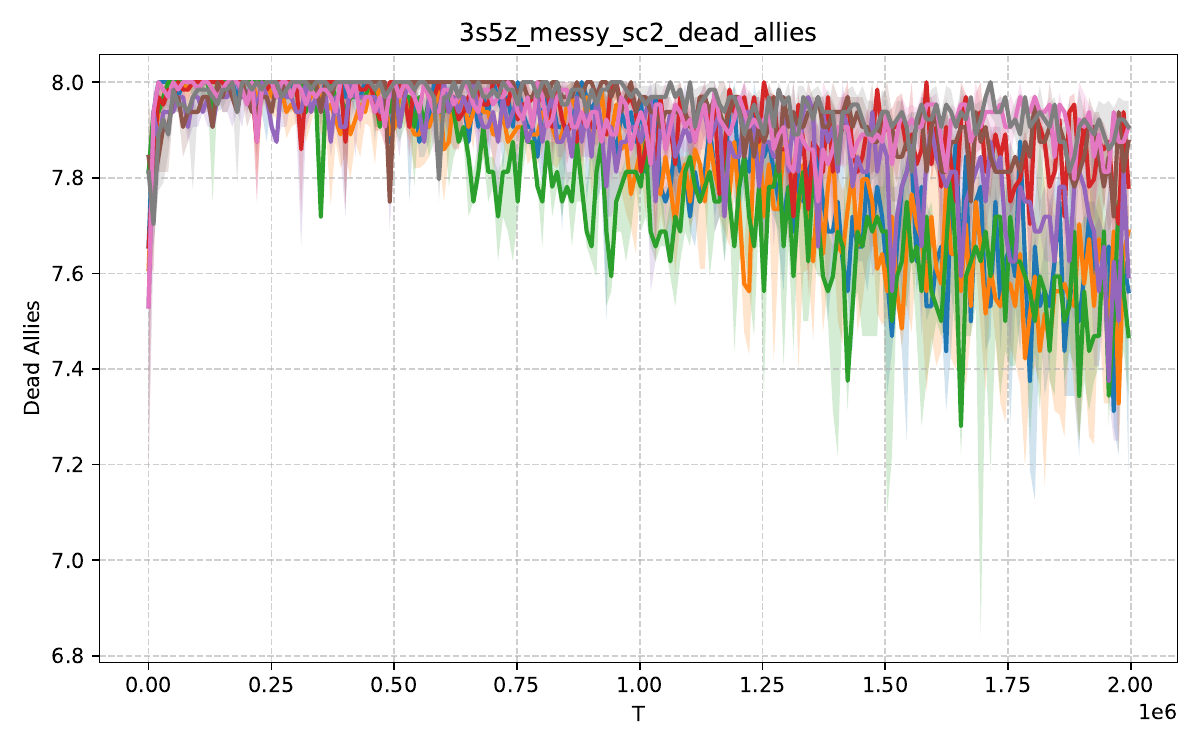}
        \caption{3s5z}
        \label{fig:3s5z-messy-allies}
    \end{subfigure}
    \begin{subfigure}[t]{0.32\textwidth}
        \includegraphics[width=\linewidth]{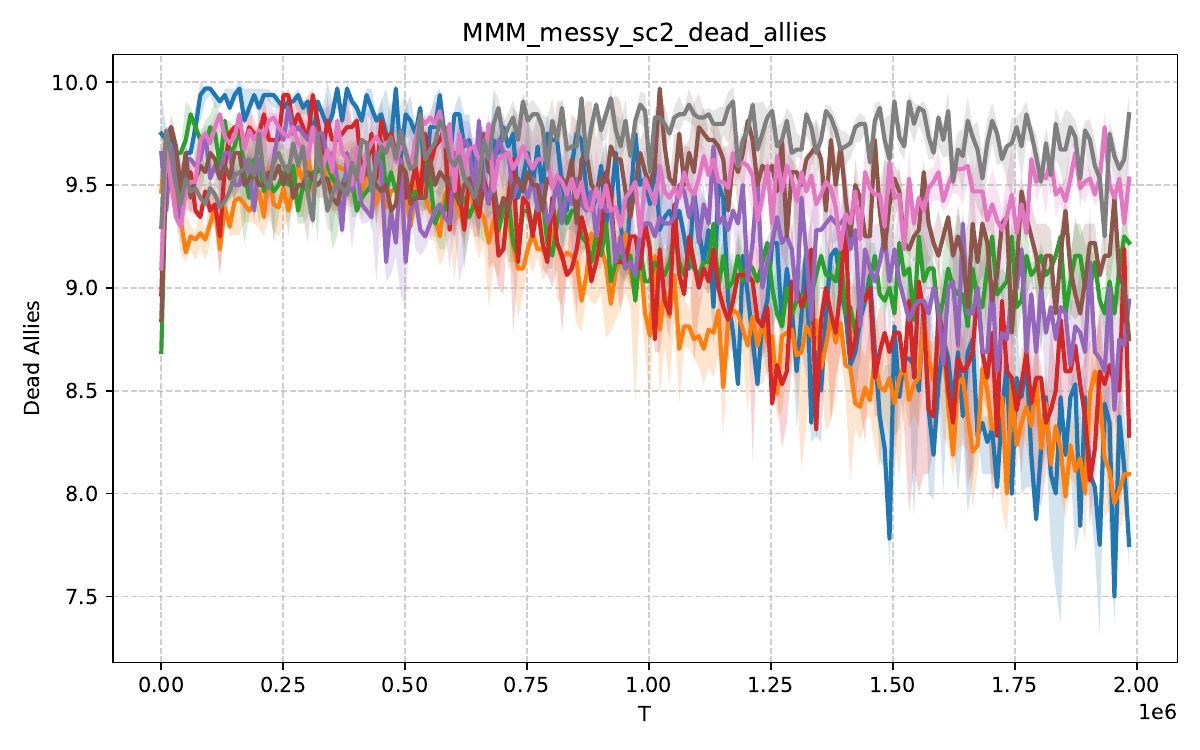}
        \caption{MMM}
        \label{fig:MMM-messy-allies}
    \end{subfigure}\hfill

    \caption{Learning curves on different MessySMAC combat maps, showing median dead ally agents with 25-75\% percentile.}
    \label{fig:allied_dead_messy}
\end{figure*}

\begin{figure*}[t!]

    \centering

        \begin{subfigure}[t]{\textwidth}
        \centering
        \includegraphics[width=0.99\linewidth]{img/legend_only.pdf}
    \end{subfigure}

    \begin{subfigure}[t]{0.32\textwidth}
        \centering
        \includegraphics[width=\linewidth]{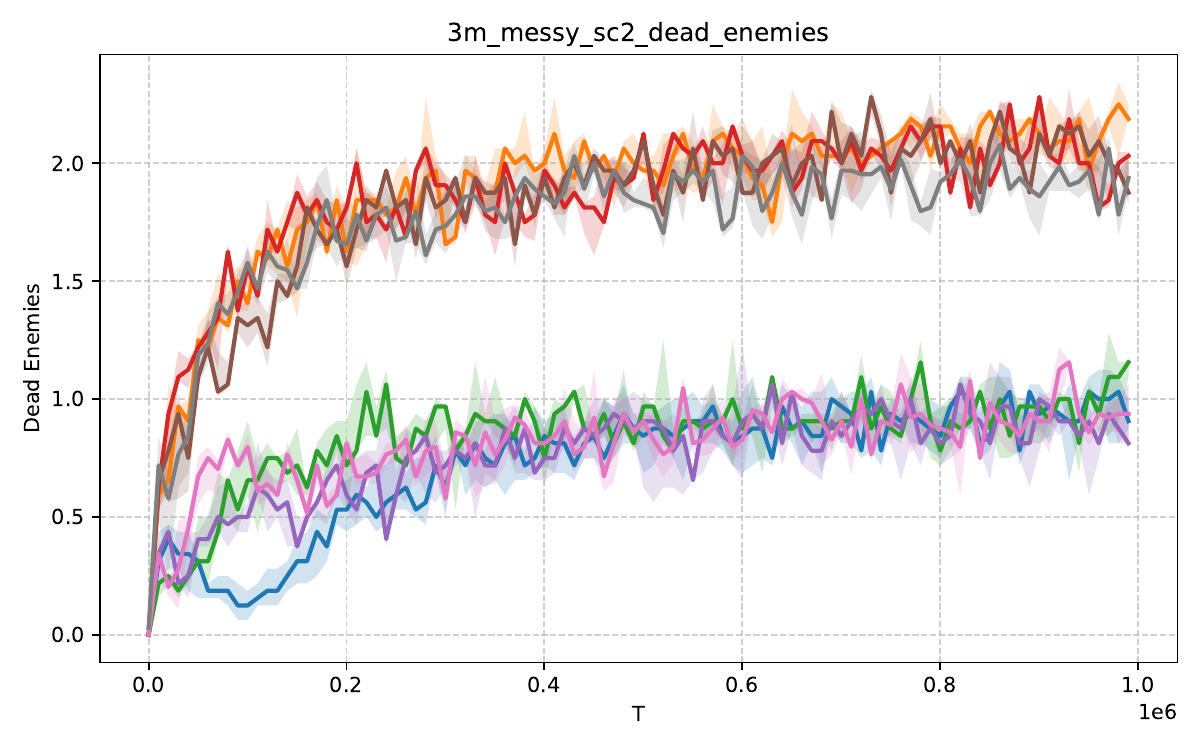}
        \caption{3m}
        \label{fig:3m-messy-enemy}
    \end{subfigure}\hfill
    \begin{subfigure}[t]{0.32\textwidth}
        \centering
        \includegraphics[width=\linewidth]{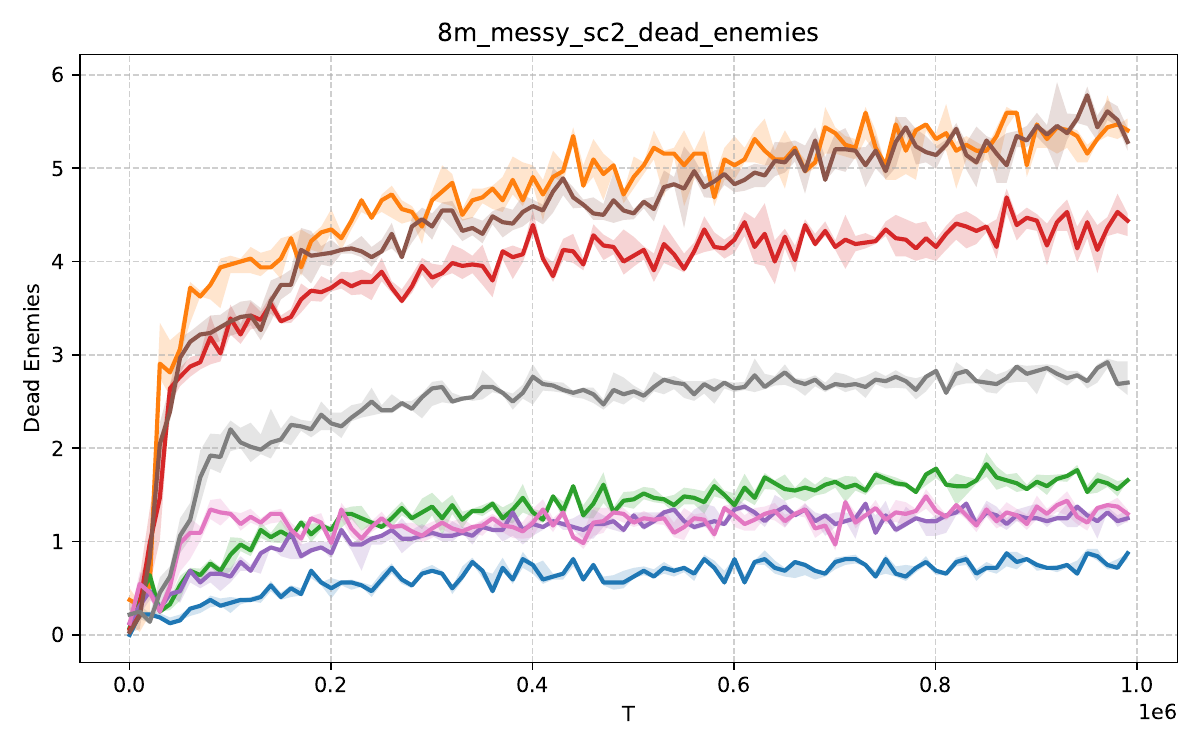}
        \caption{8m}
        \label{fig:8m-messy-enemy}
    \end{subfigure}\hfill
    \begin{subfigure}[t]{0.32\textwidth}
        \centering
        \includegraphics[width=\linewidth]{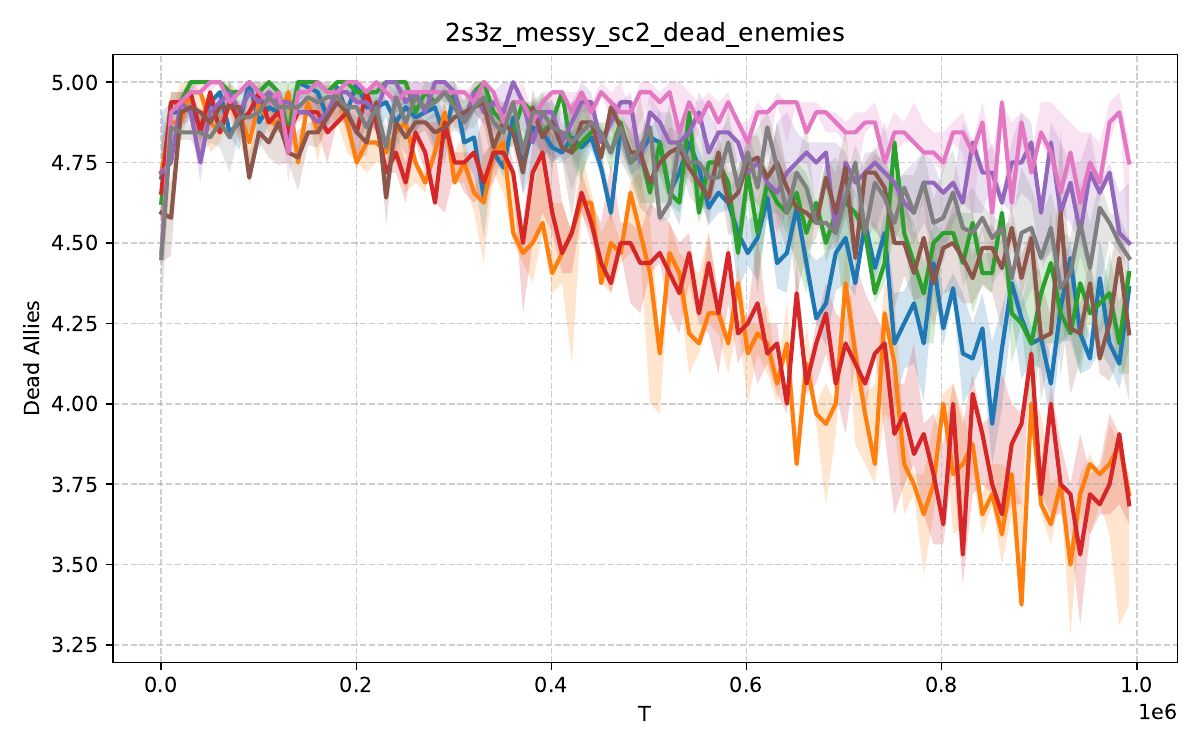}
        \caption{2s3z}
        \label{fig:2s3z-messy-enemies}
    \end{subfigure}

    \begin{subfigure}[t]{0.32\textwidth}
        \centering
        \includegraphics[width=\linewidth]{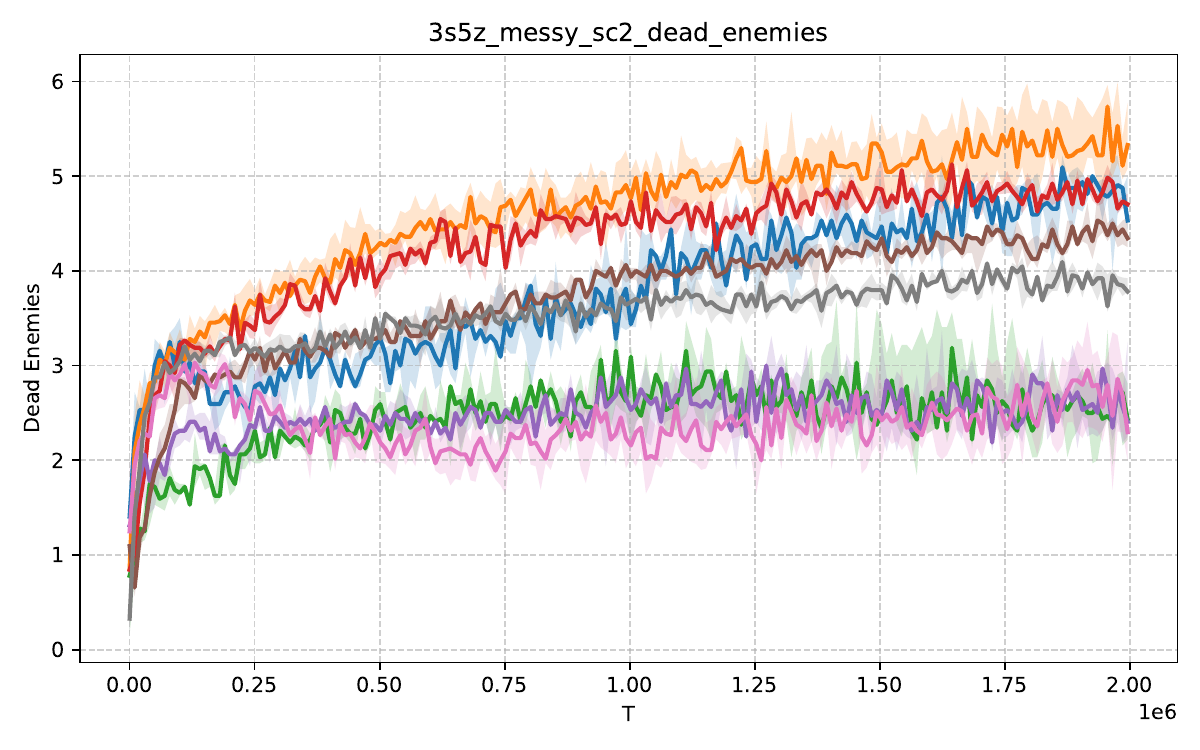}
        \caption{3s5z}
        \label{fig:3s5z-messy-enemies}
    \end{subfigure}
    \begin{subfigure}[t]{0.32\textwidth}
        \centering
        \includegraphics[width=\linewidth]{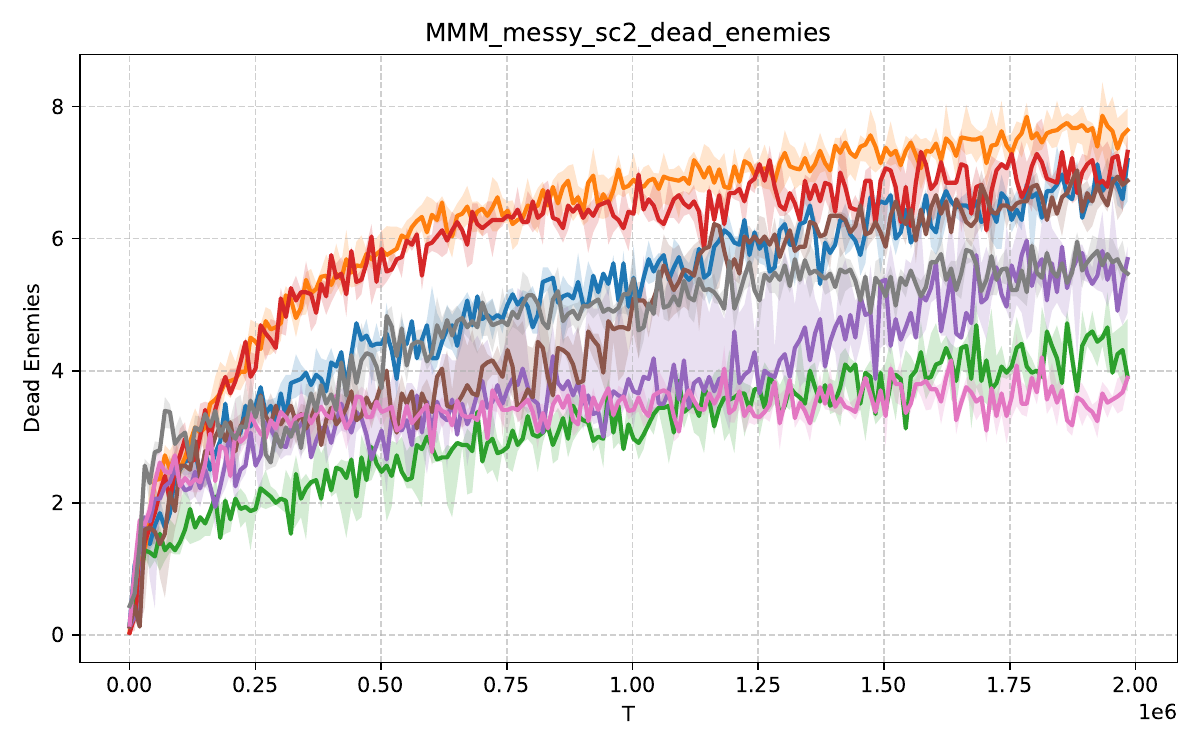}
        \caption{MMM}
        \label{fig:MMM-messy-enemies}
    \end{subfigure}\hfill

    \caption{Learning curves on different MessySMAC showing median dead enemy agents with 25-75\% percentile.}
    \label{fig:enemy_dead_messy}
\end{figure*}

We also report the number of dead enemies and dead allies across all SMAC maps in~\Cref{fig:enemy_dead,fig:allied_dead} and all MessySMAC maps in~\Cref{fig:enemy_dead_messy,fig:allied_dead_messy}.

\begin{figure*}[]
    \centering
        \begin{subfigure}[t]{\textwidth}
        \centering
        \includegraphics[width=0.99\linewidth]{img/legend_only.pdf}
    \end{subfigure}

    \begin{subfigure}[t]{0.32\textwidth}
        \centering
        \includegraphics[width=\linewidth]{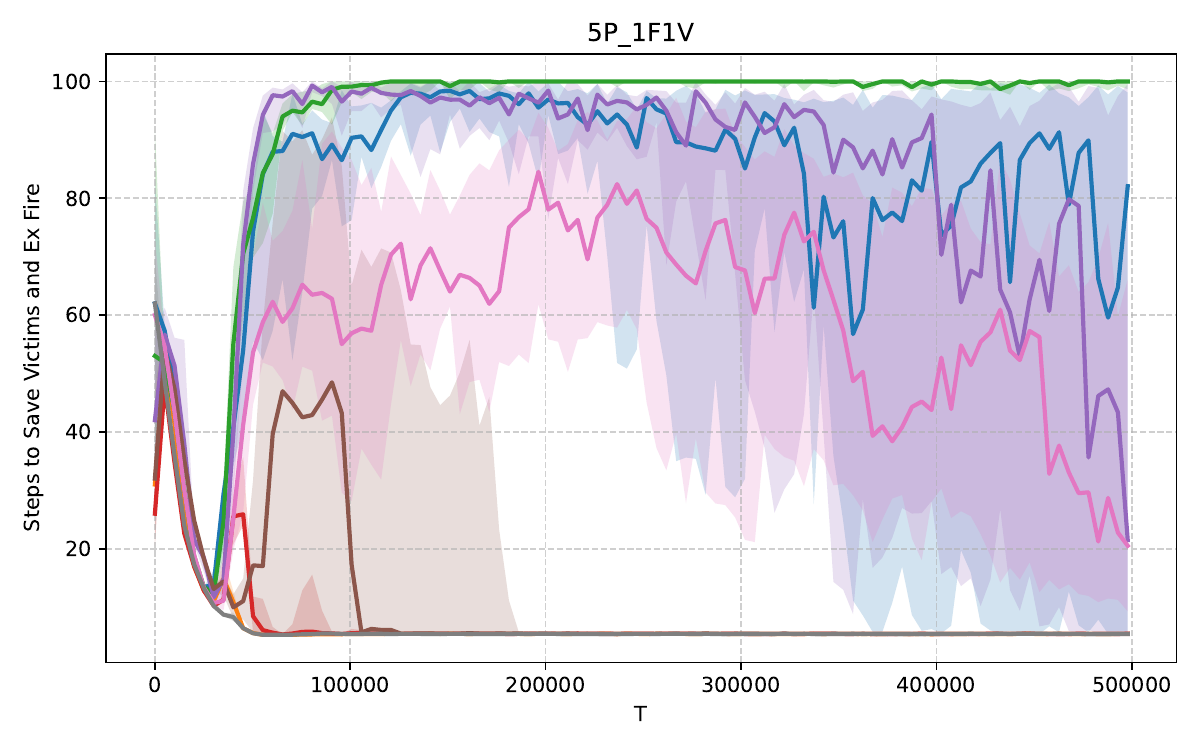}
        \caption{5P\_1F1V ($R_{\text{Wild\_1}}$)}
        \label{fig:5P_1F1V}
    \end{subfigure}\hfill
    \begin{subfigure}[t]{0.32\textwidth}
        \centering
        \includegraphics[width=\linewidth]{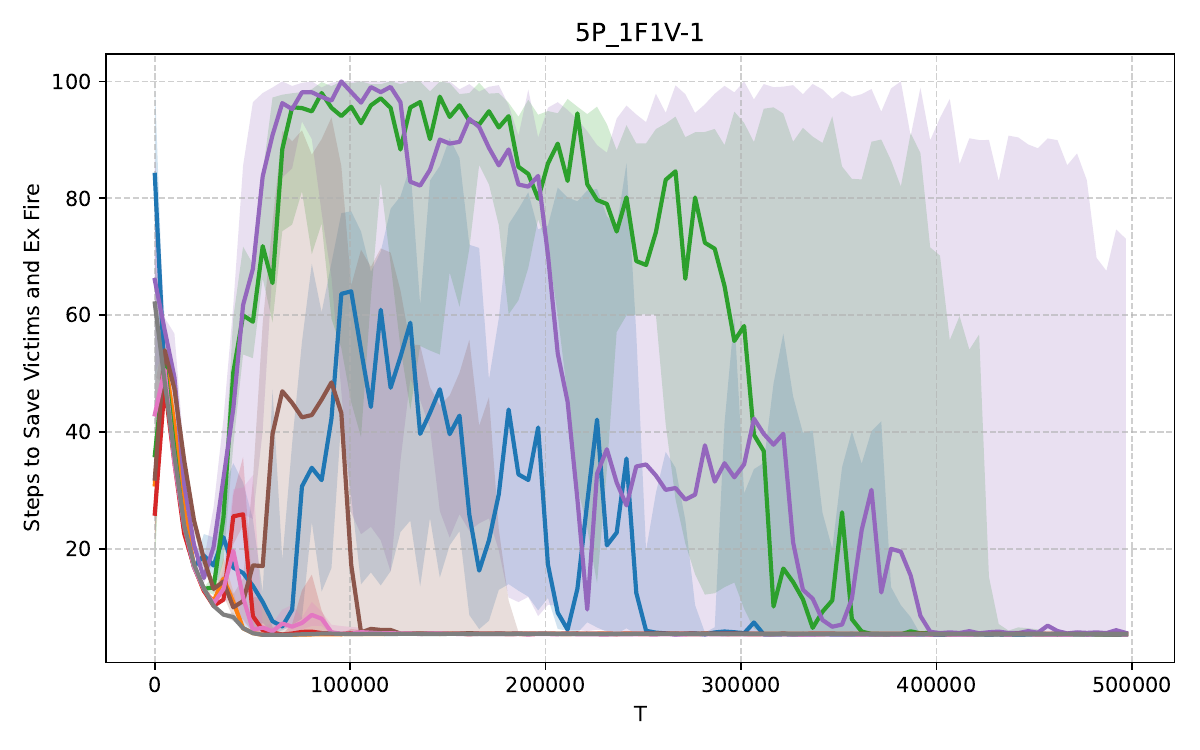}
        \caption{5P\_1F1V ($R_{\text{Wild\_2}}$)}
        \label{fig:5P_1F1V-1}
    \end{subfigure}\hfill
    \begin{subfigure}[t]{0.32\textwidth}
        \centering
        \includegraphics[width=\linewidth]{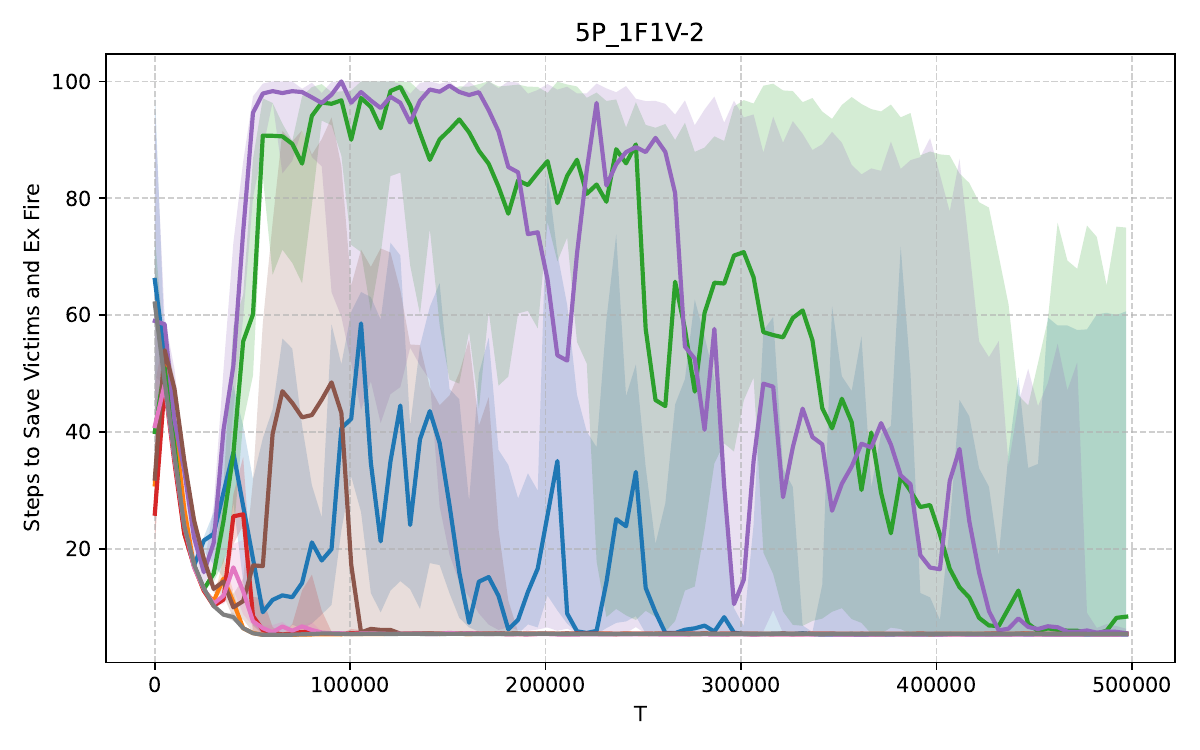}
        \caption{5P\_1F1V ($R_{\text{Wild\_3}}$)}
        \label{fig:5P_1F1V-2}
    \end{subfigure}

    \begin{subfigure}[t]{0.32\textwidth}
        \centering
        \includegraphics[width=\linewidth]{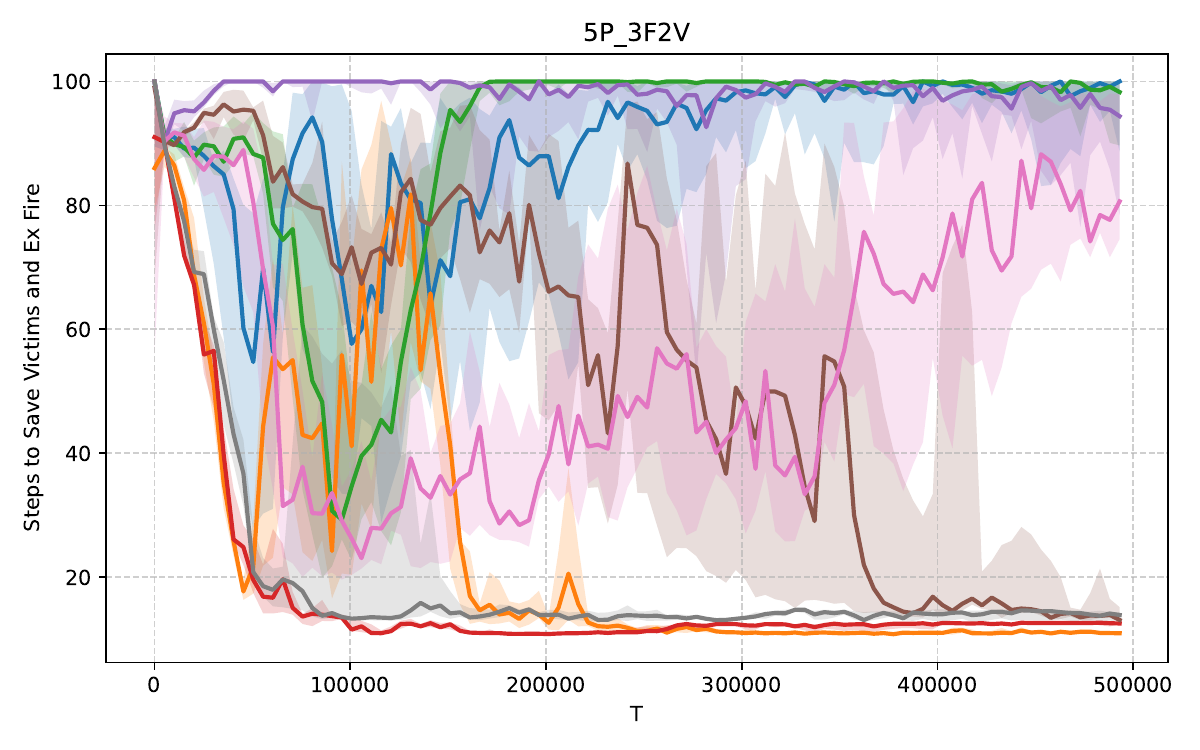}
        \caption{5P\_3F2V ($R_{\text{Wild\_1}}$)}
        \label{fig:5P_3F2V}
    \end{subfigure}\hfill
    \begin{subfigure}[t]{0.32\textwidth}
        \centering
        \includegraphics[width=\linewidth]{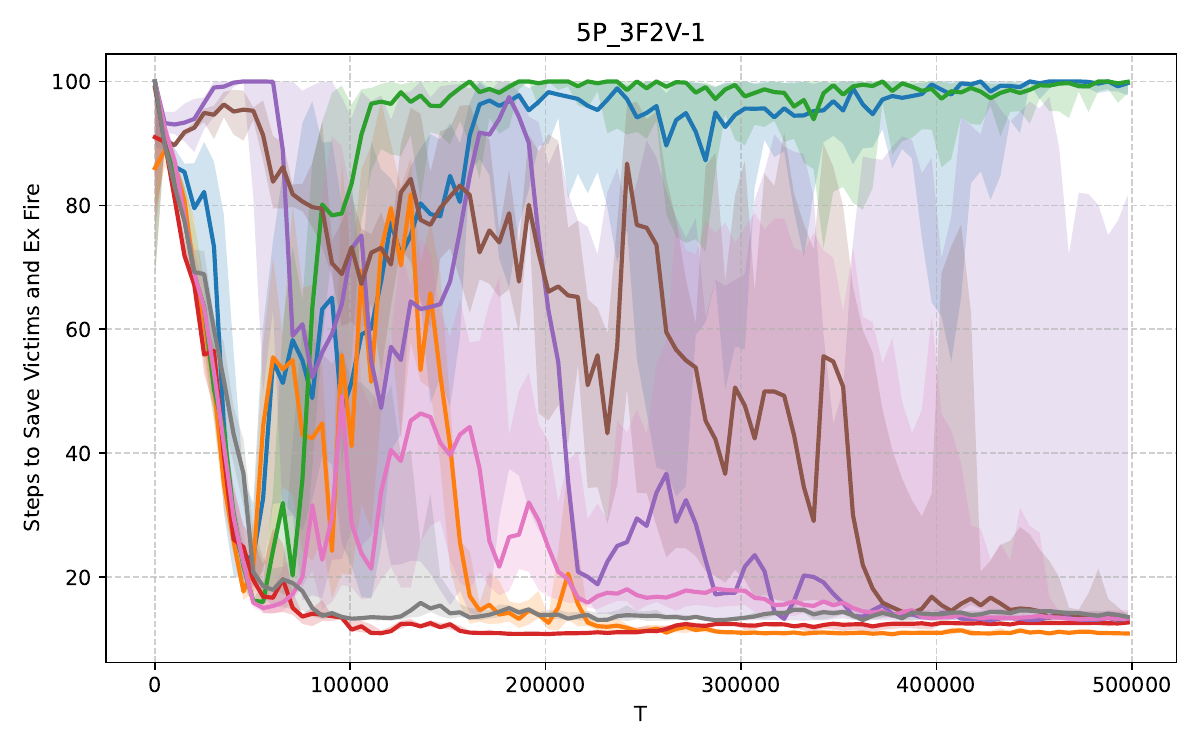}
        \caption{5P\_3F2V ($R_{\text{Wild\_2}}$)}
        \label{fig:5P_3F2V-1}
    \end{subfigure}\hfill
    \begin{subfigure}[t]{0.32\textwidth}
        \centering
        \includegraphics[width=\linewidth]{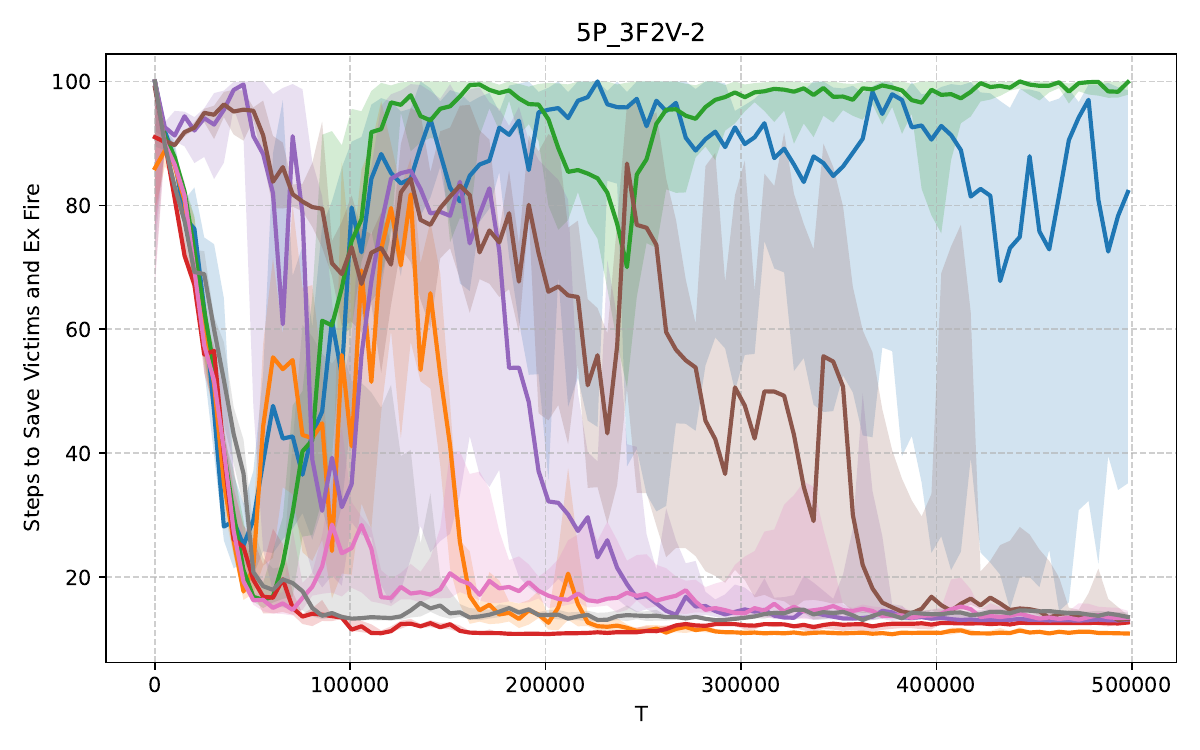}
        \caption{5P\_3F2V ($R_{\text{Wild\_3}}$)}
        \label{fig:5P_3F2V-2}
    \end{subfigure}

    \begin{subfigure}[t]{0.32\textwidth}
        \centering
        \includegraphics[width=\linewidth]{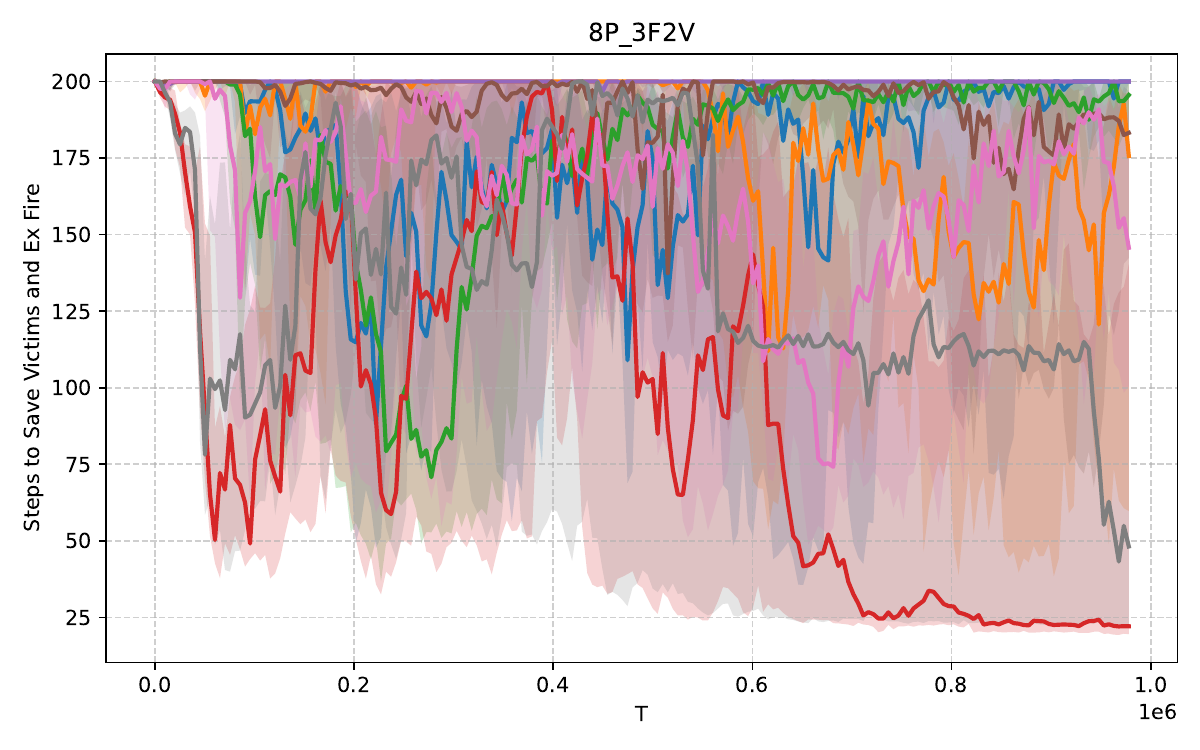}
        \caption{8P\_3F2V ($R_{\text{Wild\_1}}$)}
        \label{fig:8P_3F2V}
    \end{subfigure}\hfill
    \begin{subfigure}[t]{0.32\textwidth}
        \centering
        \includegraphics[width=\linewidth]{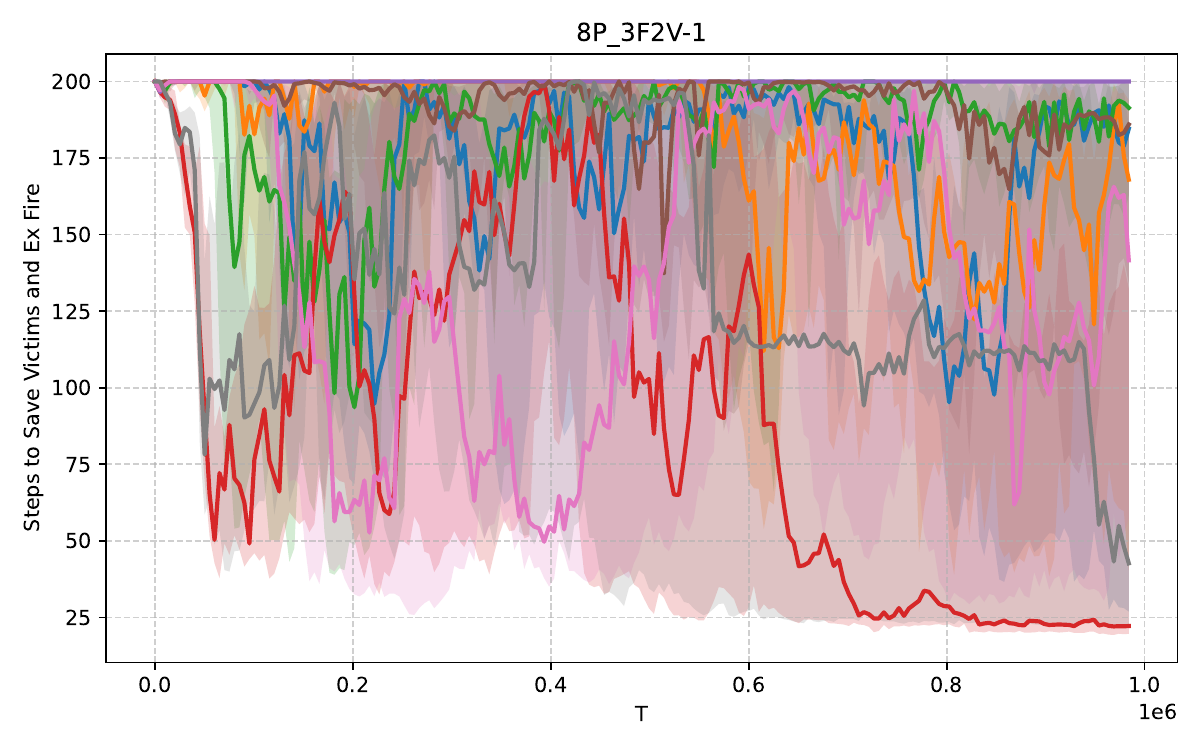}
        \caption{8P\_3F2V ($R_{\text{Wild\_2}}$)}
        \label{fig:8P_3F2V-1}
    \end{subfigure}\hfill
    \begin{subfigure}[t]{0.32\textwidth}
        \centering
        \includegraphics[width=\linewidth]{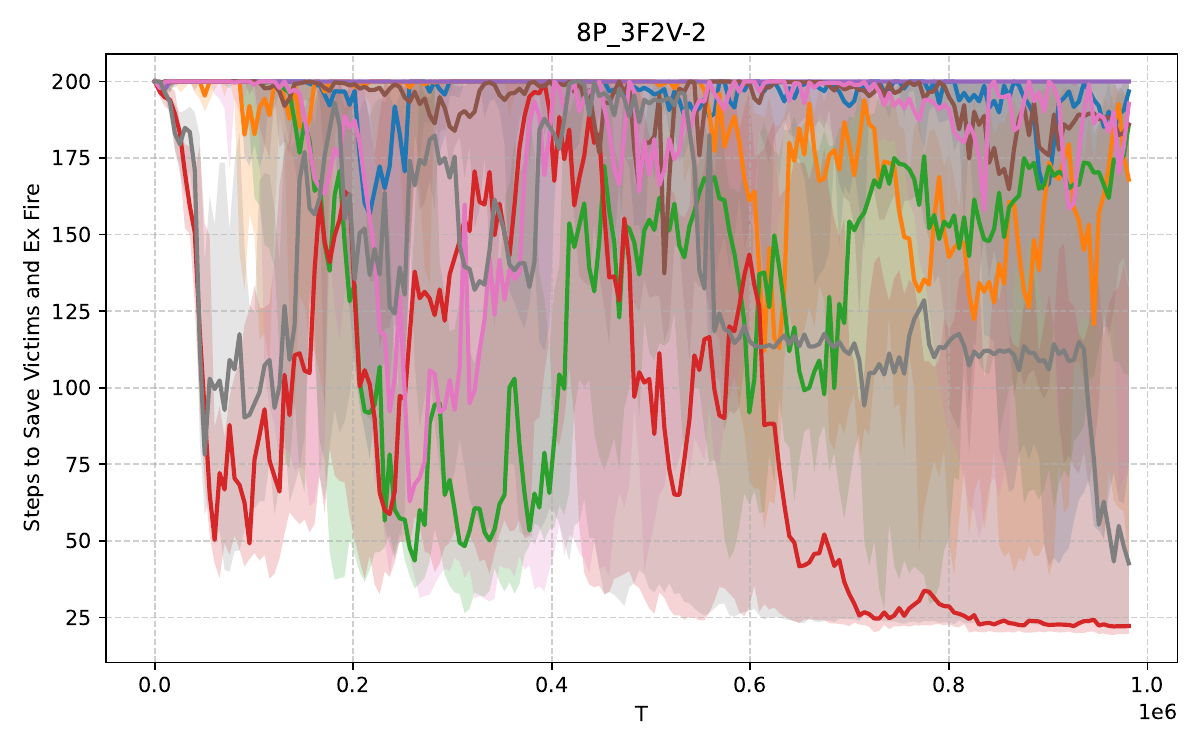}
        \caption{8P\_3F2V ($R_{\text{Wild\_3}}$)}
        \label{fig:8P_3F2V-2}
    \end{subfigure}

        \begin{subfigure}[t]{0.32\textwidth}
        \centering
        \includegraphics[width=\linewidth]{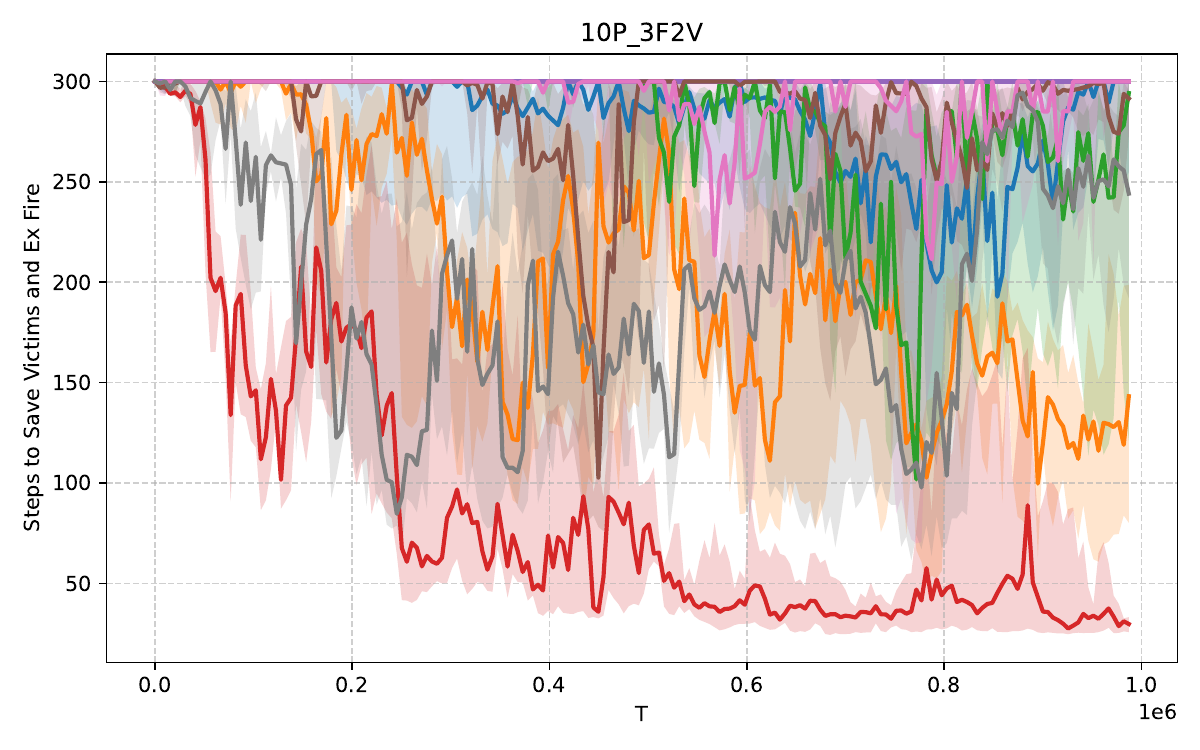}
        \caption{10P\_3F2V ($R_{\text{Wild\_1}}$)}
        \label{fig:10P_3F2V}
    \end{subfigure}\hfill
    \begin{subfigure}[t]{0.32\textwidth}
        \centering
        \includegraphics[width=\linewidth]{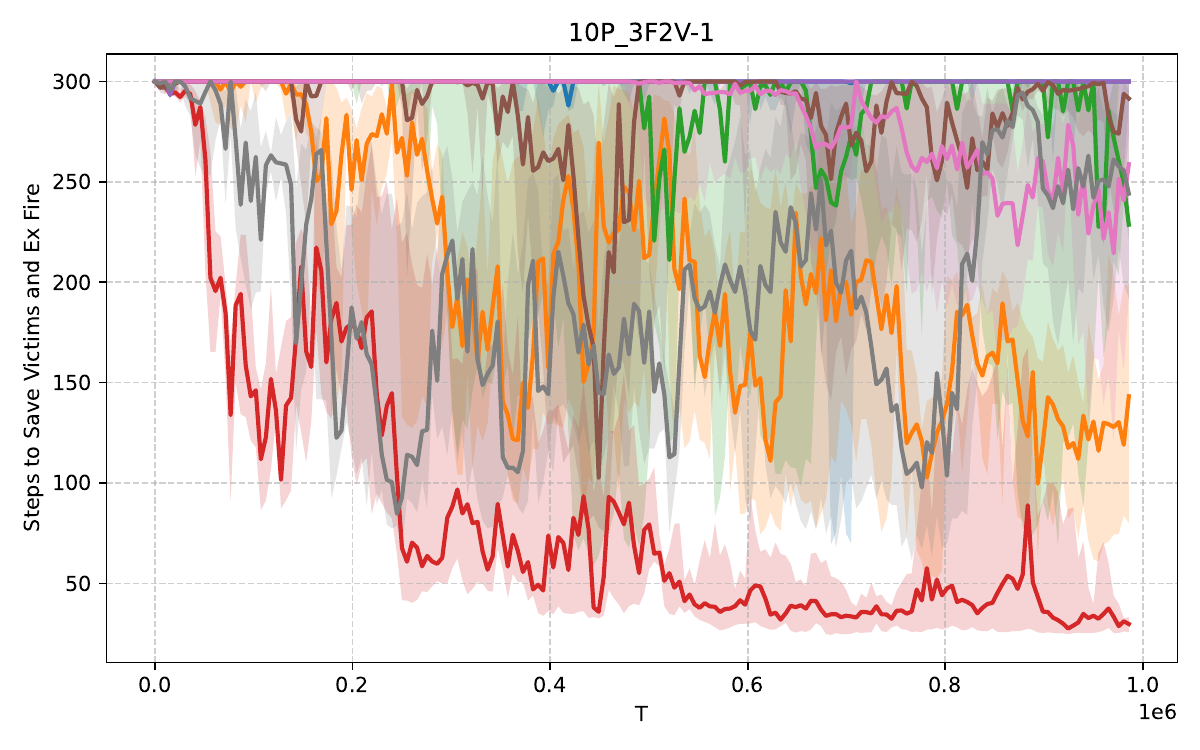}
        \caption{10P\_3F2V ($R_{\text{Wild\_2}}$)}
        \label{fig:10P_3F2V-1}
    \end{subfigure}\hfill
    \begin{subfigure}[t]{0.32\textwidth}
        \centering
        \includegraphics[width=\linewidth]{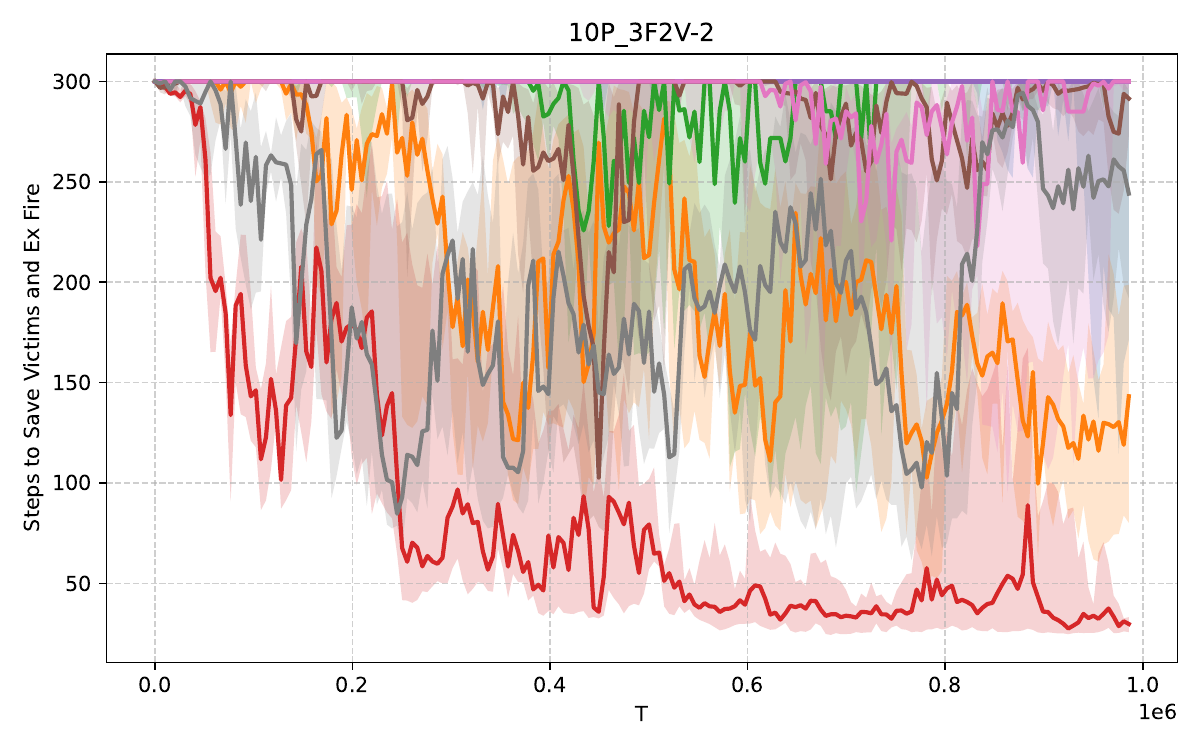}
        \caption{10P\_3F2V ($R_{\text{Wild\_3}}$)}
        \label{fig:10P_3F2V-2}
    \end{subfigure}

    \caption{Learning curves on different WildFire Maps, showing median Steps to Extinguish Fire and Save Victim with 25-75\% percentile.}
    \label{fig:wildfire}
\end{figure*}

In \cref{fig:wildfire}, we report results on the WildFire scenarios, focusing on the number of steps required to save victims and extinguish fires.
On the \texttt{5P\_1F1V} map (\cref{fig:5P_1F1V,fig:5P_1F1V-1,fig:5P_1F1V-2}), \method+MARL consistently outperforms the vanilla baselines, requiring fewer steps to complete the objectives. 
On the \texttt{5P\_3F2V} map (\cref{fig:5P_3F2V,fig:5P_3F2V-1,fig:5P_3F2V-2}), \method+MARL outperforms the vanilla baselines in most cases, except that QTRAN+$R_{\text{Wild\_2}}$ and QTRAN+$R_{\text{Wild\_3}}$ \purpleline achieve better performance than \method+QTRAN \brownline.
On the \texttt{8P\_3F2V} map (\cref{fig:8P_3F2V,fig:8P_3F2V-1,fig:8P_3F2V-2}), \method+MARL outperforms the vanilla baselines in all cases; notably, \method+VDN \redline and \method+IQL \grayline perform substantially better than the vanilla variants across all baseline reward functions.
On the \texttt{10P\_3F2V} map (\cref{fig:10P_3F2V,fig:10P_3F2V-1,fig:10P_3F2V-2}), \method+MARL outperforms the vanilla baselines in all cases; in particular, \method+VDN \redline and \method+QMIX \orangeline significantly improve over the vanilla variants under all baseline reward functions.

\subsection{$\boldsymbol{\forall\exists}$ vs.\ $\boldsymbol{\forall\forall}$ Specifications.}
\label{sec:aaee-app}
\begin{figure*}[]
    \centering

    \vspace{-2mm}
    \begin{subfigure}[t]{\textwidth}
        \centering
        \includegraphics[width=0.99\linewidth]{img/leg_AE_vs_AA.pdf}
    \end{subfigure}
\vspace{-5mm}

    \begin{subfigure}[t]{0.4\linewidth}
        \centering
        \includegraphics[width=\linewidth]{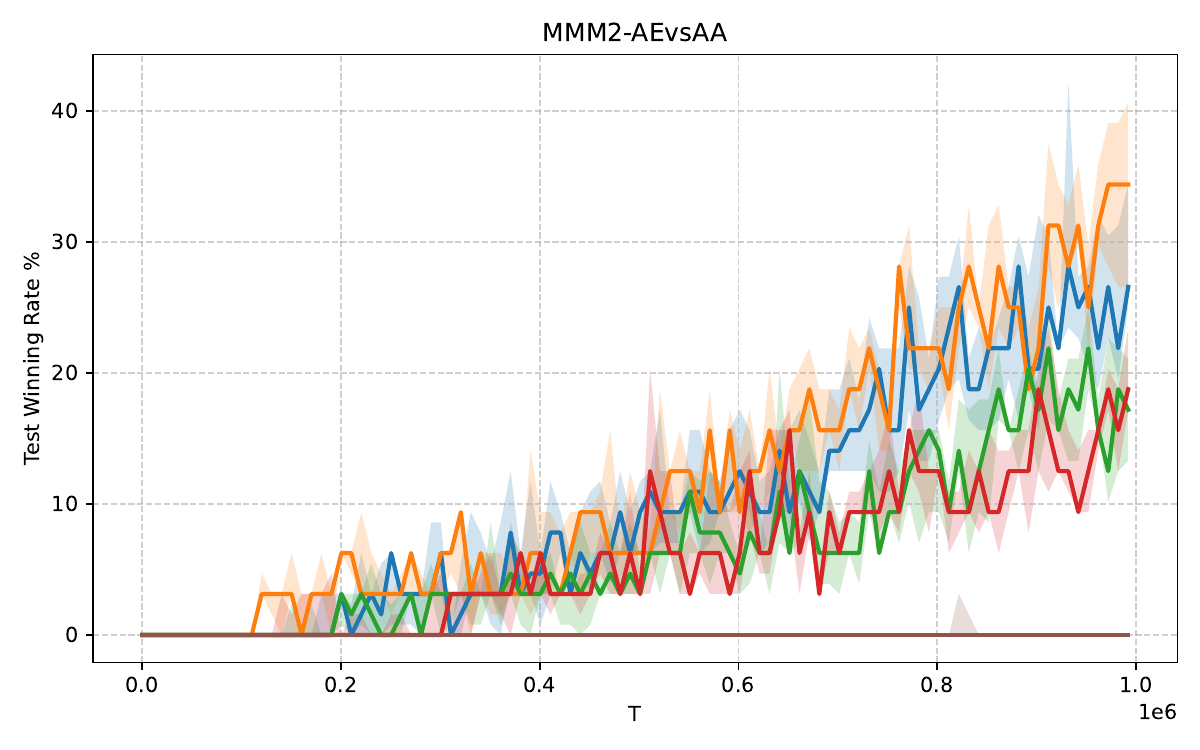}
        \caption{MMM2 (SMAC)}
        \label{fig:quantMMM2}
    \end{subfigure}\hfill
    \begin{subfigure}[t]{0.4\linewidth}
        \centering
        \includegraphics[width=\linewidth]{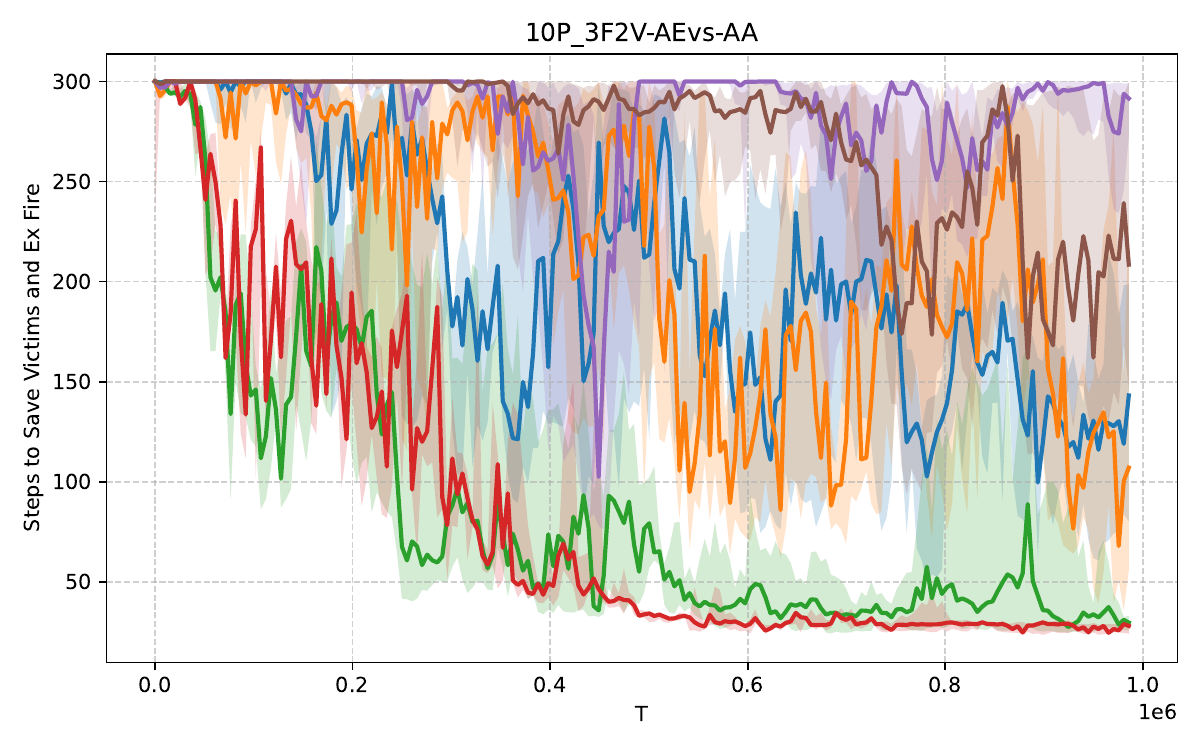}
        \caption{10P\_3F2V (WildFire)}
        \label{fig:quantwild10}
    \end{subfigure}\hfill

    \caption{$\forall\forall$ vs. $\forall\exists$ learning curves across SMAC, and WildFire, showing median test win rate with 25--75\% percentiles.}
    \vspace*{-2mm}
    \label{fig:quantefiers-app}
\end{figure*}

Since \texttt{MMM2} includes a Medivac unit, we evaluate the $\medivac$ specification using
$\forall\exists$-\HyperLTL formulas. As shown in~\Cref{fig:quantMMM2}, \method($\forall\exists$)+QMIX
\orangeline performs slightly better than \method($\forall\forall$)+QMIX \blueline.
We also provide same analysis for 10P\_3F2V map in WildFire \cref{fig:quantwild10} and we saw \method($\forall\exists$)+QTRAN \brownline performs better than \method($\forall\forall$)+QTRAN \purpleline.

\subsection{Visualization of tactics Learn by \method on SMAC}

\begin{figure*}[t!]

    \begin{subfigure}[t]{0.45\textwidth}
        \centering
        \includegraphics[width=\linewidth]{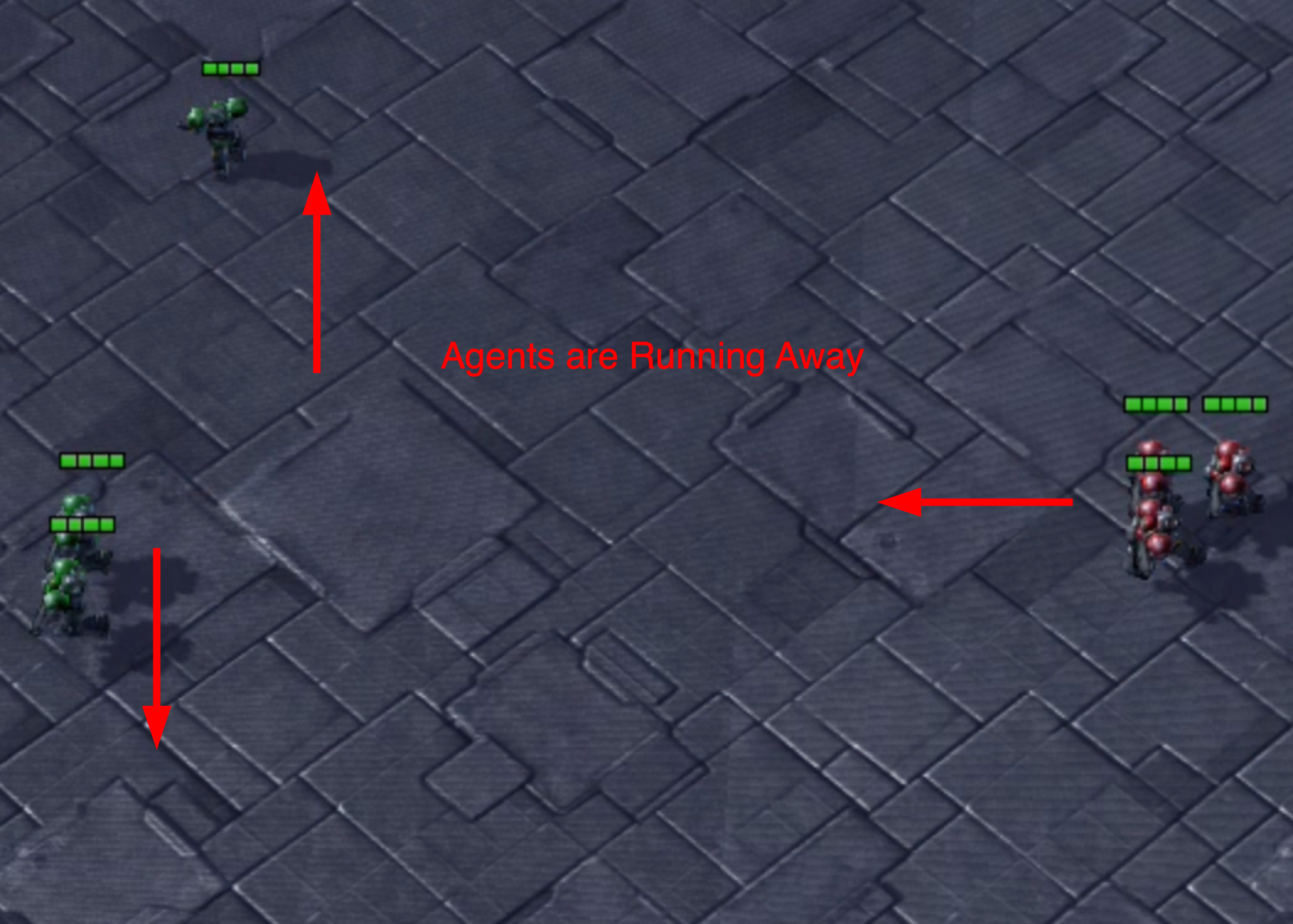}
        \caption{$\defense$ Tactic}
        \label{fig:defense-tac}
    \end{subfigure}\hfill
    \begin{subfigure}[t]{0.45\textwidth}
        \centering
        \includegraphics[width=\linewidth]{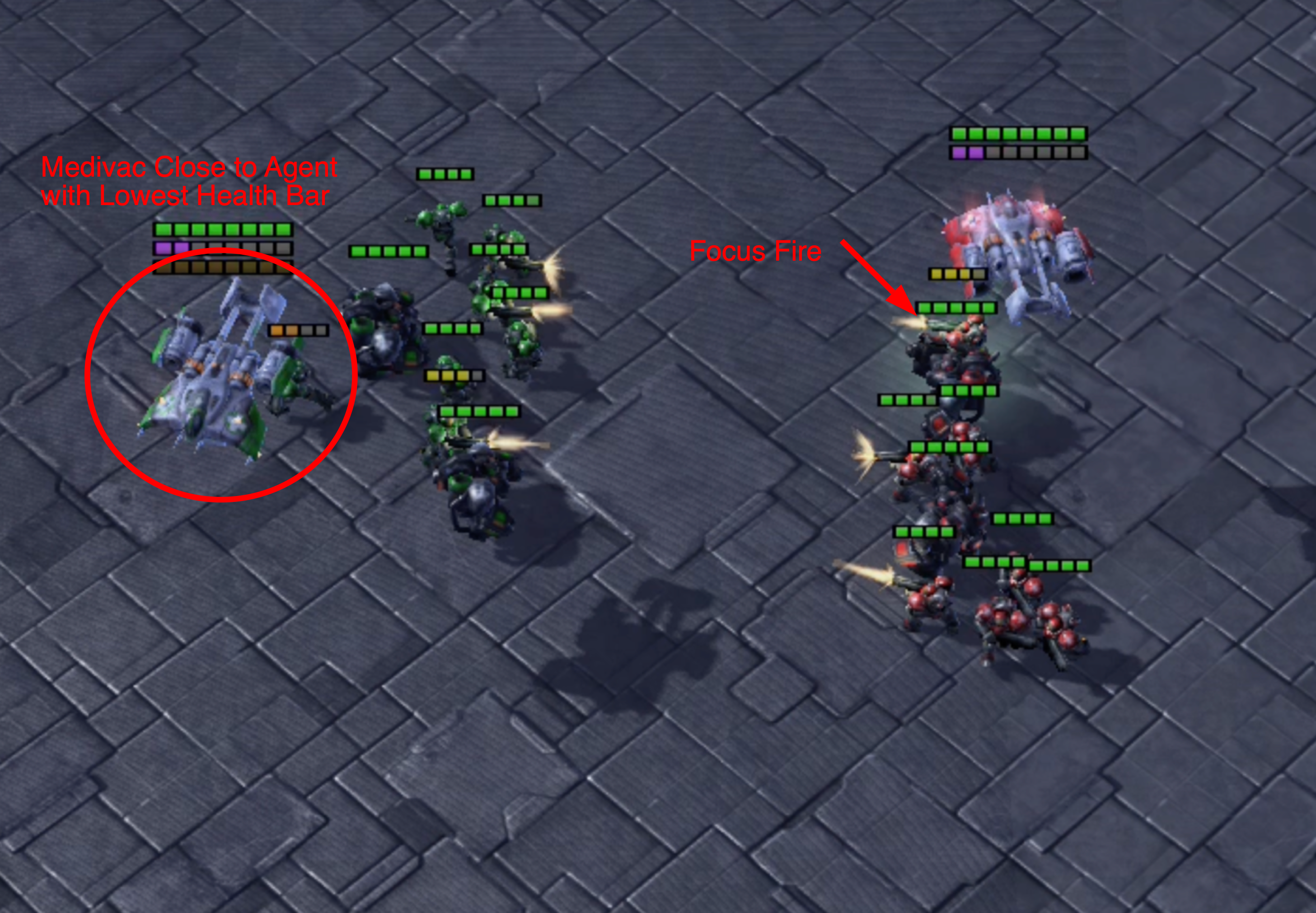}
        \caption{$\medivac$ Tactic}
        \label{fig:medivac-tac}
    \end{subfigure}

    \caption{Snapshots of scenarios under \HyperLTL-specified tactics.}
    \label{fig:Tactics}
\end{figure*}

\Cref{fig:Tactics} shows snapshots after 1M training steps in SMAC.
In \cref{fig:defense-tac} using $\defense$, the agents instead prioritize survival and retreat, moving away from enemies rather than committing to an engagement.
Under $\medivac$ \cref{fig:medivac-tac}, the Medivac consistently moves toward the ally with the lowest health bar, indicating that the learned policy captures the intended support behavior. 
At the same time, the remaining agents exhibit focused firing by concentrating fire on a specific enemy target. 
Both \cref{fig:defense-tac,fig:medivac-tac} generated by VDN.

\subsection{\method Computational Overhead}

Next, we evaluate the computational overhead introduced by \method within the CTDE training pipeline.
We run an experiment on \texttt{MMM}, with results reported in \cref{tab:run}.
In the $\forall\forall$ setting, \method incurs a negligible overhead of about 6\% at 1000K environment steps.
In contrast, in the $\forall\exists$ setting, the overhead increases to 62\%, which is substantial.
This experiment is run on an AMD EPYC 7742 64-core CPU and a single NVIDIA RTX A6000 GPU.

\begin{table}[t]
  \caption{Mean execution time on the MMM map comparing MARL with and without \method. Results for the $\forall\exists$ formula are averaged only over VDN, QMIX, and QTRAN.}
  \label{tab:run}
  \begin{center}
        \begin{tabular}{lccccc}
          \toprule
          \textbf{Method}  & \textbf{200K Steps}         & \textbf{400K Steps}      & \textbf{600K Steps}  & \textbf{800K Steps}  & \textbf{1000K Steps}  \\
          \midrule
          \textbf{MARL} & 2202 & 4294 & 6305 & 8377& 10385 \\
        \method ($\forall\forall$) + \textbf{MARL} & 2211 & 4453 & 6637 & 8918 & 11100 \\
          \method($\forall\exists$) + \textbf{CTDE} & 3366 &6821  & 10260  & 13659 &  16871\\
          \bottomrule
        \end{tabular}
  \end{center}
\end{table}

